\documentclass[10pt,journal,compsoc]{IEEEtran}

\usepackage[usenames, dvipsnames]{color}
\usepackage{amsmath}
\usepackage{array}
\usepackage{amssymb}
\usepackage[table]{xcolor}

\newcommand{\yes}{\checkmark}

\ifCLASSOPTIONcompsoc
    \usepackage[nocompress]{cite}
\else
    \usepackage{cite}
\fi

\ifCLASSINFOpdf
    \usepackage[pdftex]{graphicx}
    \graphicspath{{./resource/}}
    \DeclareGraphicsExtensions{.pdf,.jpeg,.png}
    \usepackage{epstopdf}
\else
\fi

\ifCLASSOPTIONcompsoc
    \usepackage[caption=false,font=footnotesize,labelfont=sf,textfont=sf]{subfig}
\else
    \usepackage[caption=false,font=footnotesize]{subfig}
\fi

\ifCLASSOPTIONcaptionsoff
    \usepackage[nomarkers]{endfloat}
    \let\MYoriglatexcaption\caption
    \renewcommand{\caption}[2][\relax]{\MYoriglatexcaption[#2]{#2}}
\fi

\begin{document}

\title{Deep Learning for Image Super-resolution:\\A Survey}

\author{
    Zhihao~Wang,
    Jian~Chen,
    Steven~C.H.~Hoi, Fellow, IEEE
    \IEEEcompsocitemizethanks{
        \IEEEcompsocthanksitem Corresponding author: Steven C.H. Hoi is currently with Salesforce Research Asia, and also a faculty member (on leave) of the School of Information Systems, Singapore Management University, Singapore. Email: shoi@salesforce.com or chhoi@smu.edu.sg.
        \IEEEcompsocthanksitem Z. Wang is with the South China University of Technology, China. E-mail: ptkin@outlook.com.
        This work was done when he was a visiting student with Dr Hoi's group at the School of Information Systems, Singapore Management University, Singapore.
        \IEEEcompsocthanksitem J. Chen is with the South China University of Technology, China. E-mail: ellachen@scut.edu.cn.
    }%
}

\IEEEtitleabstractindextext{%

\begin{abstract}

Image Super-Resolution (SR) is an important class of image processing techniques to enhance the resolution of images and videos in computer vision.
Recent years have witnessed remarkable progress of image super-resolution using deep learning techniques.
This article aims to provide a comprehensive survey on recent advances of image super-resolution using deep learning approaches.
In general, we can roughly group the existing studies of SR techniques into three major categories: supervised SR, unsupervised SR, and domain-specific SR.
In addition, we also cover some other important issues, such as publicly available benchmark datasets and performance evaluation metrics.
Finally, we conclude this survey by highlighting several future directions and open issues which should be further addressed by the community in the future.


\end{abstract}


}

\maketitle
\IEEEdisplaynontitleabstractindextext
\IEEEpeerreviewmaketitle

\IEEEraisesectionheading{
    \section{Introduction}
}

\begin{figure*}
    \centering
    {
        \setlength{\fboxrule}{1pt}
        \includegraphics[width=0.95\linewidth]{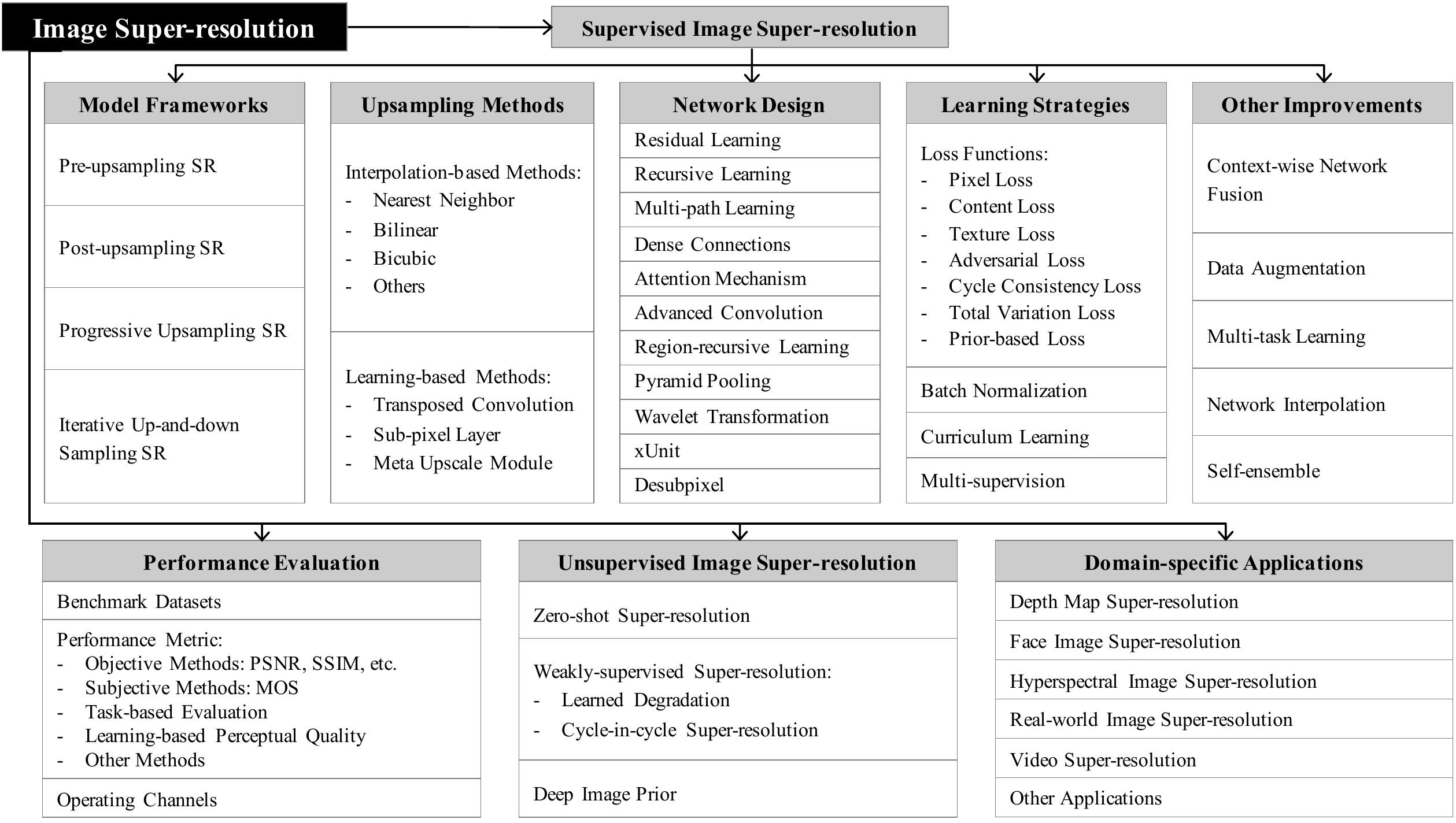}
    }
    \caption{
        Hierarchically-structured taxonomy of this survey.
    }
    \label{fig_sr_methodology}
\end{figure*}

\IEEEPARstart{I}{mage} super-resolution (SR), which refers to the process of recovering high-resolution (HR) images from low-resolution (LR) images, is an important class of image processing techniques in computer vision and image processing.
It enjoys a wide range of real-world applications, such as medical imaging \cite{TCJ2008SuperGreenspan,ICTSD2015SuperIsaac,CVPR2017SimultaneousHuang}, surveillance and security \cite{ESP2010SuperZhang,AMDO2016ConvolutionalRasti}), amongst others.  
Other than improving image perceptual quality, it also helps to improve other computer vision tasks \cite{WACV2016ImageDai,Arxiv2018TaskHaris,ICCV2017EnhancenetSajjadi,ECCV2018SODZhang}.  
In general, this problem is very challenging and inherently ill-posed since there are always multiple HR images corresponding to a single LR image.
In literature, a variety of classical SR methods have been proposed, including prediction-based methods \cite{TASSP1981CubicKeys,JAM1979LanczosDuchon,CVGIP1991ImprovingIrani}, edge-based methods \cite{TOG2011ImageFreedman,CVPR2008ImageSun}, statistical methods \cite{TPAMI2010SingleKim,TIP2010RobustXiong}, patch-based methods \cite{TOG2011ImageFreedman,CGA2002ExampleFreeman,CVPR2004SuperChang,ICCV2009SuperGlasner} and sparse representation methods \cite{CVPR2008ImageJianchao,TIP2010ImageYang}, etc.

With the rapid development of deep learning techniques in recent years, deep learning based SR models have been actively explored and often achieve the state-of-the-art performance on various benchmarks of SR.
A variety of deep learning methods have been applied to tackle SR tasks, ranging from the early Convolutional Neural Networks (CNN) based method (e.g., SRCNN \cite{ECCV2014LearningDong,TPAMI2016ImageDong}) to recent promising SR approaches using Generative Adversarial Nets (GAN) \cite{NIPS2014GenerativeGoodfellow} (e.g., SRGAN \cite{CVPR2017PhotoLedig}).
In general, the family of SR algorithms using deep learning techniques differ from each other in the following major aspects: different types of network architectures \cite{CVPR2016AccurateKim,CVPR2017DeepLai,ECCV2018FastAhn}, different types of loss functions \cite{ECCV2016PerceptualJohnson,ICCV2017EnhancenetSajjadi,CVPR2018SuperBulat}, different types of learning principles and strategies \cite{CVPRW2017EnhancedLim,ICCV2017EnhancenetSajjadi,CVPRW2018FullyWang}, etc.

In this paper, we give a comprehensive overview of recent advances in image super-resolution with deep learning.
Although there are some existing SR surveys in literature, our work differs in that we are focused in deep learning based SR techniques, while most of the earlier works \cite{ISPM2003SuperPark,MVA2014SuperNasrollahi,SIVP2011SurveyTian,IVC2006ImageVan} aim at surveying traditional SR algorithms or some studies mainly concentrate on providing quantitative evaluations based on full-reference metrics or human visual perception \cite{ECCV2014SingleYang,CEE2016PerformanceThapa}.
Unlike the existing surveys, this survey takes a unique deep learning based perspective to review the recent advances of SR techniques in a systematic and comprehensive manner.

The main contributions of this survey are three-fold:
\begin{enumerate}
    \item We give a comprehensive review of image super-resolution techniques based on deep learning, including problem settings, benchmark datasets, performance metrics, a family of SR methods with deep learning, domain-specific SR applications, etc.
    \item We provide a systematic overview of recent advances of deep learning based SR techniques in a hierarchical and structural manner, and summarize the advantages and limitations of each component for an effective SR solution.
    \item We discuss the challenges and open issues, and identify the new trends and future directions to provide an insightful guidance for the community.
\end{enumerate}

In the following sections, we will cover various aspects of recent advances in image super-resolution with deep learning.
Fig. \ref{fig_sr_methodology} shows the taxonomy of image SR to be covered in this survey in a hierarchically-structured way.
Section 2 gives the problem definition and reviews the mainstream datasets and evaluation metrics.
Section 3 analyzes main components of supervised SR modularly.
Section 4 gives a brief introduction to unsupervised SR methods.
Section 5 introduces some popular domain-specific SR applications, and Section 6 discusses future directions and open issues.

\section{Problem Setting and Terminology}
\label{sec_problem_setting}

\subsection{Problem Definitions}
\label{sec_problem_definitions}

Image super-resolution aims at recovering the corresponding HR images from the LR images.
Generally, the LR image $I_x$ is modeled as the output of the following degradation:
\begin{equation}
    I_x = \mathcal{D} (I_y ; \delta),
\end{equation}
where $\mathcal{D}$ denotes a degradation mapping function, $I_y$ is the corresponding HR image and $\delta$ is the parameters of the degradation process (e.g., the scaling factor or noise).
Generally, the degradation process (i.e., $\mathcal{D}$ and $\delta$) is unknown and only LR images are provided.
In this case, also known as blind SR, researchers are required to recover an HR approximation $\hat{I_y}$ of the ground truth HR image $I_y$ from the LR image $I_x$, following:
\begin{equation}
    \hat{I_y} = \mathcal{F} (I_x ; \theta),
\end{equation}
where $\mathcal{F}$ is the super-resolution model and $\theta$ denotes the parameters of $\mathcal{F}$.

Although the degradation process is unknown and can be affected by various factors (e.g., compression artifacts, anisotropic degradations, sensor noise and speckle noise), researchers are trying to model the degradation mapping.
Most works directly model the degradation as a single downsampling operation, as follows:
\begin{equation}
    \label{eq_degradation_impl_simple}
    \mathcal{D} (I_y ; \delta) = (I_y) \downarrow_s, \{s\} \subset \delta,
\end{equation}
where $\downarrow_s$ is a downsampling operation with the scaling factor $s$.
As a matter of fact, most datasets for generic SR are built based on this pattern, and the most commonly used downsampling operation is bicubic interpolation with anti-aliasing.
However, there are other works \cite{CVPR2018LearningZhang} modelling the degradation as a combination of several operations:
\begin{equation}
    \label{eq_degradation_impl_combine}
    \mathcal{D} (I_y ; \delta) = (I_y \otimes \kappa) \downarrow_s + n_\varsigma, \{\kappa,s,\varsigma\} \subset \delta,
\end{equation}
where $I_y \otimes \kappa$ represents the convolution between a blur kernel $\kappa$ and the HR image $I_y$, and $n_\varsigma$ is some additive white Gaussian noise with standard deviation $\varsigma$.
Compared to the naive definition of Eq. \ref{eq_degradation_impl_simple}, the combinative degradation pattern of Eq. \ref{eq_degradation_impl_combine} is closer to real-world cases and has been shown to be more beneficial for SR \cite{CVPR2018LearningZhang}.

To this end, the objective of SR is as follows:
\begin{equation}
    \hat{\theta} = \mathop{\arg \min}_{\theta} \mathcal{L} (\hat{I_y}, I_y) + \lambda \Phi (\theta),
\end{equation}
where $\mathcal{L} (\hat{I_y}, I_y)$ represents the loss function between the generated HR image $\hat{I_y}$ and the ground truth image $I_y$, $\Phi (\theta)$ is the regularization term and $\lambda$ is the tradeoff parameter.
Although the most popular loss function for SR is pixel-wise mean squared error (i.e., pixel loss), more powerful models tend to use a combination of multiple loss functions, which will be covered in Sec. \ref{sec_loss}.

\subsection{Datasets for Super-resolution}
\label{sec_datasets}

\begin{table*}[t]
    \renewcommand{\arraystretch}{1.3}
    \caption{List of public image datasets for super-resolution benchmarks.}
    \label{table_datasets}
    \centering
    \begin{tabular}{|l|c|c|c|c|l|}
        \hline
        Dataset & Amount & Avg. Resolution & Avg. Pixels & Format & Category Keywords \\
        \hline
        BSDS300      \cite{ICCV2001DatabaseMartin}    &   $300$ & $( 435,  367)$ &    $154,401$ & JPG & animal, building, food, landscape, people, plant, etc. \\
        BSDS500      \cite{TPAMI2011ContourArbelaez}  &   $500$ & $( 432,  370)$ &    $154,401$ & JPG & animal, building, food, landscape, people, plant, etc. \\
        DIV2K        \cite{CVPRW2017NTIREAgustsson}   &  $1000$ & $(1972, 1437)$ &  $2,793,250$ & PNG & environment, flora, fauna, handmade object, people, scenery, etc. \\
        General-100  \cite{ECCV2016AcceleratingDong}  &   $100$ & $( 435,  381)$ &    $181,108$ & BMP & animal, daily necessity, food, people, plant, texture, etc. \\
        L20          \cite{CVPR2016SevenTimofte}      &    $20$ & $(3843, 2870)$ & $11,577,492$ & PNG & animal, building, landscape, people, plant, etc. \\
        Manga109     \cite{MANPU2016Manga109Fujimoto} &   $109$ & $( 826, 1169)$ &    $966,011$ & PNG & manga volume \\
        OutdoorScene \cite{CVPR2018RecoveringWang}    & $10624$ & $( 553,  440)$ &    $249,593$ & PNG & animal, building, grass, mountain, plant, sky, water \\
        PIRM         \cite{ECCVW2018PIRMBlau}         &   $200$ & $( 617,  482)$ &    $292,021$ & PNG & environments, flora, natural scenery, objects, people, etc. \\
        Set5         \cite{BMVC2012LowBevilacqua}     &     $5$ & $( 313,  336)$ &    $113,491$ & PNG & baby, bird, butterfly, head, woman \\
        Set14        \cite{ICCS2010SingleZeyde}       &    $14$ & $( 492,  446)$ &    $230,203$ & PNG & humans, animals, insects, flowers, vegetables, comic, slides, etc. \\
        T91          \cite{TIP2010ImageYang}          &    $91$ & $( 264,  204)$ &     $58,853$ & PNG & car, flower, fruit, human face, etc. \\
        Urban100     \cite{CVPR2015SingleHuang}       &   $100$ & $( 984,  797)$ &    $774,314$ & PNG & architecture, city, structure, urban, etc. \\
        \hline
    \end{tabular}
\end{table*}

Today there are a variety of datasets available for image super-resolution, which greatly differ in image amounts, quality, resolution, and diversity, etc.
Some of them provide LR-HR image pairs, while others only provide HR images, in which case the LR images are typically obtained by \textit{imresize} function with default settings in MATLAB (i.e., bicubic interpolation with anti-aliasing).
In Table \ref{table_datasets} we list a number of image datasets commonly used by the SR community, and specifically indicate their amounts of HR images, average resolution, average numbers of pixels, image formats, and category keywords.

Besides these datasets, some datasets widely used for other vision tasks are also employed for SR, such as ImageNet \cite{CVPR2009ImagenetDeng}, MS-COCO \cite{ECCV2014MicrosoftLin}, VOC2012 \cite{IJCV2015PascalEveringham}, CelebA \cite{ICCV2015DeepLiu}.  
In addition, combining multiple datasets for training is also popular, such as combining T91 and BSDS300 \cite{CVPR2016AccurateKim,ICCV2017MemNetTai,CVPR2017DeepLai,CVPR2017ImageTai}, combining DIV2K and Flickr2K \cite{CVPRW2017EnhancedLim,CVPR2018DeepHaris}.

\subsection{Image Quality Assessment}
\label{sec_iqa}

Image quality refers to visual attributes of images and focuses on the perceptual assessments of viewers.
In general, image quality assessment (IQA) methods include subjective methods based on humans' perception (i.e., how realistic the image looks) and objective computational methods.  
The former is more in line with our need but often time-consuming and expensive, thus the latter is currently the mainstream.
However, these methods aren't necessarily consistent between each other, because objective methods are often unable to capture the human visual perception very accurately, which may lead to large difference in IQA results \cite{TIP2004ImageWang,CVPR2017PhotoLedig}.

In addition, the objective IQA methods are further divided into three types \cite{TIP2004ImageWang}: full-reference methods performing assessment using reference images, reduced-reference methods based on comparisons of extracted features, and no-reference methods (i.e., blind IQA) without any reference images.
Next we'll introduce several most commonly used IQA methods covering both subjective methods and objective methods.

\subsubsection{Peak Signal-to-Noise Ratio}
\label{sec_iqa_psnr}

Peak signal-to-noise ratio (PSNR) is one of the most popular reconstruction quality measurement of lossy transformation (e.g., image compression, image inpainting).
For image super-resolution, PSNR is defined via the maximum pixel value (denoted as $L$) and the mean squared error (MSE) between images.
Given the ground truth image $I$ with $N$ pixels and the reconstruction $\hat{I}$, the PSNR between $I$ and $\hat{I}$ are defined as follows:
\begin{align}
    \label{equ:mse}
    {\rm PSNR} &= 10 \cdot \log_{10} (\frac {L^2} {\frac{1}{N} \sum_{i=1}^{N} (I(i) - \hat{I}(i)) ^2}) ,
\end{align}
where $L$ equals to $255$ in general cases using 8-bit representations.
Since the PSNR is only related to the pixel-level MSE, only caring about the differences between corresponding pixels instead of visual perception, it often leads to poor performance in representing the reconstruction quality in real scenes, where we're usually more concerned with human perceptions.
However, due to the necessity to compare with literature works and the lack of completely accurate perceptual metrics, PSNR is still currently the most widely used evaluation criteria for SR models.

\subsubsection{Structural Similarity}
\label{sec_iqa_ssim}

Considering that the human visual system (HVS) is highly adapted to extract image structures \cite{ICASSP2002WhyWang}, the structural similarity index (SSIM) \cite{TIP2004ImageWang} is proposed for measuring the structural similarity between images, based on independent comparisons in terms of luminance, contrast, and structures.
For an image $I$ with $N$ pixels, the luminance $\mu_I$ and contrast $\sigma_I$ are estimated as the mean and standard deviation of the image intensity, respectively, i.e., $\mu_I = \frac{1}{N} \sum_{i=1}^N I(i)$ and $\sigma_I = (\frac{1}{N-1} \sum_{i=1}^N (I(i) - \mu_I)^2)^\frac{1}{2}$,
where $I(i)$ represents the intensity of the $i$-th pixel of image $I$.
And the comparisons on luminance and contrast, denoted as $\mathcal{C}_l(I, \hat{I})$ and $\mathcal{C}_c(I, \hat{I})$ respectively, are given by:
\begin{align}
    \mathcal{C}_l(I, \hat{I}) &= \frac {2 \mu_I \mu_{\hat{I}} + C_1} {\mu_I^2 + \mu_{\hat{I}}^2 + C_1}, \\
    \mathcal{C}_c(I, \hat{I}) &= \frac {2 \sigma_I \sigma_{\hat{I}} + C_2} {\sigma_I^2 + \sigma_{\hat{I}}^2 + C_2},
\end{align}
where $C_1 = (k_1 L)^2$ and $C_2 = (k_2 L)^2$ are constants for avoiding instability, $k_1 \ll 1$ and $k_2 \ll 1$.

Besides, the image structure is represented by the normalized pixel values (i.e., $(I - \mu_I) / \sigma_I$), whose correlations (i.e., inner product) measure the structural similarity, equivalent to the correlation coefficient between $I$ and $\hat{I}$.
Thus the structure comparison function $\mathcal{C}_s(I, \hat{I})$ is defined as:
\begin{align}
    \sigma_{I\hat{I}} &= \frac{1}{N-1} \sum_{i=1}^{N} (I(i) - \mu_I) (\hat{I}(i) - \mu_{\hat{I}}), \\
    \mathcal{C}_s(I, \hat{I}) &= \frac {\sigma_{I \hat{I}} + C_3} {\sigma_I \sigma_{\hat{I}} + C_3},
\end{align}
where $\sigma_{I,\hat{I}}$ is the covariance between $I$ and $\hat{I}$, and $C_3$ is a constant for stability.

Finally, the SSIM is given by:
\begin{equation}
    {\rm SSIM}(I, \hat{I}) = [\mathcal{C}_l(I, \hat{I})]^\alpha
                             [\mathcal{C}_c(I, \hat{I})]^\beta
                             [\mathcal{C}_s(I, \hat{I})]^\gamma,
\end{equation}
where $\alpha$, $\beta$, $\gamma$ are control parameters for adjusting the relative importance.



Since the SSIM evaluates the reconstruction quality from the perspective of the HVS, it better meets the requirements of perceptual assessment \cite{TIP2006StatisticalSheikh,ISPM2009MeanWang}, and is also widely used.

\subsubsection{Mean Opinion Score}
\label{sec_iqa_mos}

Mean opinion score (MOS) testing is a commonly used subjective IQA method, where human raters are asked to assign perceptual quality scores to tested images.
Typically, the scores are from $1$ (bad) to $5$ (good).
And the final MOS is calculated as the arithmetic mean over all ratings.

Although the MOS testing seems a faithful IQA method, it has some inherent defects, such as non-linearly perceived scales, biases and variance of rating criteria.
In reality, there are some SR models performing poorly in common IQA metrics (e.g., PSNR) but far exceeding others in terms of perceptual quality, in which case the MOS testing is the most reliable IQA method for accurately measuring the perceptual quality \cite{ICCV2015DeepWang,CVPR2017PhotoLedig,ICCV2017EnhancenetSajjadi,ICCV2017LearningXu,ICCV2017PixelDahl,CVPR2018RecoveringWang,TPAMI2018FastLai}.

\subsubsection{Learning-based Perceptual Quality}
\label{sec_iqa_learning_perceptual}

In order to better assess the image perceptual quality while reducing manual intervention, researchers try to assess the perceptual quality by learning on large datasets.
Specifically, Ma \textit{et al.} \cite{CVIU2017LearningMa} and Talebi \textit{et al.} \cite{TIP2018NimaTalebi} propose no-reference Ma and NIMA, respectively, which are learned from visual perceptual scores and directly predict the quality scores without ground-truth images.
In contrast, Kim \textit{et al.} \cite{CVPR2017DeepKim} propose DeepQA, which predicts visual similarity of images by training on triplets of distorted images, objective error maps, and subjective scores.
And Zhang \textit{et al.} \cite{CVPR2018TheZhang} collect a large-scale perceptual similarity dataset, evaluate the perceptual image patch similarity (LPIPS) according to the difference in deep features by trained deep networks, and show that the deep features learned by CNNs model perceptual similarity much better than measures without CNNs.

Although these methods exhibit better performance on capturing human visual perception, what kind of perceptual quality we need (e.g., more realistic images, or consistent identity to the original image) remains a question to be explored, thus the objective IQA methods (e.g., PSNR, SSIM) are still the mainstreams currently.

\subsubsection{Task-based Evaluation}
\label{sec_iqa_task}

According to the fact that SR models can often help other vision tasks \cite{WACV2016ImageDai,Arxiv2018TaskHaris,ICCV2017EnhancenetSajjadi,ECCV2018SODZhang}, evaluating reconstruction performance by means of other tasks is another effective way.  
Specifically, researchers feed the original and the reconstructed HR images into trained models, and evaluate the reconstruction quality by comparing the impacts on the prediction performance.
The vision tasks used for evaluation include object recognition \cite{ICCV2017EnhancenetSajjadi,ECCV2018ImageZhang}, face recognition \cite{JVCIR2012EvaluationFookes,ECCV2018SuperZhang}, face alignment and parsing \cite{CVPR2018SuperBulat,CVPR2018FSRNetChen}, etc.


\subsubsection{Other IQA Methods}

In addition to above IQA methods, there are other less popular SR metrics.
The multi-scale structural similarity (MS-SSIM) \cite{ACSSC2003MultiWang} supplies more flexibility than single-scale SSIM in incorporating the variations of viewing conditions.
The feature similarity (FSIM) \cite{TIP2011FSIMZhang} extracts feature points of human interest based on phase congruency and image gradient magnitude to evaluate image quality.
The Natural Image Quality Evaluator (NIQE) \cite{SPL2013MakingMittal} makes use of measurable deviations from statistical regularities observed in natural images, without exposure to distorted images.

Recently, Blau \textit{et al.} \cite{CVPR2018TheBlau} prove mathematically that distortion (e.g., PSNR, SSIM) and perceptual quality (e.g., MOS) are at odds with each other, and show that as the distortion decreases, the perceptual quality must be worse.
Thus how to accurately measure the SR quality is still an urgent problem to be solved.

\subsection{Operating Channels}

In addition to the commonly used RGB color space, the YCbCr color space is also widely used for SR.  
In this space, images are represented by Y, Cb, Cr channels, denoting the luminance, blue-difference and red-difference chroma components, respectively.
Although currently there is no accepted best practice for performing or evaluating super-resolution on which space, earlier models favor operating on the Y channel of YCbCr space \cite{CVPR2016AccurateKim,ECCV2016AcceleratingDong,NIPS2016ImageMao,ICCV2017ImageTong}, while more recent models tend to operate on RGB channels \cite{CVPRW2017EnhancedLim,CVPR2018DeepHaris,ECCV2018FastAhn,ECCV2018ImageZhang}.
It is worth noting that operating (training or evaluation) on different color spaces or channels can make the evaluation results differ greatly (up to $4$ dB) \cite{TPAMI2016ImageDong}.

\subsection{Super-resolution Challenges}

In this section, we will briefly introduce two most popular challenges for image SR, NTIRE \cite{CVPRW2017NTIRETimofte} and PIRM \cite{ECCVW2018PIRMBlau,ECCVW2018PIRMIgnatov}.

\textbf{NTIRE Challenge.}
The New Trends in Image Restoration and Enhancement (NTIRE) challenge \cite{CVPRW2017NTIRETimofte} is in conjunction with CVPR and includes multiple tasks like SR, denoising and colorization.
For image SR, the NTIRE challenge is built on the DIV2K \cite{CVPRW2017NTIREAgustsson} dataset and consists of bicubic downscaling tracks and blind tracks with realistic unknown degradation.
These tracks differs in degradations and scaling factors, and aim to promote the SR research under both ideal conditions and real-world adverse situations.

\textbf{PIRM Challenge.}
The Perceptual Image Restoration and Manipulation (PIRM) challenges are in conjunction with ECCV and also includes multiple tasks.
In contrast to NTIRE, one sub-challenge \cite{ECCVW2018PIRMBlau} of PIRM focuses on the tradeoff between generation accuracy and perceptual quality, and the other \cite{ECCVW2018PIRMIgnatov} focuses on SR on smartphones.
As is well-known \cite{CVPR2018TheBlau}, the models target for distortion frequently produce visually unpleasing results, while the models target for perceptual quality performs poorly on information fidelity.
Specifically, the PIRM divided the perception-distortion plane into three regions according to thresholds on root mean squared error (RMSE).
In each region, the winning algorithm is the one that achieves the best perceptual quality \cite{CVPR2018TheBlau}, evaluated by NIQE \cite{SPL2013MakingMittal} and Ma \cite{CVIU2017LearningMa}.
While in the other sub-challenge \cite{ECCVW2018PIRMIgnatov}, SR on smartphones, participants are asked to perform SR with limited smartphone hardwares (including CPU, GPU, RAM, etc.), and the evaluation metrics include PSNR, MS-SSIM and MOS testing.
In this way, PIRM encourages advanced research on the perception-distortion tradeoff, and also drives lightweight and efficient image enhancement on smartphones.

\section{Supervised Super-resolution}

Nowadays researchers have proposed a variety of super-resolution models with deep learning.
These models focus on supervised SR, i.e., trained with both LR images and corresponding HR images.
Although the differences between these models are very large, they are essentially some combinations of a set of components such as model frameworks, upsampling methods, network design, and learning strategies.
From this perspective, researchers combine these components to build an integrated SR model for fitting specific purposes.
In this section, we concentrate on modularly analyzing the fundamental components (as Fig. \ref{fig_sr_methodology} shows) instead of introducing each model in isolation, and summarizing their advantages and limitations.

\subsection{Super-resolution Frameworks}
\label{sec_sr_frameworks}

\begin{figure}[!t]
    \centering
    \subfloat[Pre-upsampling SR]{\includegraphics[width=0.4\textwidth]{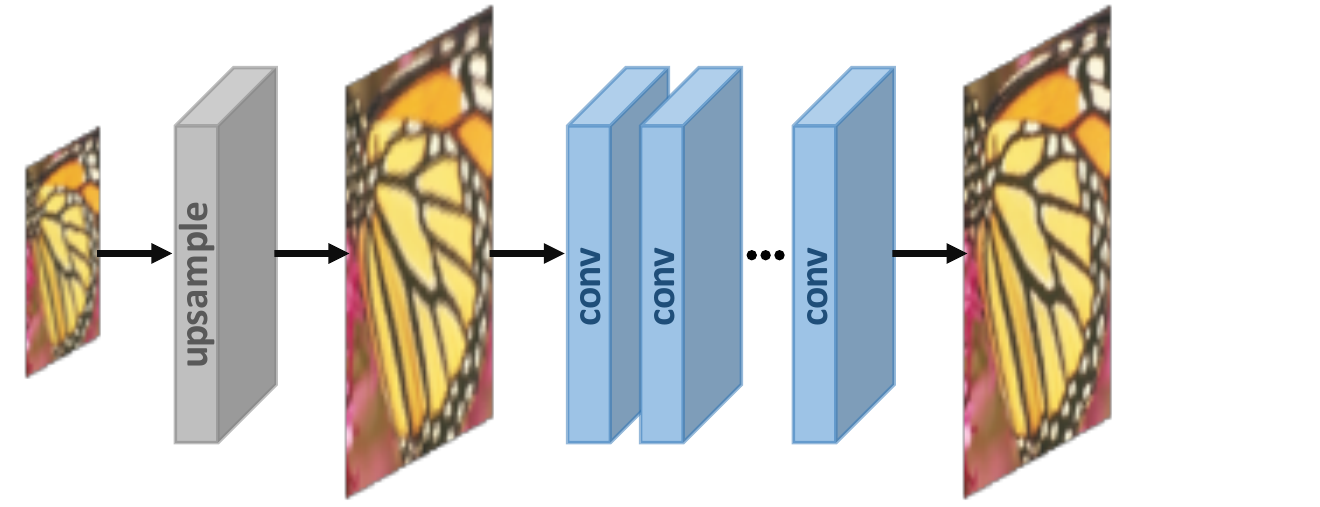}%
    \label{fig_framework_pre}} \\
    \subfloat[Post-upsampling SR]{\includegraphics[width=0.4\textwidth]{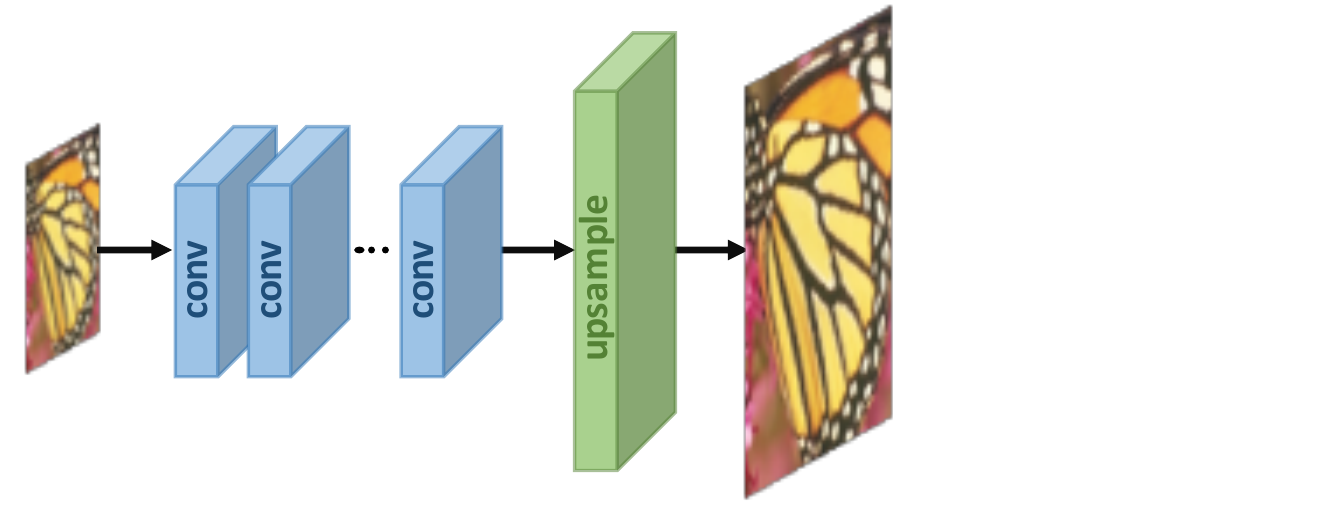}%
    \label{fig_framework_post}} \\
    \subfloat[Progressive upsampling SR]{\includegraphics[width=0.4\textwidth]{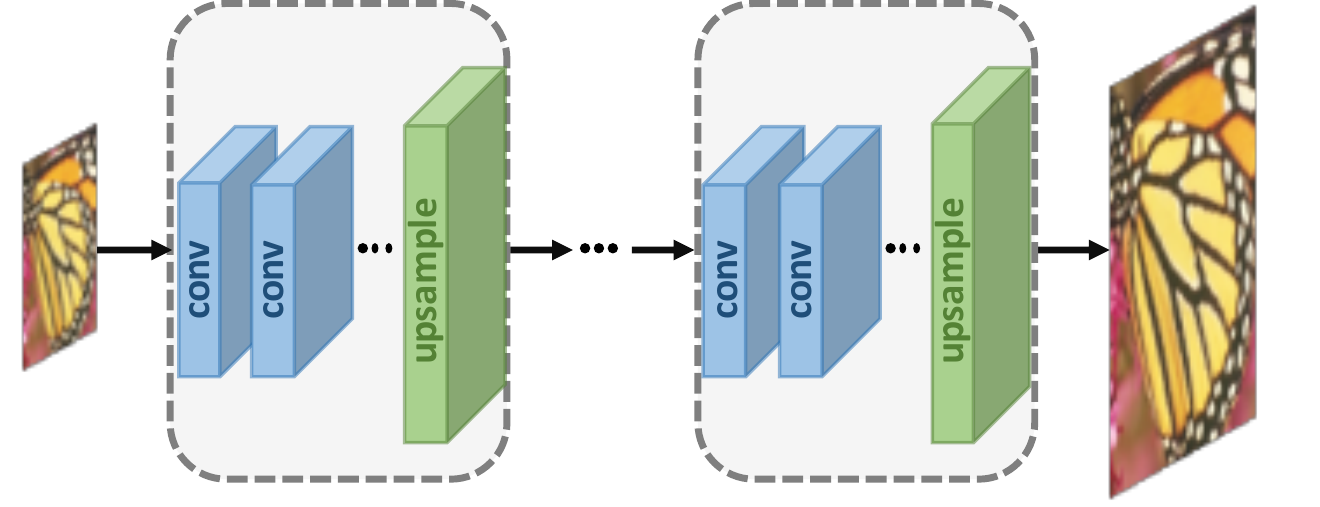}%
    \label{fig_framework_progressive}} \\
    \subfloat[Iterative up-and-down Sampling SR]{\includegraphics[width=0.4\textwidth]{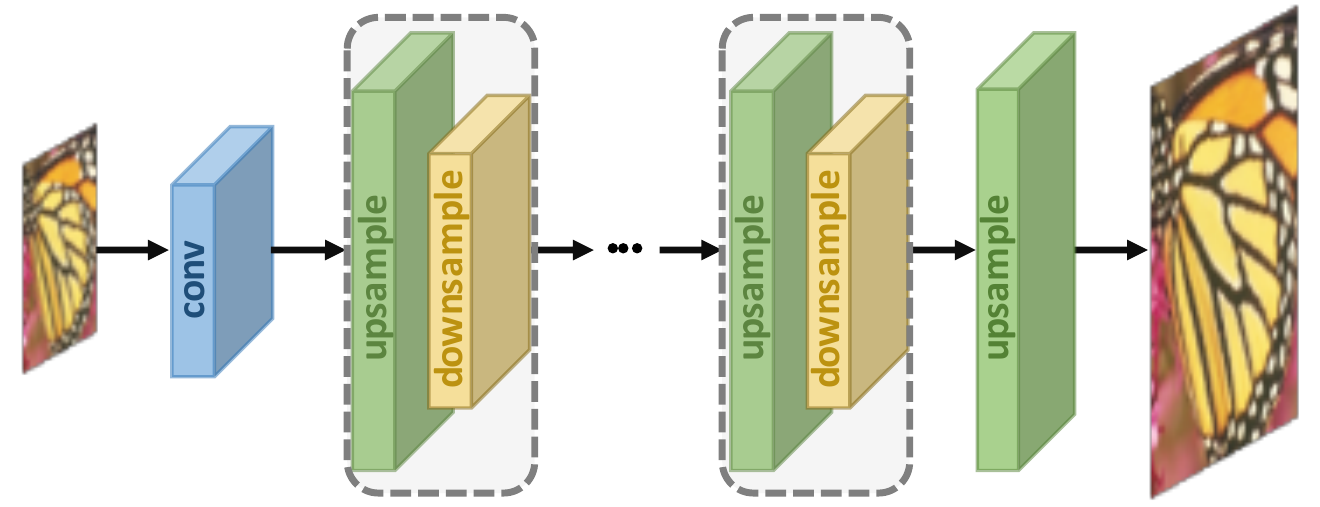}%
    \label{fig_framework_up_down}} \\
    \caption{
        Super-resolution model frameworks based on deep learning.
        The cube size represents the output size.
        The \textcolor{Gray}{gray} ones denote predefined upsampling, while the \textcolor{LimeGreen}{green}, \textcolor{Dandelion}{yellow} and \textcolor{CornflowerBlue}{blue} ones indicate learnable upsampling, downsampling and convolutional layers, respectively.
        And the blocks enclosed by dashed boxes represent stackable modules.
    }
    \label{fig_framework}
\end{figure}

Since image super-resolution is an ill-posed problem, how to perform upsampling (i.e., generating HR output from LR input) is the key problem.
Although the architectures of existing models vary widely, they can be attributed to four model frameworks (as Fig. \ref{fig_framework} shows), based on the employed upsampling operations and their locations in the model.

\subsubsection{Pre-upsampling Super-resolution}
\label{sec_framework_pre}

On account of the difficulty of directly learning the mapping from low-dimensional space to high-dimensional space, utilizing traditional upsampling algorithms to obtain higher-resolution images and then refining them using deep neural networks is a straightforward solution.
Thus Dong \textit{et al.} \cite{ECCV2014LearningDong,TPAMI2016ImageDong} firstly adopt the pre-upsampling SR framework (as Fig. \ref{fig_framework_pre} shows) and propose SRCNN to learn an end-to-end mapping from interpolated LR images to HR images.
Specifically, the LR images are upsampled to coarse HR images with the desired size using traditional methods (e.g., bicubic interpolation), then deep CNNs are applied on these images for reconstructing high-quality details.

Since the most difficult upsampling operation has been completed, CNNs only need to refine the coarse images, which significantly reduces the learning difficulty.
In addition, these models can take interpolated images with arbitrary sizes and scaling factors as input, and give refined results with comparable performance to single-scale SR models \cite{CVPR2016AccurateKim}.
Thus it has gradually become one of the most popular frameworks \cite{CVPR2016DeeplyKim,CVPR2017ImageTai,ICCV2017MemNetTai,CVPR2018ZeroShocher}, and the main differences between these models are the posterior model design (Sec. \ref{sec_network_design}) and learning strategies (Sec. \ref{sec_learning_strategies}).
However, the predefined upsampling often introduce side effects (e.g., noise amplification and blurring), and since most operations are performed in high-dimensional space, the cost of time and space is much higher than other frameworks \cite{CVPR2016RealShi,ECCV2016AcceleratingDong}.

\subsubsection{Post-upsampling Super-resolution}
\label{sec_framework_post}

In order to improve the computational efficiency and make full use of deep learning technology to increase resolution automatically, researchers propose to perform most computation in low-dimensional space by replacing the predefined upsampling with end-to-end learnable layers integrated at the end of the models.
In the pioneer works \cite{CVPR2016RealShi,ECCV2016AcceleratingDong} of this framework, namely post-upsampling SR as Fig. \ref{fig_framework_post} shows, the LR input images are fed into deep CNNs without increasing resolution, and end-to-end learnable upsampling layers are applied at the end of the network.

Since the feature extraction process with huge computational cost only occurs in low-dimensional space and the resolution increases only at the end, the computation and spatial complexity are much reduced.
Therefore, this framework also has become one of the most mainstream frameworks \cite{CVPR2017PhotoLedig,CVPRW2017EnhancedLim,ICCV2017ImageTong,CVPR2018ImageHan}.
These models differ mainly in the learnable upsampling layers (Sec. \ref{sec_upsampling_methods}), anterior CNN structures (Sec. \ref{sec_network_design}) and learning strategies ({Sec. \ref{sec_learning_strategies}), etc.

\subsubsection{Progressive Upsampling Super-resolution}
\label{sec_framework_progressive}

Although post-upsampling SR framework has immensely reduced the computational cost, it still has some shortcomings.
On the one hand, the upsampling is performed in only one step, which greatly increases the learning difficulty for large scaling factors (e.g., 4, 8).
On the other hand, each scaling factor requires training an individual SR model, which cannot cope with the need for multi-scale SR.
To address these drawbacks, a progressive upsampling framework is adopted by Laplacian pyramid SR network (LapSRN) \cite{CVPR2017DeepLai}, as Fig. \ref{fig_framework_progressive} shows.
Specifically, the models under this framework are based on a cascade of CNNs and progressively reconstruct higher-resolution images.
At each stage, the images are upsampled to higher resolution and refined by CNNs.
Other works such as MS-LapSRN \cite{TPAMI2018FastLai} and progressive SR (ProSR) \cite{CVPRW2018FullyWang} also adopt this framework and achieve relatively high performance.
In contrast to the LapSRN and MS-LapSRN using the intermediate reconstructed images as the ``base images'' for subsequent modules, the ProSR keeps the main information stream and reconstructs intermediate-resolution images by individual heads.

By decomposing a difficult task into simple tasks, the models under this framework greatly reduce the learning difficulty, especially with large factors, and also cope with the multi-scale SR without introducing overmuch spacial and temporal cost.
In addition, some specific learning strategies such as curriculum learning (Sec. \ref{sec_curriculum_learning}) and multi-supervision (Sec. \ref{sec_multi_supervision}) can be directly integrated to further reduce learning difficulty and improve final performance.
However, these models also encounter some problems, such as the complicated model designing for multiple stages and the training stability, and more modelling guidance and more advanced training strategies are needed.

\subsubsection{Iterative Up-and-down Sampling Super-resolution}
\label{sec_framework_up_down}

In order to better capture the mutual dependency of LR-HR image pairs, an efficient iterative procedure named back-projection \cite{CVGIP1991ImprovingIrani} is incorporated into SR \cite{CVPR2016SevenTimofte}.
This SR framework, namely iterative up-and-down sampling SR (as Fig. \ref{fig_framework_up_down} shows), tries to iteratively apply back-projection refinement, i.e., computing the reconstruction error then fusing it back to tune the HR image intensity.
Specifically, Haris \textit{et al.} \cite{CVPR2018DeepHaris} exploit iterative up-and-down sampling layers and propose DBPN, which connects upsampling and downsampling layers alternately and reconstructs the final HR result using all of the intermediately reconstructions.
Similarly, the SRFBN \cite{CVPR2019FeedbackLi} employs a iterative up-and-down sampling feedback block with more dense skip connections and learns better representations.
And the RBPN \cite{CVPR2019RecurrentHaris} for video super-resolution extracts context from continuous video frames and combines these context to produce recurrent output frames by a back-projection module.

The models under this framework can better mine the deep relationships between LR-HR image pairs and thus provide higher-quality reconstruction results.
Nevertheless, the design criteria of the back-projection modules are still unclear.
Since this mechanism has just been introduced into deep learning-based SR, the framework has great potential and needs further exploration.

\subsection{Upsampling Methods}
\label{sec_upsampling_methods}

In addition to the upsampling positions in the model, how to perform upsampling is of great importance.
Although there has been various traditional upsampling methods \cite{CVPR2008ImageJianchao,TIP2010ImageYang,ACCV2014AdjustedTimofte,CVPR2015FastSchulter}, making use of CNNs to learn end-to-end upsampling has gradually become a trend.  
In this section, we'll introduce some traditional interpolation-based algorithms and deep learning-based upsampling layers.

\subsubsection{Interpolation-based Upsampling}

\begin{figure}[!t]
    \centering
    \subfloat[Starting]{\includegraphics[width=0.1\textwidth]{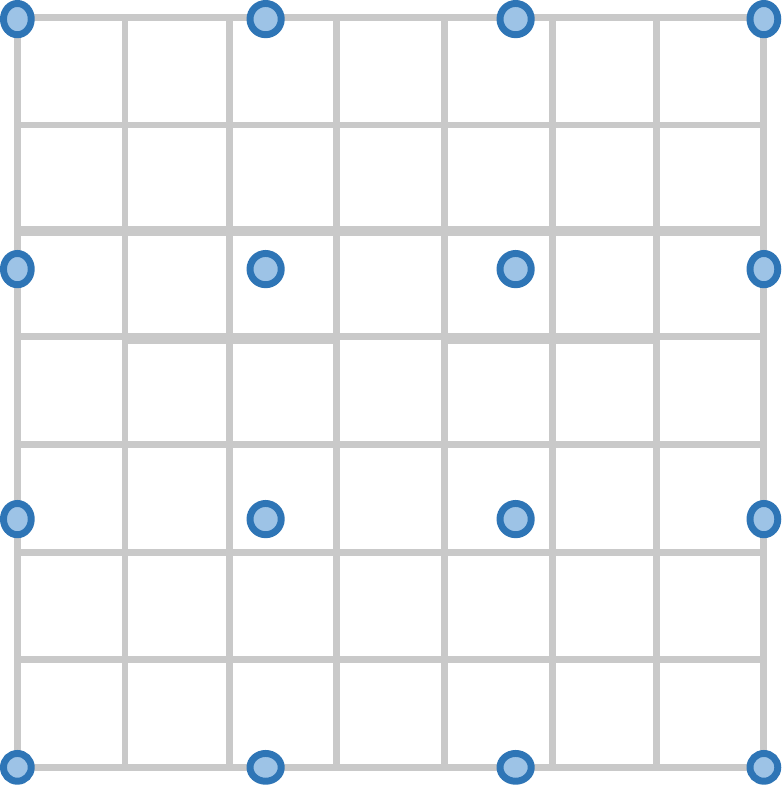}%
    \label{fig_interpolation_0}}
    \hfil
    \subfloat[Step 1]{\includegraphics[width=0.1\textwidth]{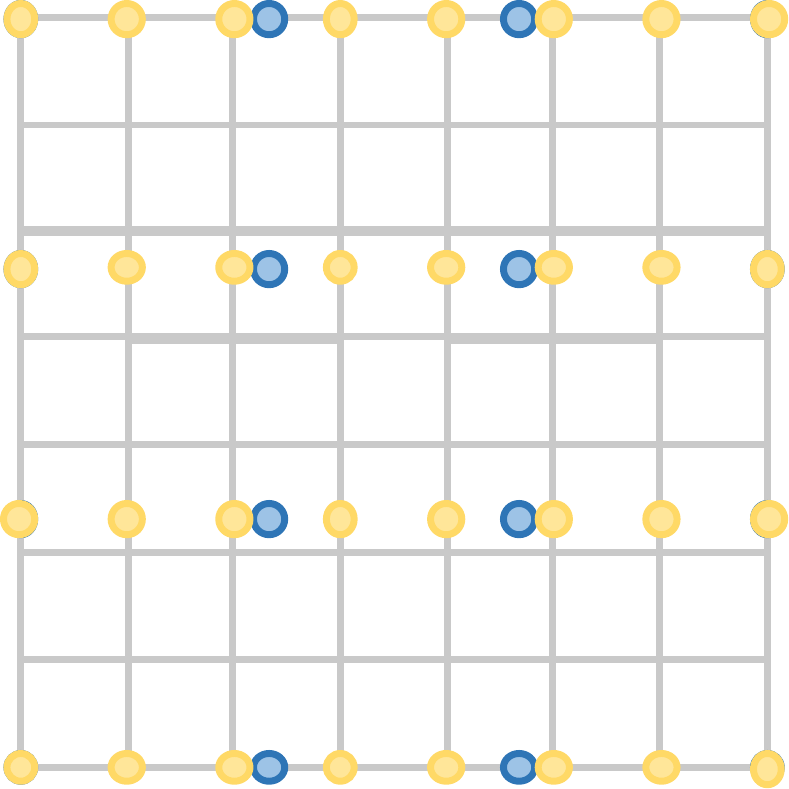}%
    \label{fig_interpolation_1}}
    \hfil
    \subfloat[Step 2]{\includegraphics[width=0.1\textwidth]{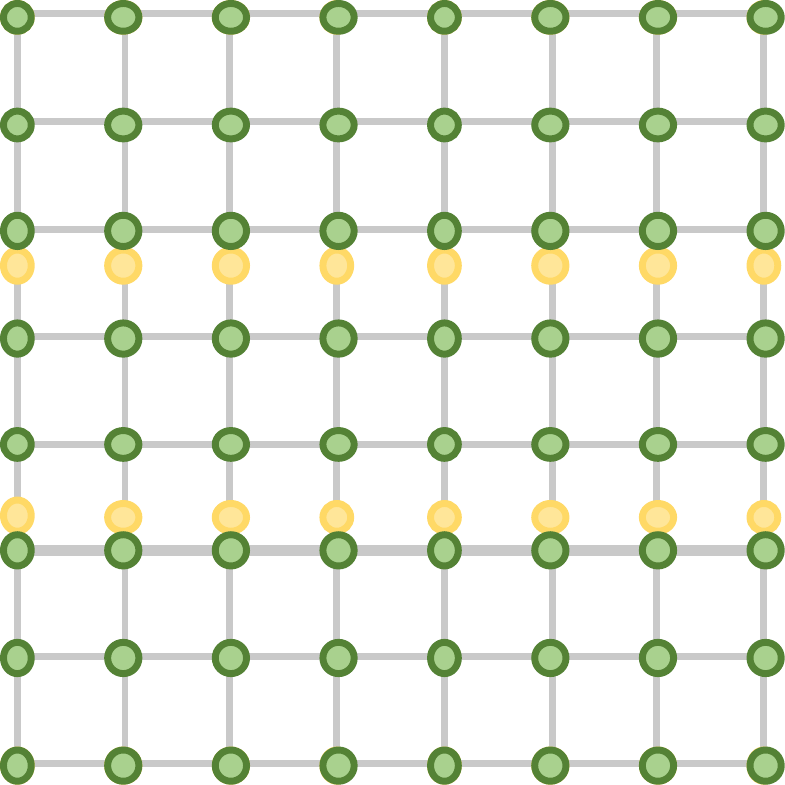}%
    \label{fig_interpolation_2}}
    \hfil
    \subfloat[End]{\includegraphics[width=0.1\textwidth]{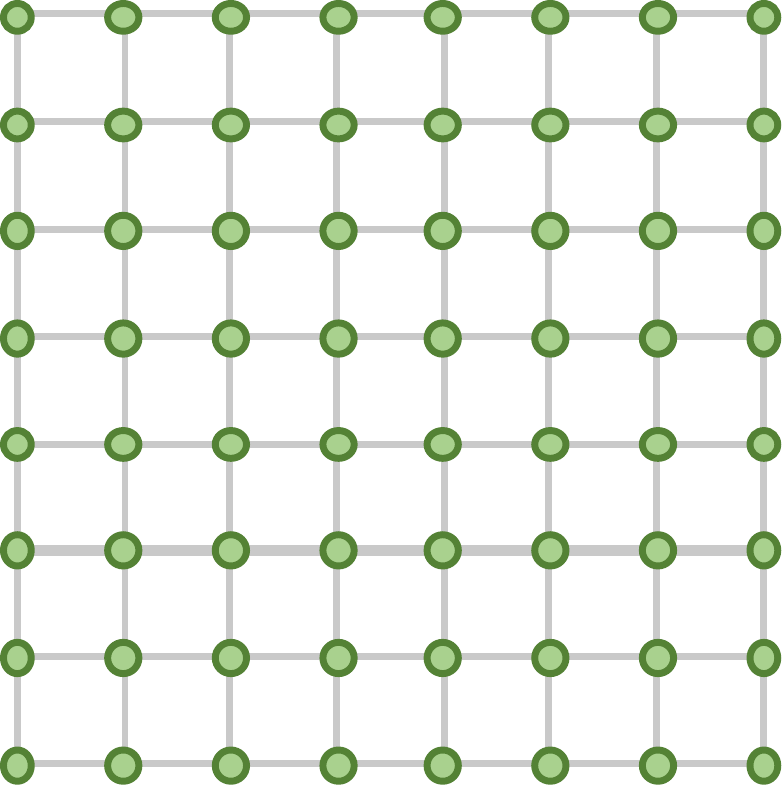}%
    \label{fig_interpolation_3}}
    \caption{
        Interpolation-based upsampling methods.
        The \textcolor{Gray}{gray} board denotes the coordinates of pixels, and the \textcolor{CornflowerBlue}{blue}, \textcolor{Dandelion}{yellow} and \textcolor{LimeGreen}{green} points represent the initial, intermediate and output pixels, respectively.
    }
    \label{fig_interpolation}
\end{figure}

Image interpolation, a.k.a. image scaling, refers to resizing digital images and is widely used by image-related applications.
The traditional interpolation methods include nearest-neighbor interpolation, bilinear and bicubic interpolation, Sinc and Lanczos resampling, etc.
Since these methods are interpretable and easy to implement, some of them are still widely used in CNN-based SR models.

\textbf{Nearest-neighbor Interpolation.}
The nearest-neighbor interpolation is a simple and intuitive algorithm.
It selects the value of the nearest pixel for each position to be interpolated regardless of any other pixels.
Thus this method is very fast but usually produces blocky results of low quality.

\textbf{Bilinear Interpolation.}
The bilinear interpolation (BLI) first performs linear interpolation on one axis of the image and then performs on the other axis, as Fig. \ref{fig_interpolation} shows.
Since it results in a quadratic interpolation with a receptive field sized $2 \times 2$, it shows much better performance than nearest-neighbor interpolation while keeping relatively fast speed.

\textbf{Bicubic Interpolation.}
Similarly, the bicubic interpolation (BCI) \cite{TASSP1981CubicKeys} performs cubic interpolation on each of the two axes, as Fig. \ref{fig_interpolation} shows.
Compared to BLI, the BCI takes $4 \times 4$ pixels into account, and results in smoother results with fewer artifacts but much lower speed.
In fact, the BCI with anti-aliasing is the mainstream method for building SR datasets (i.e., degrading HR images to LR images), and is also widely used in pre-upsampling SR framework (Sec. \ref{sec_framework_pre}).

As a matter of fact, the interpolation-based upsampling methods improve the image resolution only based on its own image signals, without bringing any more information.
Instead, they often introduce some side effects, such as computational complexity, noise amplification, blurring results.
Therefore, the current trend is to replace the interpolation-based methods with learnable upsampling layers.

\subsubsection{Learning-based Upsampling}
\label{sec_learning_based_upsampling}

\begin{figure}[!t]
    \centering
    \subfloat[Starting]{\includegraphics[height=60pt]{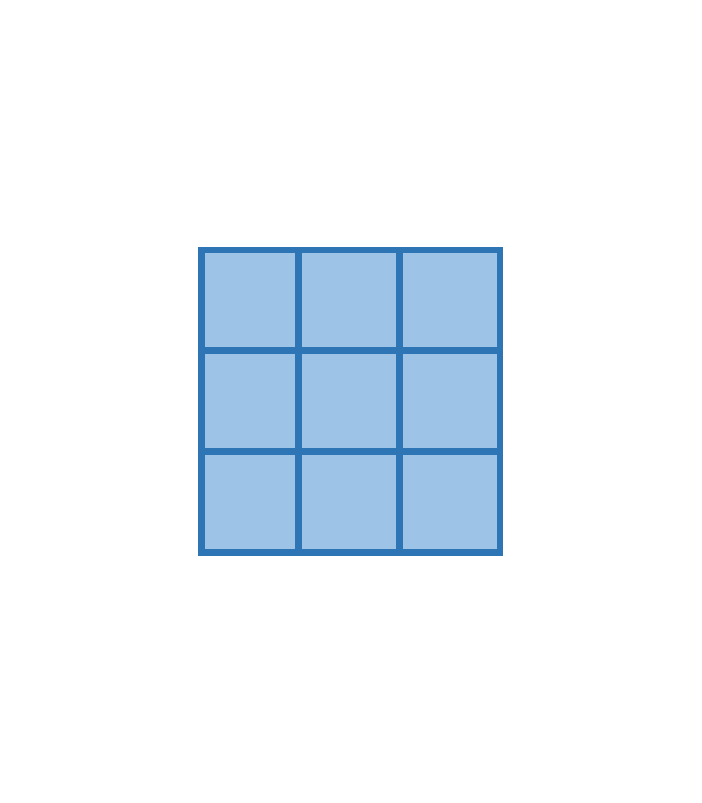}%
    \label{fig_deconv_0}}
    \hfil
    \subfloat[Expanding]{\includegraphics[height=60pt]{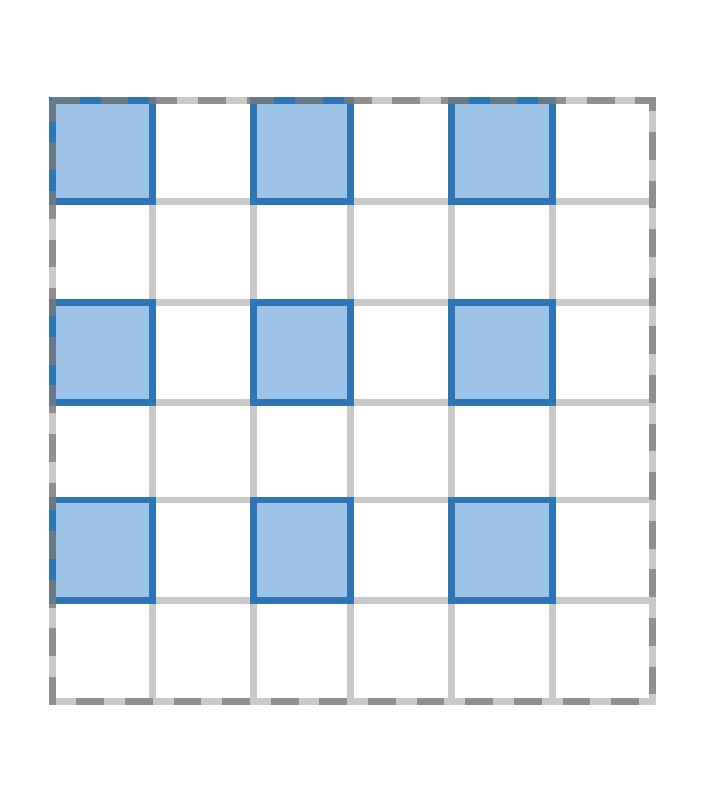}%
    \label{fig_deconv_1}}
    \hfil
    \subfloat[Convolution]{\includegraphics[height=60pt]{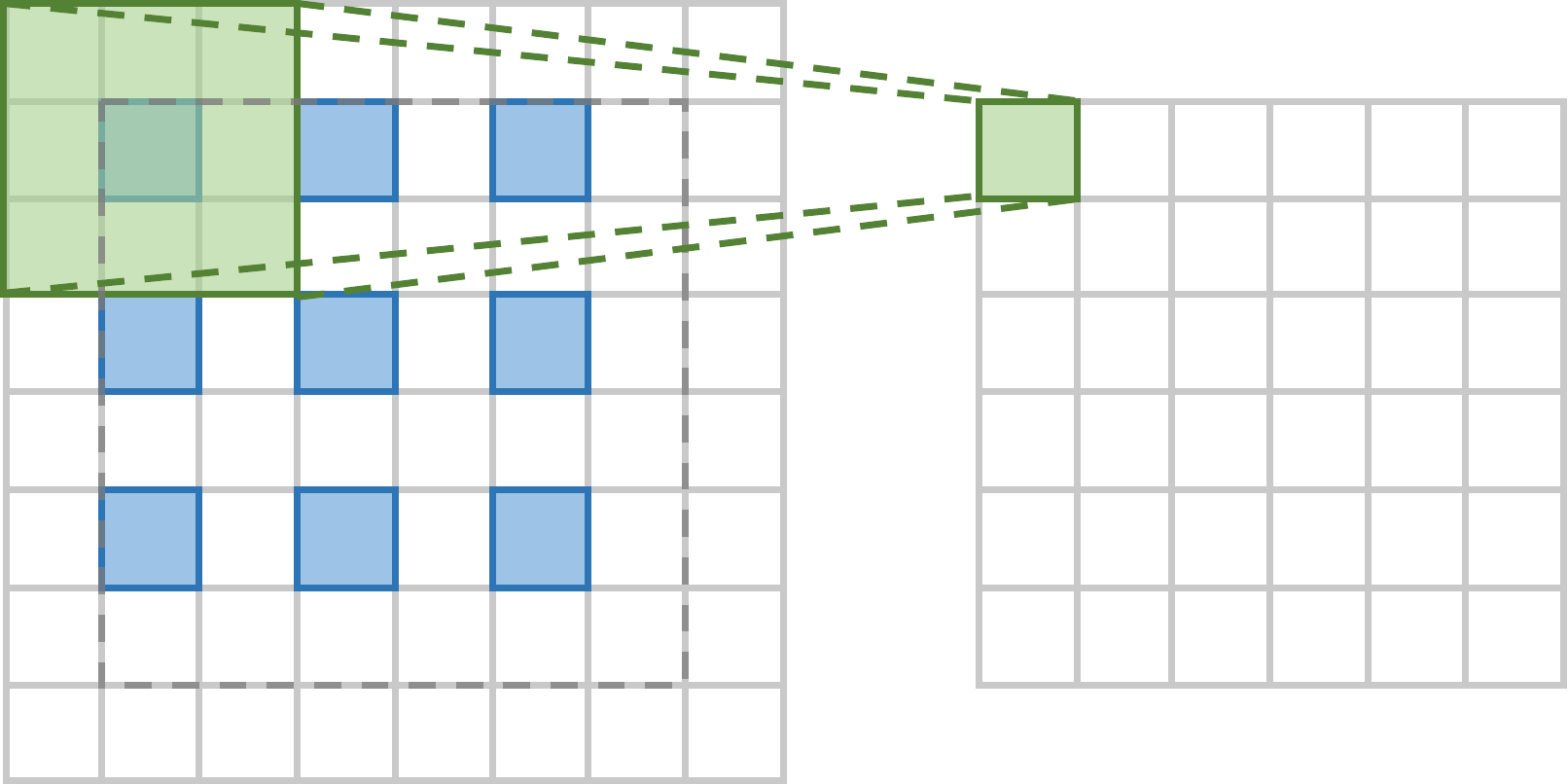}%
    \label{fig_deconv_2}}
    \caption{
        Transposed convolution layer.
        The \textcolor{CornflowerBlue}{blue} boxes denote the input, and the \textcolor{LimeGreen}{green} boxes indicate the kernel and the convolution output.
    }
    \label{fig_deconv}
\end{figure}

\begin{figure}[!t]
    \centering
    \subfloat[Starting]{\includegraphics[height=45pt]{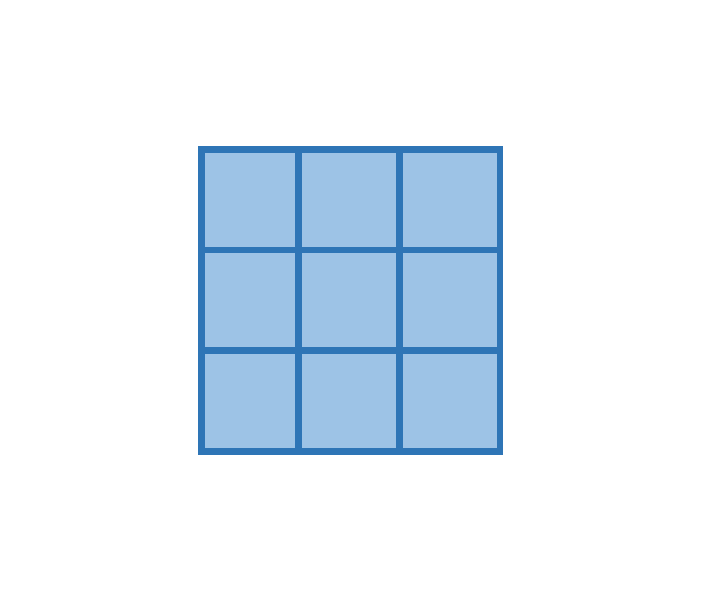}%
    \label{fig_sub_pixel_0}}
    \hfil
    \subfloat[Convolution]{\includegraphics[height=45pt]{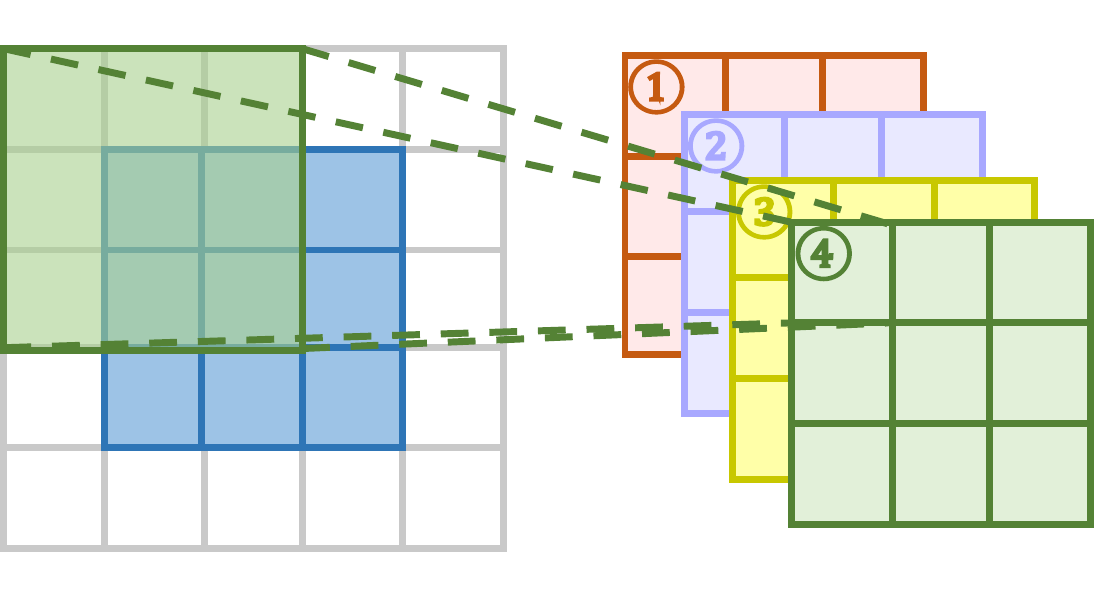}%
    \label{fig_sub_pixel_1}}
    \hfil
    \subfloat[Reshaping]{\includegraphics[height=45pt]{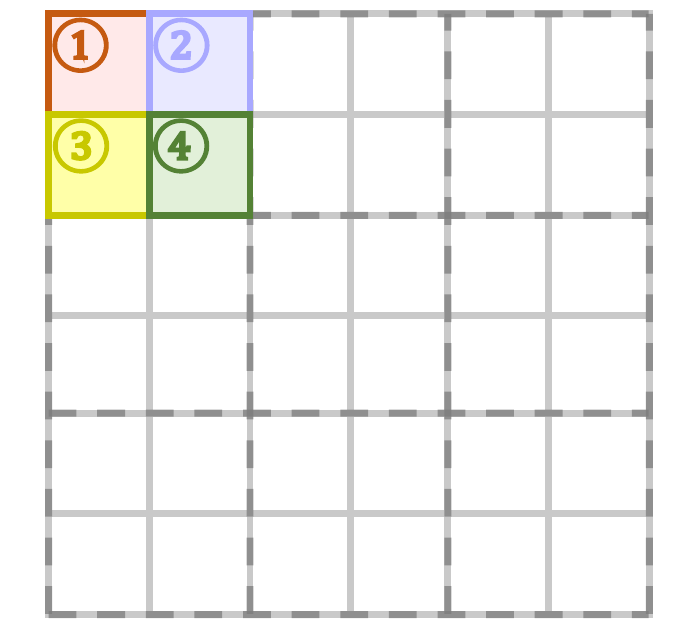}%
    \label{fig_sub_pixel_2}}
    \caption{
        Sub-pixel layer.
        The \textcolor{CornflowerBlue}{blue} boxes denote the input, and the boxes with other colors indicate different convolution operations and different output feature maps.
    }
    \label{fig_sub_pixel}
\end{figure}

\begin{figure}[!t]
    \centering
    {
        \setlength{\fboxrule}{1pt}
        {\includegraphics[width=0.35\textwidth]{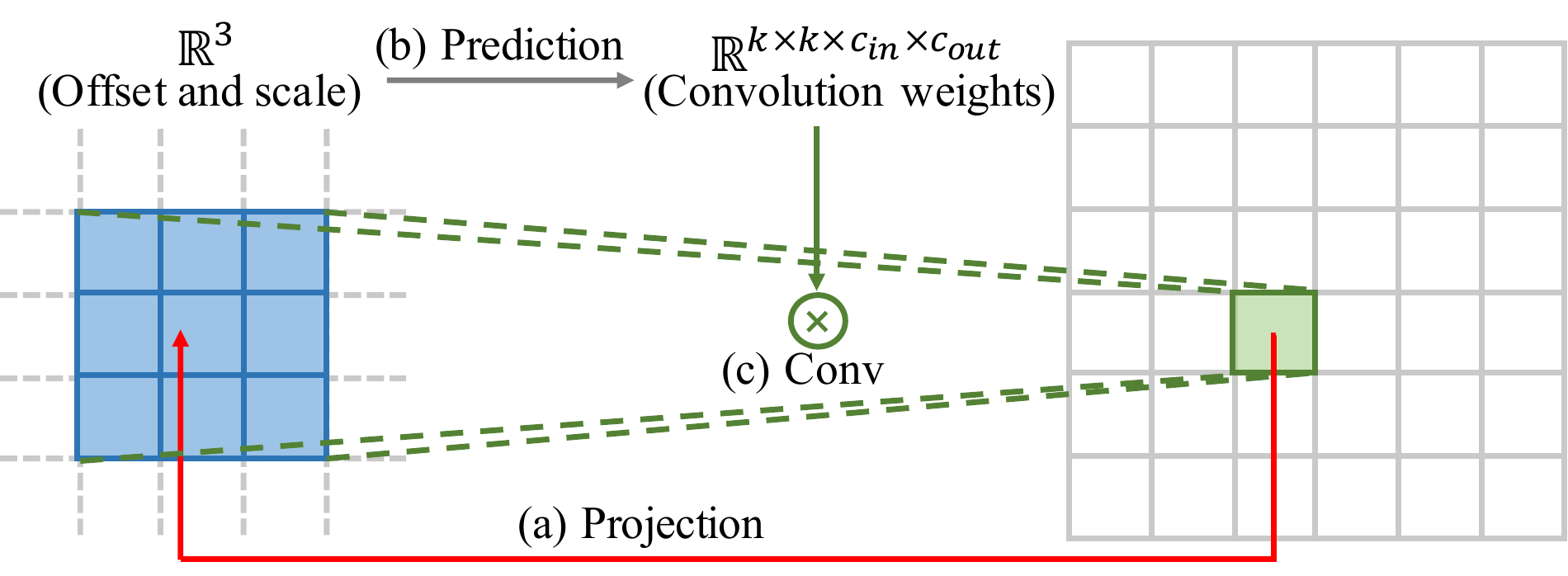}%
        }
    }
    \caption{
        Meta upscale module.
        The \textcolor{CornflowerBlue}{blue} boxes denote the projection patch, and the \textcolor{LimeGreen}{green} boxes and lines indicate the convolution operation with predicted weights.
    }
    \label{fig_meta_upscale}
\end{figure}

In order to overcome the shortcomings of interpolation-based methods and learn upsampling in an end-to-end manner, transposed convolution layer and sub-pixel layer are introduced into the SR field.

\textbf{Transposed Convolution Layer.}
Transposed convolution layer, a.k.a. deconvolution layer \cite{CVPRW2010DeconvolutionalZeiler,ECCV2014VisualizingZeiler}, tries to perform transformation opposite a normal convolution, i.e., predicting the possible input based on feature maps sized like convolution output.
Specifically, it increases the image resolution by expanding the image by inserting zeros and performing convolution.
Taking $2\times$ SR with $3 \times 3$ kernel as example (as Fig. \ref{fig_deconv} shows), the input is firstly expanded twice of the original size, where the added pixel values are set to $0$ (Fig. \ref{fig_deconv_1}).
Then a convolution with kernel sized $3 \times 3$, stride $1$ and padding $1$ is applied (Fig. \ref{fig_deconv_2}).
In this way, the input is upsampled by a factor of 2, in which case the receptive field is at most $2 \times 2$.
Since the transposed convolution enlarges the image size in an end-to-end manner while maintaining a connectivity pattern compatible with vanilla convolution, it is widely used as upsampling layers in SR models \cite{NIPS2016ImageMao,ICCV2017ImageTong,CVPR2018ImageHan,CVPR2018DeepHaris}.
However, this layer can easily cause ``uneven overlapping'' on each axis \cite{Distill2016DeconvolutionOdena}, and the multiplied results on both axes further create a checkerboard-like pattern of varying magnitudes and thus hurt the SR performance.

\textbf{Sub-pixel Layer.}
The sub-pixel layer \cite{CVPR2016RealShi}, another end-to-end learnable upsampling layer, performs upsampling by generating a plurality of channels by convolution and then reshaping them, as Fig. \ref{fig_sub_pixel} shows.
Within this layer, a convolution is firstly applied for producing outputs with $s^2$ times channels, where $s$ is the scaling factor (Fig. \ref{fig_sub_pixel_1}).
Assuming the input size is $h \times w \times c$, the output size will be $h \times w \times s^2 c$.
After that, the reshaping operation (a.k.a. \textit{shuffle} \cite{CVPR2016RealShi}) is performed to produce outputs with size $sh \times sw \times c$ (Fig. \ref{fig_sub_pixel_2}).
In this case, the receptive field can be up to $3 \times 3$.
Due to the end-to-end upsampling manner, this layer is also widely used by SR models \cite{CVPR2017PhotoLedig,CVPR2018LearningZhang,CVPR2018ResidualZhang,ECCV2018FastAhn}.
Compared with transposed convolution layer, the sub-pixel layer has a larger receptive field, which provides more contextual information to help generate more realistic details.
However, since the distribution of the receptive fields is uneven and blocky regions actually share the same receptive field, it may result in some artifacts near the boundaries of different blocks.
On the other hand, independently predicting adjacent pixels in a blocky region may cause unsmooth outputs.
Thus Gao \textit{et al.} \cite{TPAMI2019PixelGao} propose PixelTCL, which replaces the independent prediction to interdependent sequential prediction, and produces smoother and more consistent results.

\textbf{Meta Upscale Module.}
 The previous methods need to predefine the scaling factors, i.e., training different upsampling modules for different factors, which is inefficient and not in line with real needs.
So that Hu \textit{et al.} \cite{CVPR2019MetaHu} propose meta upscale module (as Fig. \ref{fig_meta_upscale} shows), which firstly solves SR of arbitrary scaling factors based on meta learning.
Specifically, for each target position on the HR images, this module project it to a small patch on the LR feature maps (i.e., $k \times k \times c_{in}$), predicts convolution weights (i.e., $k \times k \times c_{in} \times c_{out}$) according to the projection offsets and the scaling factor by dense layers and perform convolution.
In this way, the meta upscale module can continuously zoom in it with arbitrary factors by a single model.
And due to the large amount of training data (multiple factors are simultaneously trained), the module can exhibit comparable or even better performance on fixed factors.
Although this module needs to predict weights during inference, the execution time of the upsampling module only accounts for about 1\% of the time of feature extraction \cite{CVPR2019MetaHu}.
However, this method predicts a large number of convolution weights for each target pixel based on several values independent of the image contents, so the prediction result may be unstable and less efficient when faced with larger magnifications.

Nowadays, these learning-based layers have become the most widely used upsampling methods.
Especially in the post-upsampling framework (Sec. \ref{sec_framework_post}), these layers are usually used in the final upsampling phase for reconstructing HR images based on high-level representations extracted in low-dimensional space, and thus achieve end-to-end SR while avoiding overwhelming operations in high-dimensional space.

\subsection{Network Design}
\label{sec_network_design}

\begin{figure*}[!t]
    \centering
    \subfloat[Residual Learning]{\includegraphics[width=0.25\textwidth]{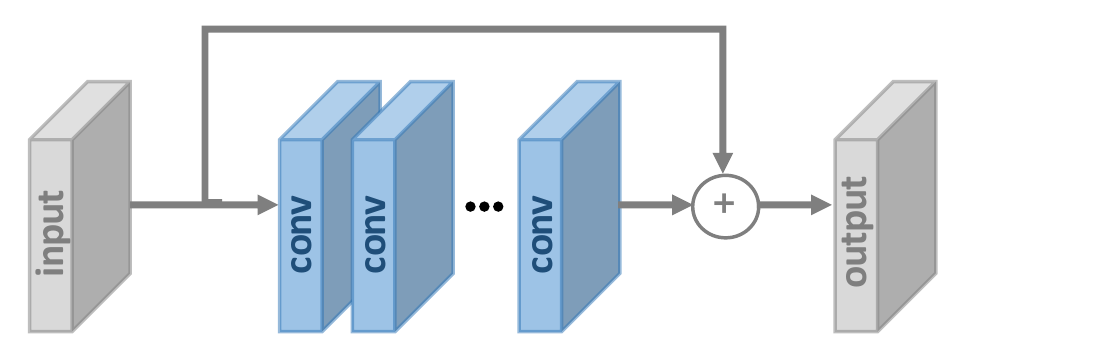}%
    \label{fig_residual_learning}}
    \subfloat[Recursive learning]{\includegraphics[width=0.25\textwidth]{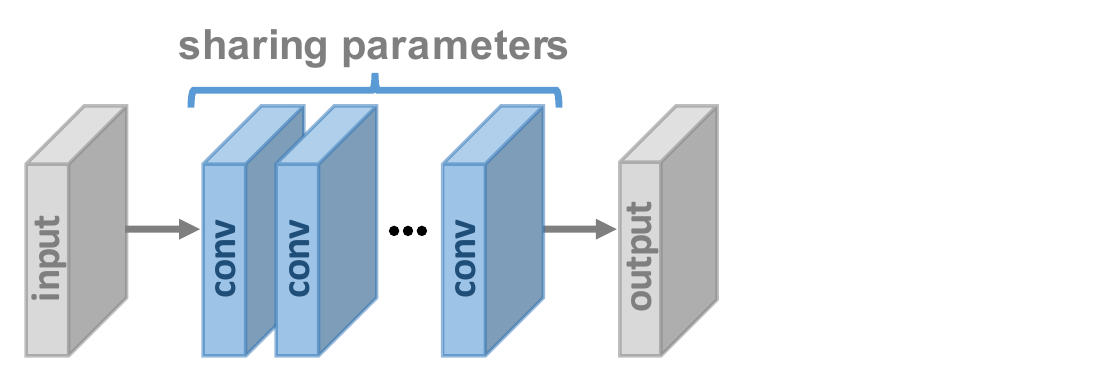}%
    \label{fig_recursive_learning}}
    \subfloat[Channel attention]{\includegraphics[width=0.25\textwidth]{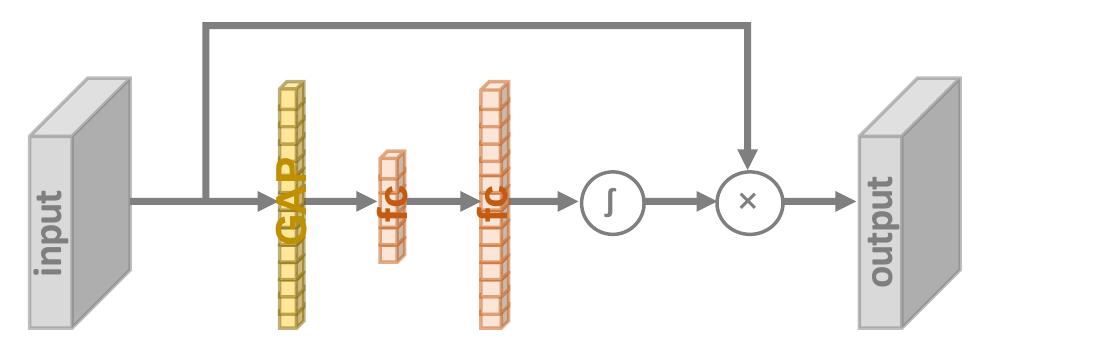}%
    \label{fig_channel_attention}}
    \subfloat[Dense connections]{\includegraphics[width=0.25\textwidth]{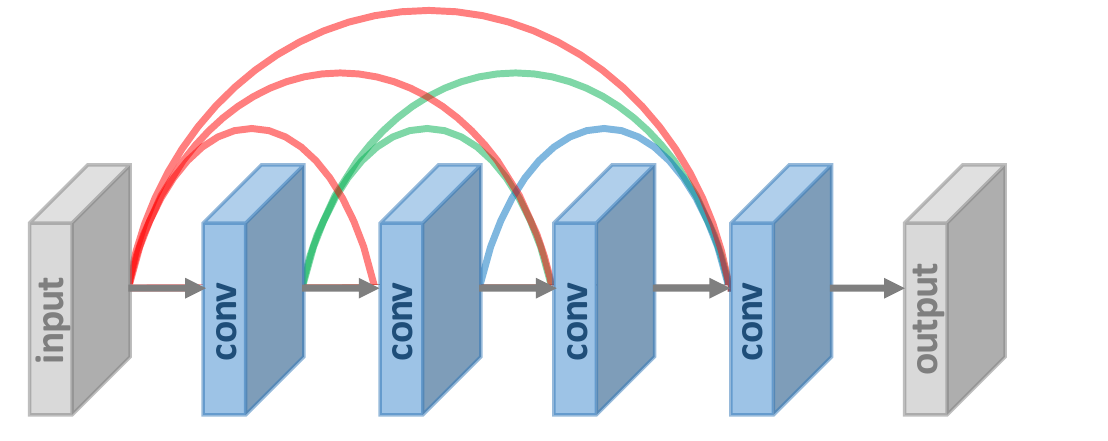}%
    \label{fig_dense_connections}} \\
    \subfloat[Local multi-path learning]{\includegraphics[width=0.25\textwidth]{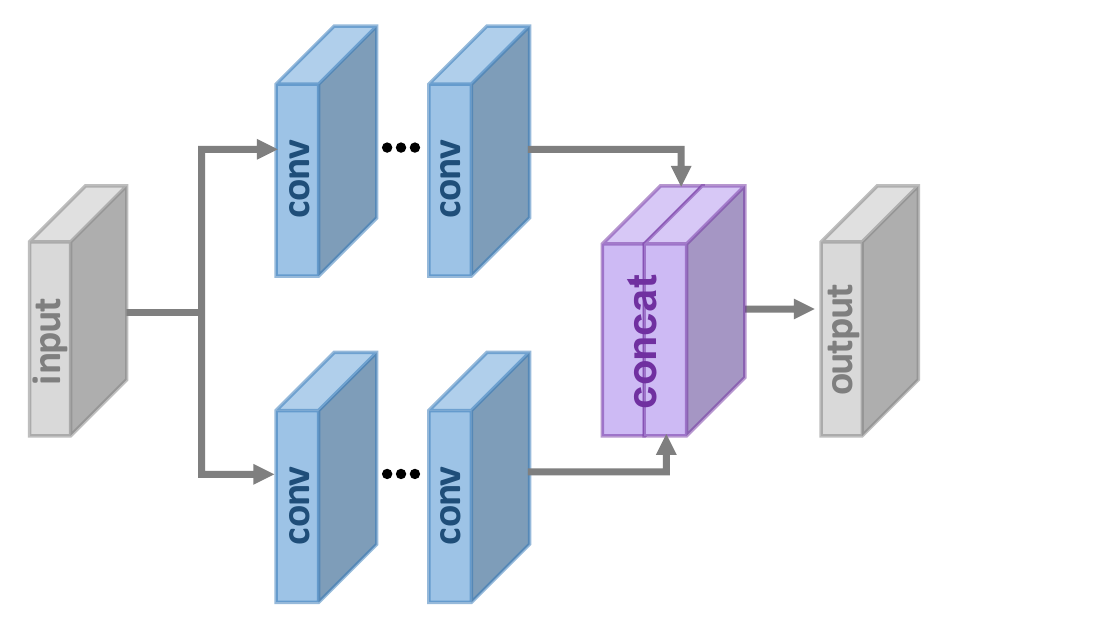}%
    \label{fig_multi_path_learning}}
    \subfloat[Scale-specific multi-path learning]{\includegraphics[width=0.25\textwidth]{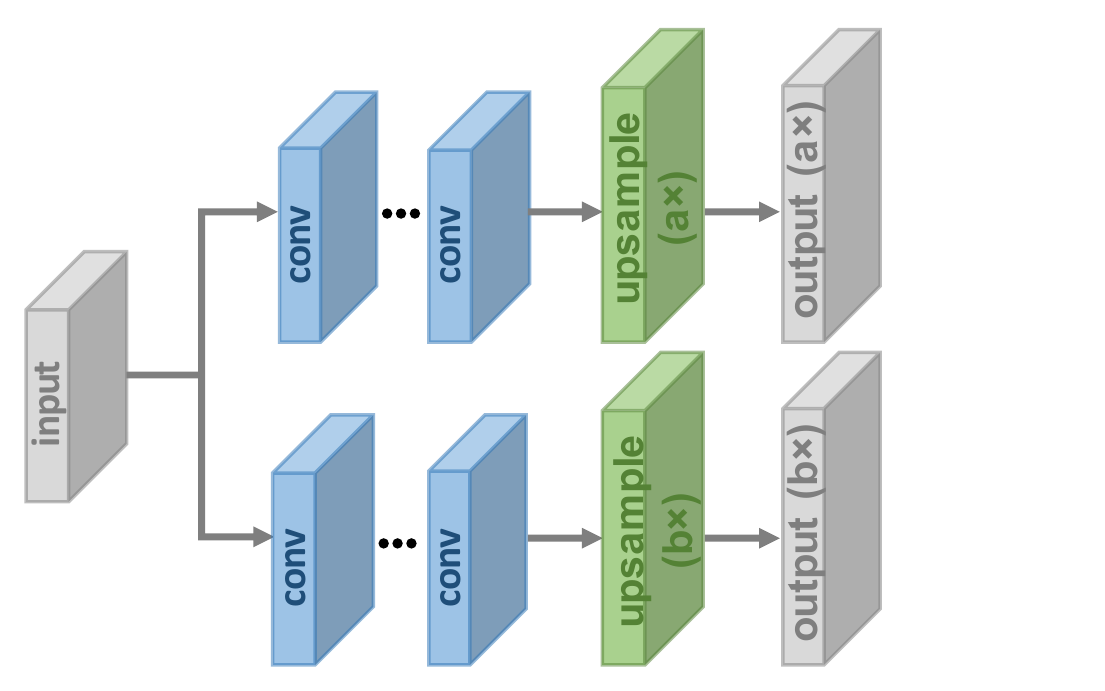}%
    \label{fig_multi_path_learning_specific}}
    \subfloat[Group convolution]{\includegraphics[width=0.25\textwidth]{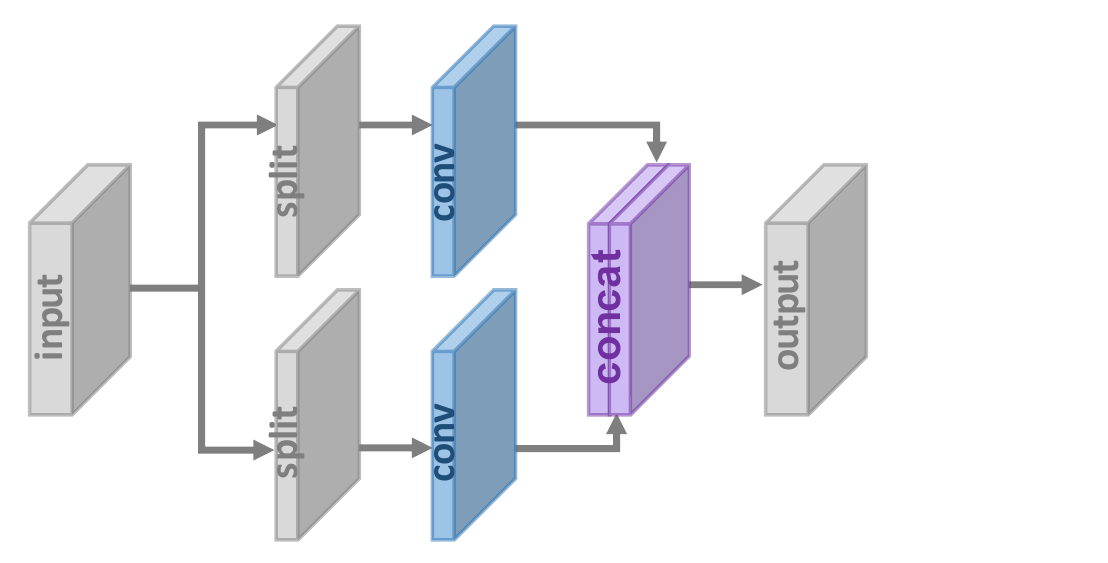}%
    \label{fig_group_conv}}
    \subfloat[Pyramid pooling]{\includegraphics[width=0.25\textwidth]{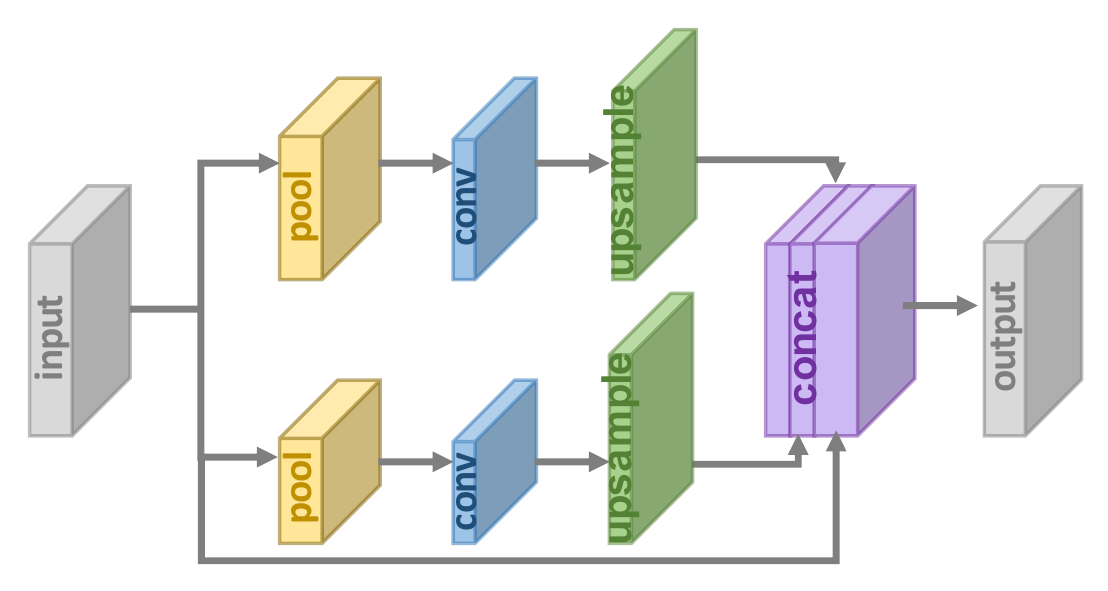}%
    \label{fig_pyramid_pooling}} \\
    \caption{
        Network design strategies.
    }
    \label{fig_modelling_strategies}
\end{figure*}

Nowadays the network design has been one of the most important parts of deep learning.
In the super-resolution field, researchers apply all kinds of network design strategies on top of the four SR frameworks (Sec. \ref{sec_sr_frameworks}) to construct the final networks.
In this section, we decompose these networks to essential principles or strategies for network design, introduce them and analyze the advantages and limitations one by one.

\subsubsection{Residual Learning}
\label{sec_residual_learning}

Before He \textit{et al.} \cite{CVPR2016DeepHe} propose ResNet for learning residuals instead of a thorough mapping, residual learning has been widely employed by SR models \cite{ICCV2013AnchoredTimofte,BMVC2012LowBevilacqua,ACCV2014AdjustedTimofte}, as Fig. \ref{fig_residual_learning} shows.
Among them, the residual learning strategies can be roughly divided into global and local residual learning.

\textbf{Global Residual Learning.}
Since the image SR is an image-to-image translation task where the input image is highly correlated with the target image, researchers try to learn only the residuals between them, namely global residual learning.
In this case, it avoids learning a complicated transformation from a complete image to another, instead only requires learning a residual map to restore the missing high-frequency details.
Since the residuals in most regions are close to zero, the model complexity and learning difficulty are greatly reduced.
Thus it is widely used by SR models \cite{CVPR2016AccurateKim,CVPR2017ImageTai,ICCV2017MemNetTai,CVPR2018FastHui}.

\textbf{Local Residual Learning.}
The local residual learning is similar to the residual learning in ResNet \cite{CVPR2016DeepHe} and used to alleviate the degradation problem \cite{CVPR2016DeepHe} caused by ever-increasing network depths, reduce training difficulty and improve the learning ability.
It is also widely used for SR \cite{NIPS2016ImageMao,CVPR2018ImageHan,ECCV2018ImageZhang,ECCV2018MultiLi}.

In practice, the above methods are both implemented by shortcut connections (often scaled by a small constant) and element-wise addition, while the difference is that the former directly connects the input and output images, while the latter usually adds multiple shortcuts between layers with different depths inside the network.

\subsubsection{Recursive Learning}
\label{sec_recursive_learning}

In order to learn higher-level features without introducing overwhelming parameters, recursive learning, which means applying the same modules multiple times in a recursive manner, is introduced into the SR field, as Fig. \ref{fig_recursive_learning} shows.

Among them, the 16-recursive DRCN\cite{CVPR2016DeeplyKim} employs a single convolutional layer as the recursive unit and reaches a receptive field of $41 \times 41$, which is much larger than $13 \times 13$ of SRCNN \cite{ECCV2014LearningDong}, without over many parameters.
The DRRN \cite{CVPR2017ImageTai} uses a ResBlock \cite{CVPR2016DeepHe} as the recursive unit for 25 recursions and obtains even better performance than the 17-ResBlock baseline.
Later Tai \textit{et al.} \cite{ICCV2017MemNetTai} propose MemNet based on the memory block, which is composed of a 6-recursive ResBlock where the outputs of every recursion are concatenated and go through an extra $1 \times 1$ convolution for memorizing and forgetting.
The cascading residual network (CARN) \cite{ECCV2018FastAhn} also adopts a similar recursive unit including several ResBlocks.
Recently, Li \textit{et al.} \cite{CVPR2019FeedbackLi} employ iterative up-and-down sampling SR framework, and propose a feedback network based on recursive learning, where the weights of the entire network are shared across all recursions.

Besides, researchers also employ different recursive modules in different parts.
Specifically, Han \textit{et al.} \cite{CVPR2018ImageHan} propose dual-state recurrent network (DSRN) to exchange signals between the LR and HR states.
At each time step (i.e., recursion), the representations of each branch are updated and exchanged for better exploring LR-HR relationships.
Similarly, Lai \textit{et al.} \cite{TPAMI2018FastLai} employ the embedding and upsampling modules as recursive units, and thus much reduce the model size at the expense of little performance loss.  

In general, the recursive learning can indeed learn more advanced representations without introducing excessive parameters, but still can't avoid high computational costs.
And it inherently brings vanishing or exploding gradient problems, consequently some techniques such as residual learning (Sec. \ref{sec_residual_learning}) and multi-supervision (Sec. \ref{sec_multi_supervision}) are often integrated with recursive learning for mitigating these problems \cite{CVPR2016DeeplyKim,CVPR2017ImageTai,ICCV2017MemNetTai,CVPR2018ImageHan}.

\subsubsection{Multi-path Learning}
\label{sec_multi_path_learning}

Multi-path learning refers to passing features through multiple paths, which perform different operations, and fusing them back for providing better modelling capabilities.
Specifically, it could be divided into global, local and scale-specific multi-path learning, as bellow.

\textbf{Global Multi-path Learning.}
Global multi-path learning refers to making use of multiple paths to extract features of different aspects of the images.
These paths can cross each other in the propagation and thus greatly enhance the learning ability.
Specifically, the LapSRN \cite{CVPR2017DeepLai} includes a feature extraction path predicting the sub-band residuals in a coarse-to-fine fashion and another path to reconstruct HR images based on the signals from both paths.
Similarly, the DSRN \cite{CVPR2018ImageHan} utilizes two paths to extract information in low-dimensional and high-dimensional space, respectively, and continuously exchanges information for further improving learning ability.
And the pixel recursive super-resolution \cite{ICCV2017PixelDahl} adopts a conditioning path to capture the global structure of images, and a prior path to capture the serial dependence of generated pixels.
In contrast, Ren \textit{et al.} \cite{CVPRW2017ImageRen} employ multiple paths with unbalanced structures to perform upsampling and fuse them at the end of the model.

\textbf{Local Multi-path Learning.}
Motivated by the inception module \cite{CVPR2015GoingSzegedy}, the MSRN \cite{ECCV2018MultiLi} adopts a new block for multi-scale feature extraction, as Fig. \ref{fig_multi_path_learning} shows.
In this block, two convolution layers with kernel size $3\times 3$ and $5\times 5$ are adopted to extract features simultaneously, then the outputs are concatenated and go through the same operations again, and finally an extra $1\times 1$ convolution is applied.
A shortcut connects the input and output by element-wise addition.
Through such local multi-path learning, the SR models can better extract image features from multiple scales and further improve performance.

\textbf{Scale-specific Multi-path Learning.}
Considering that SR models for different scales need to go through similar feature extraction, Lim \textit{et al.} \cite{CVPRW2017EnhancedLim} propose scale-specific multi-path learning to cope with multi-scale SR with a single network.
To be concrete, they share the principal components of the model (i.e., the intermediate layers for feature extraction), and attach scale-specific pre-processing paths and upsampling paths at the beginning and the end of the network, respectively (as Fig. \ref{fig_multi_path_learning_specific} shows).
During training, only the paths corresponding to the selected scale are enabled and updated.
In this way, the proposed MDSR \cite{CVPRW2017EnhancedLim} greatly reduce the model size by sharing most of the parameters for different scales and exhibits comparable performance as single-scale models.
The similar scale-specific multi-path learning is also adopted by CARN \cite{ECCV2018FastAhn} and ProSR \cite{CVPRW2018FullyWang}.

\subsubsection{Dense Connections}
\label{sec_dense_connections}

Since Huang \textit{et al.} \cite{CVPR2017DenselyHuang} propose DenseNet based on dense blocks, the dense connections have become more and more popular in vision tasks.
For each layer in a dense block, the feature maps of all preceding layers are used as inputs, and its own feature maps are used as inputs into all subsequent layers, so that it leads to $l \cdot (l - 1) / 2$ connections in a $l$-layer dense block ($l \ge 2$).
The dense connections not only help alleviate gradient vanishing, enhance signal propagation and encourage feature reuse, but also substantially reduce the model size by employing small growth rate (i.e., number of channels in dense blocks) and squeezing channels after concatenating all input feature maps.

For the sake of fusing low-level and high-level features to provide richer information for reconstructing high-quality details, dense connections are introduced into the SR field, as Fig. \ref{fig_dense_connections} shows.
Tong \textit{et al.} \cite{ICCV2017ImageTong} not only adopt dense blocks to construct a 69-layers SRDenseNet, but also insert dense connections between different dense blocks, i.e., for every dense block, the feature maps of all preceding blocks are used as inputs, and its own feature maps are used as inputs into all subsequent blocks.
These layer-level and block-level dense connections are also adopted by MemNet \cite{ICCV2017MemNetTai}, CARN \cite{ECCV2018FastAhn}, RDN \cite{CVPR2018ResidualZhang} and ESRGAN \cite{ECCVW2018ESRGANWang}.
The DBPN \cite{CVPR2018DeepHaris} also adopts dense connections extensively, but their dense connections are between all the upsampling units, as are the downsampling units.

\subsubsection{Attention Mechanism}
\label{sec_attention_mechanism}

\textbf{Channel Attention.}
Considering the interdependence and interaction of the feature representations between different channels, Hu \textit{et al.} \cite{CVPR2018SqueezeHu} propose a ``squeeze-and-excitation'' block to improve learning ability by explicitly modelling channel interdependence, as Fig. \ref{fig_channel_attention} shows.
In this block, each input channel is squeezed into a channel descriptor (i.e., a constant) using global average pooling (GAP), then these descriptors are fed into two dense layers to produce channel-wise scaling factors for input channels.
Recently, Zhang \textit{et al.} \cite{ECCV2018ImageZhang} incorporate the channel attention mechanism with SR and propose RCAN, which markedly improves the representation ability of the model and SR performance.
In order to better learn the feature correlations, Dai \textit{et al.} \cite{CVPR2019SecondDai} further propose a second-order channel attention (SOCA) module.
The SOCA adaptively rescales the channel-wise features by using second-order feature statistics instead of GAP, and enables extracting more informative and discriminative representations.

\textbf{Non-local Attention.}
Most existing SR models have very limited local receptive fields.
However, some distant objects or textures may be very important for local patch generation.
So that Zhang \textit{et al.} \cite{ICLR2019ResidualZhang} propose local and non-local attention blocks to extract features that capture the long-range dependencies between pixels.
Specifically, they propose a trunk branch for extracting features, and a (non-)local mask branch for adaptively rescaling features of trunk branch.
Among them, the local branch employs an encoder-decoder structure to learn the local attention, while the non-local branch uses the embedded Gaussian function to evaluate pairwise relationships between every two position indices in the feature maps to predict the scaling weights.
Through this mechanism, the proposed method captures the spatial attention well and further enhances the representation ability.
Similarly, Dai \textit{et al.} \cite{CVPR2019SecondDai} also incorporate the non-local attention mechanism to capture long-distance spatial contextual information.

\subsubsection{Advanced Convolution}
\label{sec_advanced_convolution}

Since convolution operations are the basis of deep neural networks, researchers also attempt to improve convolution operations for better performance or greater efficiency.

\textbf{Dilated Convolution.}
It is well known that the contextual information facilitates generating realistic details for SR.
Thus Zhang \textit{et al.} \cite{CVPR2017LearningZhang} replace the common convolution by dilated convolution in SR models, increase the receptive field over twice and achieve much better performance.

\textbf{Group Convolution.}
Motivated by recent advances on lightweight CNNs \cite{CVPR2017AggregatedXie,CVPR2017XceptionChollet}, Hui \textit{et al.} \cite{CVPR2018FastHui} and Ahn \textit{et al.} \cite{ECCV2018FastAhn} propose IDN and CARN-M, respectively, by replacing the vanilla convolution by group convolution.
As some previous works have proven, the group convolution much reduces the number of parameters and operations at the expense of a little performance loss \cite{CVPR2018FastHui,ECCV2018FastAhn}.

\textbf{Depthwise Separable Convolution}
Since Howard \textit{et al.} \cite{Arxiv2017MobilenetsHoward} propose depthwise separable convolution for efficient convolution, it has been expanded to into various fields.
Specifically, it consists of a factorized depthwise convolution and a pointwise convolution (i.e., $1 \times 1$ convolution), and thus reduces plenty of parameters and operations at only a small reduction in accuracy \cite{Arxiv2017MobilenetsHoward}.
And recently, Nie \textit{et al.} \cite{ECCVW2018PIRMIgnatov} employ the depthwise separable convolution and much accelerate the SR architecture.

\subsubsection{Region-recursive Learning}
\label{sec_region_recursive_learning}

Most SR models treat SR as a pixel-independent task and thus cannot source the interdependence between generated pixels properly.
Inspired by PixelCNN \cite{NIPS2016ConditionalAaron}, Dahl \textit{et al.} \cite{ICCV2017PixelDahl} firstly propose pixel recursive learning to perform pixel-by-pixel generation, by employing two networks to capture global contextual information and serial generation dependence, respectively.
In this way, the proposed method synthesizes realistic hair and skin details on super-resolving very low-resolution face images (e.g., $8\times 8$) and far exceeds the previous methods on MOS testing \cite{ICCV2017PixelDahl} (Sec. \ref{sec_iqa_mos}).

Motivated by the human attention shifting mechanism \cite{Nature2005OptimalNajemnik}, the Attention-FH \cite{CVPR2017AttentionCao} also adopts this strategy by resorting to a recurrent policy network for sequentially discovering attended patches and performing local enhancement.
In this way, it is capable of adaptively personalizing an optimal searching path for each image according to its own characteristic, and thus fully exploits the global intra-dependence of images.

Although these methods show better performance to some extent, the recursive process requiring a long propagation path greatly increases the computational cost and training difficulty, especially for super-resolving HR images.

\subsubsection{Pyramid Pooling}
\label{sec_pyramid_pooling}

Motivated by the spatial pyramid pooling layer \cite{ECCV2014SpatialHe}, Zhao \textit{et al.} \cite{CVPR2017PyramidZhao} propose the pyramid pooling module to better utilize global and local contextual information.
Specifically, for feature maps sized $h\times w\times c$, each feature map is divided into $M\times M$ bins, and goes through global average pooling, resulting in $M\times M\times c$ outputs.
Then a $1\times 1$ convolution is performed for compressing the outputs to a single channel.
After that, the low-dimensional feature map is upsampled to the same size as the original feature map via bilinear interpolation.
By using different $M$, the module integrates global as well as local contextual information effectively.
By incorporating this module, the proposed EDSR-PP model \cite{CVPRW2018EfficientPark} further improve the performance over baselines.

\subsubsection{Wavelet Transformation}
\label{sec_wavelet_transformation}

As is well-known, the wavelet transformation (WT) \cite{SBook1992TenDaubechies,EBook1999WaveletMallat} is a highly efficient representation of images by decomposing the image signal into high-frequency sub-bands denoting texture details and low-frequency sub-bands containing global topological information.
Bae \textit{et al.} \cite{CVPRW2017BeyondBae} firstly combine WT with deep learning based SR model, take sub-bands of interpolated LR wavelet as input and predict residuals of corresponding HR sub-bands.
WT and inverse WT are applied for decomposing the LR input and reconstructing the HR output, respectively.
Similarly, the DWSR \cite{CVPRW2017DeepGuo} and Wavelet-SRNet \cite{ICCV2017WaveletHuang} also perform SR in the wavelet domain but with more complicated structures.
In contrast to the above works processing each sub-band independently, the MWCNN \cite{CVPRW2018MultiLiu} adopts multi-level WT and takes the concatenated sub-bands as the input to a single CNN for better capturing the dependence between them.
Due to the efficient representation by wavelet transformation, the models using this strategy often much reduce the model size and computational cost, while maintain competitive performance \cite{CVPRW2017BeyondBae,CVPRW2018MultiLiu}.

\subsubsection{Desubpixel}
\label{sec_desubpixel}

In order to speed up the inference speed, Vu \textit{et al.} \cite{ECCVW2018FastVu} propose to perform the time-consuming feature extraction in a lower-dimensional space, and propose desubpixel, an inverse of the shuffle operation of sub-pixel layer (Sec. \ref{sec_learning_based_upsampling}).  
Specifically, the desubpixel operation splits the images spatially, stacks them as extra channels and thus avoids loss of information.
In this way, they downsample input images by desubpixel at the beginning of the model, learn representations in a lower-dimensional space, and upsample to the target size at the end.  
The proposed model achieves the best scores in the PIRM Challenge on Smartphones \cite{ECCVW2018PIRMIgnatov} with very high-speed inference and good performance.

\subsubsection{xUnit}
\label{sec_xunit}

In order to combine spatial feature processing and nonlinear activations to learn complex features more efficiently, Kligvasser \textit{et al.} \cite{CVPR2018xUnitKligvasser} propose xUnit for learning a spatial activation function.
Specifically, the ReLU is regarded as determining a weight map to perform element-wise multiplication with the input, while the xUnit directly learn the weight map through convolution and Gaussian gating.
Although the xUnit is more computationally demanding, due to its dramatic effect on the performance, it allows greatly reducing the model size while matching the performance with ReLU.
In this way, the authors reduce the model size by nearly 50\% without any performance degradation.

\subsection{Learning Strategies}
\label{sec_learning_strategies}

\subsubsection{Loss Functions}
\label{sec_loss}

In the super-resolution field, loss functions are used to measure reconstruction error and guide the model optimization.
In early times, researchers usually employ the pixel-wise L2 loss, but later discover that it cannot measure the reconstruction quality very accurately.
Therefore, a variety of loss functions (e.g., content loss \cite{ECCV2016PerceptualJohnson}, adversarial loss \cite{CVPR2017PhotoLedig}) are adopted for better measuring the reconstruction error and producing more realistic and higher-quality results.
Nowadays these loss functions have been playing an important role.
In this section, we'll take a closer look at the loss functions used widely.
The notations in this section follow Sec. \ref{sec_problem_definitions}, except that we ignore the subscript $y$ of the target HR image $\hat{I_y}$ and generated HR image $I_y$ for brevity.

\textbf{Pixel Loss.}
Pixel loss measures pixel-wise difference between two images and mainly includes L1 loss (i.e., mean absolute error) and L2 loss (i.e., mean square error):
\begin{align}
    \mathcal{L}_{\text{pixel\_l1}} (\hat{I}, I) &= \frac{1}{hwc} \sum_{i,j,k} |\hat{I}_{i,j,k} - I_{i,j,k}| , \\
    \mathcal{L}_{\text{pixel\_l2}} (\hat{I}, I) &= \frac{1}{hwc} \sum_{i,j,k} (\hat{I}_{i,j,k} - I_{i,j,k})^2 ,
\end{align}
where $h$, $w$ and $c$ are the height, width and number of channels of the evaluated images, respectively.
In addition, there is a variant of the pixel L1 loss, namely Charbonnier loss \cite{IJCV2005LucasBruhn,CVPR2017DeepLai}, given by:.
\begin{equation}
    \mathcal{L}_{\text{pixel\_Cha}} (\hat{I}, I) = \frac{1}{hwc} \sum_{i,j,k} \sqrt{(\hat{I}_{i,j,k} - I_{i,j,k})^2 + \epsilon^2} ,
\end{equation}
where $\epsilon$ is a constant (e.g., $10^{-3}$) for numerical stability.

The pixel loss constrains the generated HR image $\hat{I}$ to be close enough to the ground truth $I$ on the pixel values.
Comparing with L1 loss, the L2 loss penalizes larger errors but is more tolerant to small errors, and thus often results in too smooth results.
In practice, the L1 loss shows improved performance and convergence over L2 loss \cite{CVPRW2017EnhancedLim,TCI2017LossZhao,ECCV2018FastAhn}.
Since the definition of PSNR (Sec. \ref{sec_iqa_psnr}) is highly correlated with pixel-wise difference and minimizing pixel loss directly maximize PSNR, the pixel loss gradual becomes the most widely used loss function.
However, since the pixel loss actually doesn't take image quality (e.g., perceptual quality \cite{ECCV2016PerceptualJohnson}, textures \cite{ICCV2017EnhancenetSajjadi}) into account, the results often lack high-frequency details and are perceptually unsatisfying with oversmooth textures \cite{TIP2004ImageWang,ACSSC2003MultiWang,ECCV2016PerceptualJohnson,CVPR2017PhotoLedig}.

\textbf{Content Loss.}
In order to evaluate perceptual quality of images, the content loss is introduced into SR \cite{NIPS2016GeneratingDosovitskiy,ECCV2016PerceptualJohnson}.
Specifically, it measures the semantic differences between images using a pre-trained image classification network.
Denoting this network as $\phi$ and the extracted high-level representations on $l$-th layer as $\phi^{(l)}(I)$, the content loss is indicated as the Euclidean distance between high-level representations of two images, as follows:
\begin{equation}
    \mathcal{L}_{\text{content}} (\hat{I}, I ; \phi, l) = \frac{1}{h_l w_l c_l} \sqrt{\sum_{i,j,k} (\phi^{(l)}_{i,j,k}(\hat{I}) - \phi^{(l)}_{i,j,k}(I))^2} ,
\end{equation}
where $h_l$, $w_l$ and $c_l$ are the height, width and number of channels of the representations on layer $l$, respectively.

Essentially the content loss transfers the learned knowledge of hierarchical image features from the classification network $\phi$ to the SR network.
In contrast to the pixel loss, the content loss encourages the output image $\hat{I}$ to be perceptually similar to the target image $I$ instead of forcing them to match pixels exactly.
Thus it produces visually more perceptible results and is also widely used in this field \cite{ECCV2016PerceptualJohnson,CVPR2017PhotoLedig,ICCV2017EnhancenetSajjadi,CVPR2018RecoveringWang,CVPR2018SuperBulat,ECCVW2018ESRGANWang}, where the VGG \cite{ICLR2015VerySimonyan} and ResNet \cite{CVPR2016DeepHe} are the most commonly used pre-trained CNNs.

\textbf{Texture Loss.}
On account that the reconstructed image should have the same style (e.g., colors, textures, contrast) with the target image, and motivated by the style representation by Gatys \textit{et al.} \cite{NIPS2015TextureGatys,CVPR2016ImageGatys}, the texture loss (a.k.a style reconstruction loss) is introduced into SR.
Following \cite{NIPS2015TextureGatys,CVPR2016ImageGatys}, the image texture is regarded as the correlations between different feature channels and defined as the Gram matrix $G^{(l)} \in \mathbb{R}^{c_l \times c_l}$, where $G^{(l)}_{ij}$ is the inner product between the vectorized feature maps $i$ and $j$ on layer $l$:
\begin{equation}
    G^{(l)}_{ij}(I) = \operatorname{vec}(\phi_i^{(l)}(I)) \cdot \operatorname{vec}(\phi_j^{(l)}(I)) ,
\end{equation}
where $\operatorname{vec}(\cdot)$ denotes the vectorization operation, and $\phi_i^{(l)}(I)$ denotes the $i$-th channel of the feature maps on layer $l$ of image $I$.
Then the texture loss is given by:
\begin{equation}
    \mathcal{L}_{\text{texture}} (\hat{I}, I ; \phi, l) = \frac{1}{c_l^2} \sqrt{\sum_{i,j} (G^{(l)}_{i,j}(\hat{I}) - G^{(l)}_{i,j}(I))^2} .
\end{equation}

By employing texture loss, the EnhanceNet \cite{ICCV2017EnhancenetSajjadi} proposed by Sajjadi \textit{et al.} creates much more realistic textures and produces visually more satisfactory results.
Despite this, determining the patch size to match textures is still empirical.
Too small patches lead to artifacts in textured regions, while too large patches lead to artifacts throughout the entire image because texture statistics are averaged over regions of varying textures.


\textbf{Adversarial Loss.}
In recent years, due to the powerful learning ability, the GANs \cite{NIPS2014GenerativeGoodfellow} receive more and more attention and are introduced to various vision tasks.
To be concrete, the GAN consists of a generator performing generation (e.g., text generation, image transformation), and a discriminator which takes the generated results and instances sampled from the target distribution as input and discriminates whether each input comes from the target distribution.
During training, two steps are alternately performed:
(a) fix the generator and train the discriminator to better discriminate,
(b) fix the discriminator and train the generator to fool the discriminator.
Through adequate iterative adversarial training, the resulting generator can produce outputs consistent with the distribution of real data, while the discriminator can't distinguish between the generated data and real data.

In terms of super-resolution, it is straightforward to adopt adversarial learning, in which case we only need to treat the SR model as a generator, and define an extra discriminator to judge whether the input image is generated or not.
Therefore, Ledig \textit{et al.} \cite{CVPR2017PhotoLedig} firstly propose SRGAN using adversarial loss based on cross entropy, as follows:
\begin{align}
    \mathcal{L}_{\text{gan\_ce\_g}} (\hat{I} ; D) &= - \log D(\hat{I}) , \\
    \mathcal{L}_{\text{gan\_ce\_d}} (\hat{I}, I_s ; D) &= - \log D(I_s) - \log (1 - D(\hat{I})) ,
\end{align}
where $\mathcal{L}_{\text{gan\_ce\_g}}$ and $\mathcal{L}_{\text{gan\_ce\_d}}$ denote the adversarial loss of the generator (i.e., the SR model) and the discriminator $D$ (i.e., a binary classifier), respectively, and $I_s$ represents images randomly sampled from the ground truths.
Besides, the Enhancenet \cite{ICCV2017EnhancenetSajjadi} also adopts the similar adversarial loss.

Besides, Wang \textit{et al.} \cite{CVPRW2018FullyWang} and Yuan \textit{et al.} \cite{CVPRW2018UnsupervisedYuan} use adversarial loss based on least square error for more stable training process and higher quality results \cite{ICCV2017LeastMao}, given by:
\begin{align}
    \mathcal{L}_{\text{gan\_ls\_g}} (\hat{I} ; D) &= (D(\hat{I}) - 1)^2 , \\
    \mathcal{L}_{\text{gan\_ls\_d}} (\hat{I}, I_s ; D) &= (D(\hat{I}))^2 + (D(I_s) - 1)^2 .
\end{align}

In contrast to the above works focusing on the specific forms of adversarial loss, Park \textit{et al.} \cite{ECCV2018SRFeatPark} argue that the pixel-level discriminator causes generating meaningless high-frequency noise, and attach another feature-level discriminator to operate on high-level representations extracted by a pre-trained CNN which captures more meaningful attributes of real HR images.  
Xu \textit{et al.} \cite{ICCV2017LearningXu} incorporate a multi-class GAN consisting of a generator and multiple class-specific discriminators.
And the ESRGAN \cite{ECCVW2018ESRGANWang} employs relativistic GAN \cite{Arxiv2018RelativisticJolicoeur} to predict the probability that real images are relatively more realistic than fake ones, instead of the probability that input images are real or fake, and thus guide recovering more detailed textures.

Extensive MOS tests (Sec. \ref{sec_iqa_mos}) show that even though the SR models trained with adversarial loss and content loss achieve lower PSNR compared to those trained with pixel loss, they bring significant gains in perceptual quality \cite{CVPR2017PhotoLedig,ICCV2017EnhancenetSajjadi}.
As a matter of fact, the discriminator extracts some difficult-to-learn latent patterns of real HR images, and pushes the generated HR images to conform, thus helps to generate more realistic images.
However, currently the training process of GAN is still difficult and unstable.
Although there have been some studies on how to stabilize the GAN training \cite{ICML2017WassersteinArjovsky,NIPS2017ImprovedGulrajani,ICLR2018SpectralMiyato}, how to ensure that the GANs integrated into SR models are trained correctly and play an active role remains a problem.

\textbf{Cycle Consistency Loss.}
Motivated by the CycleGAN proposed by Zhu \textit{et al.} \cite{ICCV2017UnpairedZhu}, Yuan \textit{et al.} \cite{CVPRW2018UnsupervisedYuan} present a cycle-in-cycle approach for super-resolution.  
Concretely speaking, they not only super-resolve the LR image $I$ to the HR image $\hat{I}$ but also downsample $\hat{I}$ back to another LR image $I'$ through another CNN.
The regenerated $I'$ is required to be identical to the input $I$, thus the cycle consistency loss is introduced for constraining their pixel-level consistency:
\begin{equation}
    \mathcal{L}_{\text{cycle}} (I', I) = \frac{1}{hwc} \sqrt{\sum_{i,j,k} (I'_{i,j,k} - I_{i,j,k})^2} .
\end{equation}


\textbf{Total Variation Loss.}
In order to suppress noise in generated images, the total variation (TV) loss \cite{PDNP1992NonlinearRudin} is introduced into SR by Aly \textit{et al.} \cite{TIP2005ImageAly}.
It is defined as the sum of the absolute differences between neighboring pixels and measures how much noise is in the images, as follows:
\begin{equation}
    \mathcal{L}_{\text{TV}} (\hat{I}) = \frac{1}{hwc} \sum_{i,j,k} \sqrt{
        (\hat{I}_{i,j+1,k} - \hat{I}_{i,j,k})^2 +
        (\hat{I}_{i+1,j,k} - \hat{I}_{i,j,k})^2} .
\end{equation}
Lai \textit{et al.} \cite{CVPR2017PhotoLedig} and Yuan \textit{et al.} \cite{CVPRW2018UnsupervisedYuan} also adopt the TV loss for imposing spatial smoothness.

\textbf{Prior-Based Loss.}
In addition to the above loss functions, external prior knowledge is also introduced to constrain the generation.
Specifically, Bulat \textit{et al.} \cite{CVPR2018SuperBulat} focus on face image SR and introduce a face alignment network (FAN) to constrain the consistency of facial landmarks.  
The FAN is pre-trained and integrated for providing face alignment priors, and then trained jointly with the SR.
In this way, the proposed Super-FAN improves performance both on LR face alignment and face image SR.

As a matter of fact, the content loss and the texture loss, both of which introduce a classification network, essentially provide prior knowledge of hierarchical image features for SR.
By introducing more prior knowledge, the SR performance can be further improved.

In this section, we introduce various loss functions for SR.
In practice, researchers often combine multiple loss functions by weighted average \cite{CVPR2017DeepLai,CVPR2017PhotoLedig,ICCV2017EnhancenetSajjadi,Arxiv2018DualGuo,CVPR2018RecoveringWang} for constraining different aspects of the generation process, especially for distortion-perception tradeoff \cite{CVPR2017PhotoLedig,ECCVW2018AnalyzingVasu,ECCVW2018ESRGANWang,ECCVW2018GenerativeCheon,ECCVW2018DeepChoi}.
However, the weights of different loss functions require a lot of empirical exploration, and how to combine reasonably and effectively remains a problem.

\subsubsection{Batch Normalization}

In order to accelerate and stabilize training of deep CNNs, Sergey \textit{et al.} \cite{ICML2015BatchSergey} propose batch normalization (BN) to reduce internal covariate shift of networks.
Specifically, they perform normalization for each mini-batch and train two extra transformation parameters for each channel to preserve the representation ability.
Since the BN calibrates the intermediate feature distribution and mitigates vanishing gradients, it allows using higher learning rates and being less careful about initialization.
Thus this technique is widely used by SR models \cite{CVPR2017ImageTai,CVPR2017PhotoLedig,ICCV2017MemNetTai,ICLR2017AmortisedSonderby,CVPR2018LearningZhang,CVPRW2018MultiLiu}.

However, Lim \textit{et al.} \cite{CVPRW2017EnhancedLim} argue that the BN loses the scale information of each image and gets rid of range flexibility from networks.
So they remove BN and use the saved memory cost (up to 40\%) to develop a much larger model, and thus increase the performance substantially.
Some other models \cite{CVPRW2018FullyWang,CVPRW2018PersistentChen,ECCVW2018ESRGANWang} also adopt this experience and achieve performance improvements.

\subsubsection{Curriculum Learning}
\label{sec_curriculum_learning}

Curriculum learning \cite{ICML2009CurriculumBengio} refers to starting from an easier task and gradually increasing the difficulty.
Since super-resolution is an ill-posed problem and always suffers adverse conditions such as large scaling factors, noise and blurring, the curriculum training is incorporated for reducing learning difficulty.

In order to reduce the difficulty of SR with large scaling factors, Wang \textit{et al.} \cite{CVPRW2018FullyWang}, Bei \textit{et al.} \cite{CVPRW2018NewBei} and Ahn \textit{et al.} \cite{CVPRW2018ImageAhn} propose ProSR, ADRSR and progressive CARN, respectively, which are progressive not only on architectures (Sec. \ref{sec_framework_progressive}) but also on training procedure.
The training starts with the $2\times$ upsampling, and after finishing training, the portions with $4\times$ or larger scaling factors are gradually mounted and blended with the previous portions.
Specifically, the ProSR blends by linearly combining the output of this level and the upsampled output of previous levels following \cite{ICLR2018ProgressiveKarras}, the ADRSR concatenates them and attaches another convolutional layer, while the progressive CARN replace the previous reconstruction block with the one that produces the image in double resolution.


In addition, Park \textit{et al.} \cite{CVPRW2018EfficientPark} divide the $8\times$ SR problem to three sub-problems (i.e., $1\times$ to $2\times$, $2\times$ to $4\times$, $4\times$ to $8\times$) and train independent networks for each problem.
Then two of them are concatenated and fine-tuned, and then with the third one.
Besides, they also decompose the $4\times$ SR under difficult conditions into $1\times$ to $2\times$, $2\times$ to $4\times$ and denoising or deblurring sub-problems.
In contrast, the SRFBN \cite{CVPR2019FeedbackLi} uses this strategy for SR under adverse conditions, i.e., starting from easy degradation and gradually increasing degradation complexity.

Compared to common training procedure, the curriculum learning greatly reduces the training difficulty and shortens the total training time, especially for large factors.

\subsubsection{Multi-supervision}
\label{sec_multi_supervision}

Multi-supervision refers to adding multiple supervision signals within the model for enhancing the gradient propagation and avoiding vanishing and exploding gradients.
In order to prevent the gradient problems introduced by recursive learning (Sec. \ref{sec_recursive_learning}), the DRCN \cite{CVPR2016DeeplyKim} incorporates multi-supervision with recursive units.
Specifically, they feed each output of recursive units into a reconstruction module to generate an HR image, and build the final prediction by incorporating all the intermediate reconstructions.  
Similar strategies are also taken by MemNet \cite{ICCV2017MemNetTai} and DSRN \cite{CVPR2018ImageHan}, which are also based on recursive learning.

Besides, since the LapSRN \cite{CVPR2017DeepLai,TPAMI2018FastLai} under the progressive upsampling framework (Sec. \ref{sec_framework_progressive}) generates intermediate results of different scales during propagation, it is straightforward to adopt multi-supervision strategy.
Specifically, the intermediate results are forced to be the same as the intermediate images downsampled from the ground truth HR images.

In practice, this multi-supervision technique is often implemented by adding some terms in the loss function, and in this way, the supervision signals are back-propagated more effectively, and thus reduce the training difficulty and enhance the model training.

\subsection{Other Improvements}
\label{sec_other_improvements}

In addition to the network design and learning strategies, there are other techniques further improving SR models.

\subsubsection{Context-wise Network Fusion}

Context-wise network fusion (CNF) \cite{CVPRW2017ImageRen} refers to a stacking technique fusing predictions from multiple SR networks (i.e., a special case of multi-path learning in Sec. \ref{sec_multi_path_learning}).
To be concrete, they train individual SR models with different architectures separately, feed the prediction of each model into individual convolutional layers, and finally sum the outputs up to be the final prediction result.
Within this CNF framework, the final model constructed by three lightweight SRCNNs \cite{ECCV2014LearningDong,TPAMI2016ImageDong} achieves comparable performance with state-of-the-art models with acceptable efficiency \cite{CVPRW2017ImageRen}.

\subsubsection{Data Augmentation}

Data augmentation is one of the most widely used techniques for boosting performance with deep learning.
For image super-resolution, some useful augmentation options include cropping, flipping, scaling, rotation, color jittering, etc. \cite{CVPR2016SevenTimofte,CVPRW2017EnhancedLim,CVPR2017DeepLai,CVPR2017ImageTai,CVPR2018ImageHan,CVPR2018FastHui}.
In addition, Bei \textit{et al.} \cite{CVPRW2018NewBei} also randomly shuffle RGB channels, which not only augments data, but also alleviates color bias caused by the dataset with color unbalance.

\subsubsection{Multi-task Learning}
\label{sec_mtl}

Multi-task learning \cite{ML1997MultitaskCaruana} refers to improving generalization ability by leveraging domain-specific information contained in training signals of related tasks, such as object detection and semantic segmentation \cite{ICCV2017MaskHe}, head pose estimation and facial attribute inference \cite{ECCV2014FicialZhang}.
In the SR field, Wang \textit{et al.} \cite{CVPR2018RecoveringWang} incorporate a semantic segmentation network for providing semantic knowledge and generating semantic-specific details.  
Specifically, they propose spatial feature transformation to take semantic maps as input and predict spatial-wise parameters of affine transformation performed on the intermediate feature maps.
The proposed SFT-GAN thus generates more realistic and visually pleasing textures on images with rich semantic regions.  
Besides, considering that directly super-resolving noisy images may cause noise amplification, the DNSR \cite{CVPRW2018NewBei} proposes to train a denoising network and an SR network separately, then concatenates them and fine-tunes together.
Similarly, the cycle-in-cycle GAN (CinCGAN) \cite{CVPRW2018UnsupervisedYuan} combines a cycle-in-cycle denoising framework and a cycle-in-cycle SR model to joint perform noise reduction and super-resolution.
Since different tasks tend to focus on different aspects of the data, combining related tasks with SR models usually improves the SR performance by providing extra information and knowledge.

\subsubsection{Network Interpolation}

PSNR-based models produce images closer to ground truths but introduce blurring problems, while GAN-based models bring better perceptual quality but introduce unpleasant artifacts (e.g., meaningless noise making images more ``realistic'').
In order to better balance the distortion and perception, Wang \textit{et al.} \cite{ECCVW2018ESRGANWang,CVPR2019DeepWang} propose a network interpolation strategy.
Specifically, they train a PSNR-based model and train a GAN-based model by fine-tuning, then interpolate all the corresponding parameters of both networks to derive intermediate models.
By tuning the interpolation weights without retraining networks, they produce meaningful results with much less artifacts.

\subsubsection{Self-Ensemble}
\label{sec_self_ensemble}

Self-ensemble, a.k.a. enhanced prediction \cite{CVPR2016SevenTimofte}, is an inference technique commonly used by SR models.
Specifically, rotations with different angles (0$^\circ$, 90$^\circ$, 180$^\circ$, 270$^\circ$) and horizontal flipping are applied on the LR images to get a set of 8 images.
Then these images are fed into the SR model and the corresponding inverse transformation is applied to the reconstructed HR images to get the outputs.
The final prediction result is conducted by the mean \cite{CVPR2016SevenTimofte,NIPS2016ImageMao,CVPRW2017EnhancedLim,CVPR2018ResidualZhang,ECCV2018ImageZhang,CVPRW2018FullyWang} or the median \cite{CVPR2018ZeroShocher} of these outputs.
In this way, these models further improve performance.

\begin{table*}[t]
    \renewcommand{\arraystretch}{1.3}
    \caption{
        Super-resolution methodology employed by some representative models.
        The ``Fw.'', ``Up.'', ``Rec.'', ``Res.'', ``Dense.'', ``Att.'' represent SR frameworks, upsampling methods, recursive learning, residual learning, dense connections, attention mechanism, respectively.
    }
    \label{tab_sota_sr_models}
    \rowcolors{2}{gray!0}{gray!8}
    \centering
    {
        \setlength{\fboxrule}{1pt}
        \begin{tabular}{|l|l|c|c|c|c|c|c|c|c|l|}
            \hline
            Method                                      & Publication & Fw.   & Up.       & Rec. & Res. & Dense& Att. & $\mathcal{L}_{\text{L1}}$ & $\mathcal{L}_{\text{L2}}$ & Keywords \\
            \hline
            SRCNN \cite{ECCV2014LearningDong}           & 2014, ECCV  & Pre.  & Bicubic   & ~    & ~    & ~    & ~    & ~    & \yes & ~ \\
            DRCN \cite{CVPR2016DeeplyKim}               & 2016, CVPR  & Pre.  & Bicubic   & \yes & \yes & ~    & ~    & ~    & \yes & Recursive layers \\
            FSRCNN \cite{ECCV2016AcceleratingDong}      & 2016, ECCV  & Post. & Deconv    & ~    & ~    & ~    & ~    & ~    & \yes & Lightweight design \\
            ESPCN \cite{CVPR2017RealCaballero}          & 2017, CVPR  & Pre.  & Sub-Pixel & ~    & ~    & ~    & ~    & ~    & \yes & Sub-pixel \\
            LapSRN \cite{CVPR2017DeepLai}               & 2017, CVPR  & Pro.  & Bicubic   & ~    & \yes & ~    & ~    & \yes & ~    & $\mathcal{L}_{\text{pixel\_Cha}}$ \\
            DRRN \cite{CVPR2017ImageTai}                & 2017, CVPR  & Pre.  & Bicubic   & \yes & \yes & ~    & ~    & ~    & \yes & Recursive blocks \\
            SRResNet \cite{CVPR2017PhotoLedig}          & 2017, CVPR  & Post. & Sub-Pixel & ~    & \yes & ~    & ~    & ~    & \yes & $\mathcal{L}_{\text{Con.}}$, $\mathcal{L}_{\text{TV}}$ \\
            SRGAN \cite{CVPR2017PhotoLedig}             & 2017, CVPR  & Post. & Sub-Pixel & ~    & \yes & ~    & ~    & ~    & ~    & $\mathcal{L}_{\text{Con.}}$, $\mathcal{L}_{\text{GAN}}$ \\
            EDSR \cite{CVPRW2017EnhancedLim}            & 2017, CVPRW & Post. & Sub-Pixel & ~    & \yes & ~    & ~    & \yes & ~    & Compact and large-size design \\
            EnhanceNet \cite{ICCV2017EnhancenetSajjadi} & 2017, ICCV  & Pre.  & Bicubic   & ~    & \yes & ~    & ~    & ~    & ~    & $\mathcal{L}_{\text{Con.}}$, $\mathcal{L}_{\text{GAN}}$, $\mathcal{L}_{\text{texture}}$ \\
            MemNet \cite{ICCV2017MemNetTai}             & 2017, ICCV  & Pre.  & Bicubic   & \yes & \yes & \yes & ~    & ~    & \yes & Memory block \\
            SRDenseNet \cite{ICCV2017ImageTong}         & 2017, ICCV  & Post. & Deconv    & ~    & \yes & \yes & ~    & ~    & \yes & Dense connections \\
            DBPN \cite{CVPR2018DeepHaris}               & 2018, CVPR  & Iter. & Deconv    & ~    & \yes & \yes & ~    & ~    & \yes & Back-projection \\
            DSRN \cite{CVPR2018ImageHan}                & 2018, CVPR  & Pre.  & Deconv    & \yes & \yes & ~    & ~    & ~    & \yes & Dual state \\
            RDN \cite{CVPR2018ResidualZhang}            & 2018, CVPR  & Post. & Sub-Pixel & ~    & \yes & \yes & ~    & \yes & ~    & Residual dense block \\
            CARN \cite{ECCV2018FastAhn}                 & 2018, ECCV  & Post. & Sub-Pixel & \yes & \yes & \yes & ~    & \yes & ~    & Cascading \\
            MSRN \cite{ECCV2018MultiLi}                 & 2018, ECCV  & Post. & Sub-Pixel & ~    & \yes & ~    & ~    & \yes & ~    & Multi-path \\
            RCAN \cite{ECCV2018ImageZhang}              & 2018, ECCV  & Post. & Sub-Pixel & ~    & \yes & ~    & \yes & \yes & ~    & Channel attention \\
            ESRGAN \cite{ECCVW2018ESRGANWang}           & 2018, ECCVW & Post. & Sub-Pixel & ~    & \yes & \yes &      & \yes & ~    & $\mathcal{L}_{\text{Con.}}$, $\mathcal{L}_{\text{GAN}}$ \\
            RNAN \cite{ICLR2019ResidualZhang}           & 2019, ICLR  & Post. & Sub-Pixel & ~    & \yes & ~    & \yes & \yes & ~    & Non-local attention \\
            Meta-RDN \cite{CVPR2019MetaHu}              & 2019, CVPR  & Post. & Meta Upscale & ~    & \yes & \yes & ~    & \yes & ~    & Magnification-arbitrary \\
            SAN \cite{CVPR2019SecondDai}                & 2019, CVPR  & Post. & Sub-Pixel & ~    & \yes & ~    & \yes & \yes & ~    & Second-order attention \\
            SRFBN \cite{CVPR2019FeedbackLi}             & 2019, CVPR  & Post. & Deconv & \yes & \yes & \yes & ~    & \yes & ~    & Feedback mechanism \\
            \hline
        \end{tabular}
    }
\end{table*}

\begin{figure*}
    \centering
    {
        \setlength{\fboxrule}{1pt}
        \includegraphics[width=0.95\linewidth]{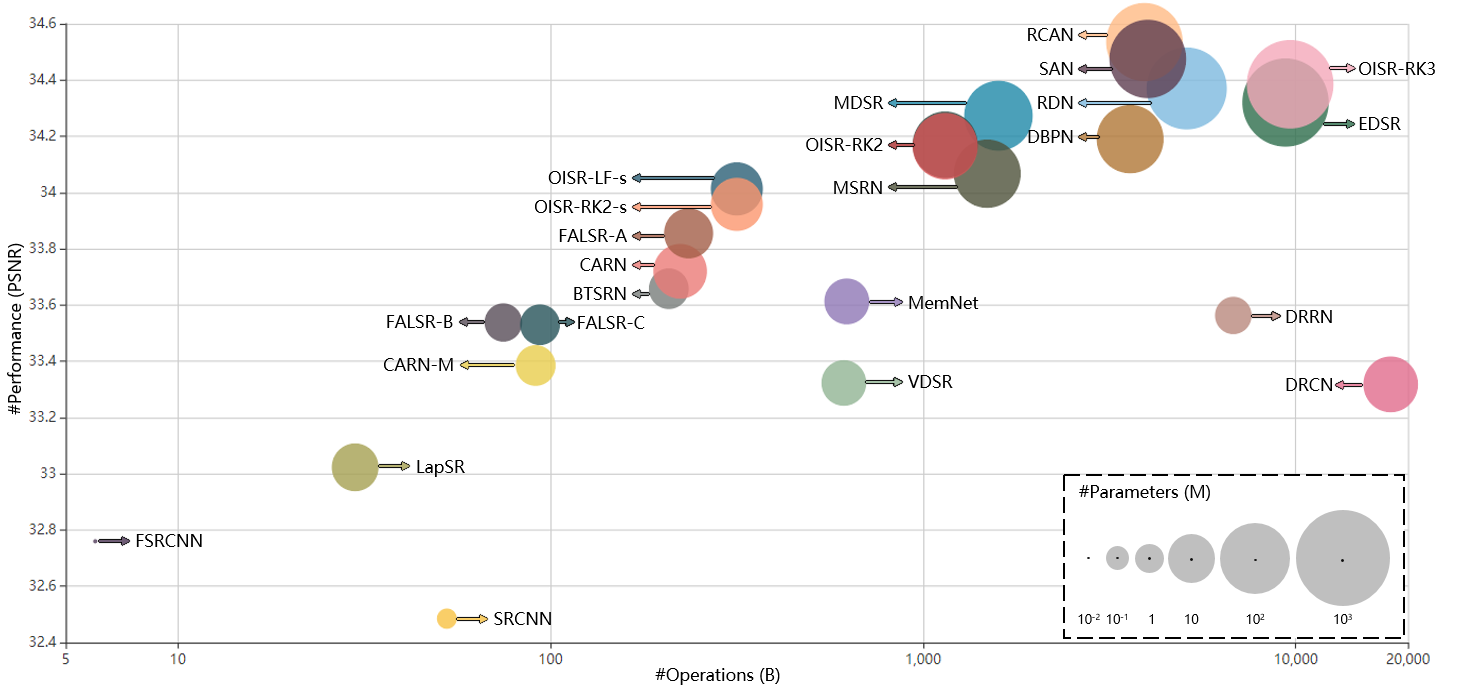}
    }
    \caption{
        Super-resolution benchmarking.
        The $x$-axis and the $y$-axis denote the Multi-Adds and PSNR, respectively, and the circle size represents the number of parameters.
    }
    \label{fig_sr_benchmarking}
\end{figure*}

\subsection{State-of-the-art Super-resolution Models}
\label{sec_sota_sr_models}

In recent years, image super-resolution models based on deep learning have received more and more attention and achieved state-of-the-art performance.
In previous sections, we decompose SR models into specific components, including model frameworks (Sec. \ref{sec_sr_frameworks}), upsampling methods (Sec. \ref{sec_upsampling_methods}), network design (Sec. \ref{sec_network_design}) and learning strategies (Sec. \ref{sec_learning_strategies}), analyze these components hierarchically and identify their advantages and limitations.
As a matter of fact, most of the state-of-the-art SR models today can basically be attributed to a combination of multiple strategies we summarize above.
For example, the biggest contribution of the RCAN \cite{ECCV2018ImageZhang} comes from the channel attention mechanism (Sec. \ref{sec_attention_mechanism}), and it also employs other strategies like sub-pixel upsampling (Sec. \ref{sec_learning_based_upsampling}), residual learning (Sec. \ref{sec_residual_learning}), pixel L1 loss (Sec. \ref{sec_loss}), and self-ensemble (Sec. \ref{sec_self_ensemble}).
In similar manners, we summarize some representative models and their key strategies, as Table \ref{tab_sota_sr_models} shows.

In addition to SR accuracy, the efficiency is another very important aspect and different strategies have more or less impact on efficiency.
So in the previous sections, we not only analyze the accuracy of the presented strategies, but also indicate the concrete impacts on efficiency for the ones with a greater impact on efficiency, such as the post-upsampling (Sec. \ref{sec_framework_post}), recursive learning (Sec. \ref{sec_recursive_learning}), dense connections (Sec. \ref{sec_dense_connections}), xUnit (Sec. \ref{sec_xunit}).
And we also benchmark some representative SR models on the SR accuracy (i.e., PSNR), model size (i.e., number of parameters) and computation cost (i.e., number of Multi-Adds), as shown in Fig. \ref{fig_sr_benchmarking}.
The accuracy is measured by the mean of the PSNR on 4 benchmark datasets (i.e., Set5 \cite{BMVC2012LowBevilacqua}, Set14 \cite{ICCS2010SingleZeyde}, B100 \cite{ICCV2001DatabaseMartin} and Urban100 \cite{CVPR2015SingleHuang}).
And the model size and computational cost are calculated with PyTorch-OpCounter \cite{Code2019OpCounterZhu}, where the output resolution is 720p (i.e., $1080 \times 720$).
All statistics are derived from the original papers or calculated on official models, with a scaling factor of 2.
For better viewing and comparison, we also provide an interactive online version\footnote{https://github.com/ptkin/Awesome-Super-Resolution}.

\section{Unsupervised Super-resolution}
\label{sec_unsupervised_sr}

Existing super-resolution works mostly focus on supervised learning, i.e., learning with matched LR-HR image pairs.
However, since it is difficult to collect images of the same scene but with different resolutions, the LR images in SR datasets are often obtained by performing predefined degradation on HR images.
Thus the trained SR models actually learn a reverse process of the predefined degradation.
In order to learn the real-world LR-HR mapping without introducing manual degradation priors, researchers pay more and more attention to unsupervised SR, in which case only unpaired LR-HR images are provided for training, so that the resulting models are more likely to cope with the SR problems in real-world scenarios.
Next we'll briefly introduce several existing unsupervised SR models with deep learning, and more methods are yet to be explored.

\subsection{Zero-shot Super-resolution}

Considering that the internal image statistics inside a single image have provided sufficient information for SR, Shocher \textit{et al.} \cite{CVPR2018ZeroShocher} propose zero-shot super-resolution (ZSSR) to cope with unsupervised SR by training image-specific SR networks at test time rather than training a generic model on large external datasets.
Specifically, they estimate the degradation kernel from a single image using \cite{ICCV2013NonparametricMichaeli} and use this kernel to build a small dataset by performing degradation with different scaling factors and augmentation on this image.
Then a small CNN for SR is trained on this dataset and used for the final prediction.

In this way, the ZSSR leverages on the cross-scale internal recurrence inside every image, and thus outperforms previous approaches by a large margin ($1$ dB for estimated kernels and $2$ dB for known kernels) on images under non-ideal conditions (i.e., images obtained by non-bicubic degradation and suffered effects like blurring, noise, compression artifacts), which is closer to the real-world scenes, while give competitive results under ideal conditions (i.e., images obtained by bicubic degradation).
However, since it needs to train different networks for different images during testing, the inference time is much longer than others.

\subsection{Weakly-supervised Super-resolution}

To cope with super-resolution without introducing predefined degradation, researchers attempt to learn SR models with weakly-supervised learning, i.e., using unpaired LR-HR images.
Among them, some researchers first learn the HR-to-LR degradation and use it to construct datasets for training the SR model, while others design cycle-in-cycle networks to learn the LR-to-HR and HR-to-LR mappings simultaneously.
Next we'll detail these models.

\textbf{Learned Degradation.}
Since the predefined degradation is suboptimal, learning the degradation from unpaired LR-HR datasets is a feasible direction.
Bulat \textit{et al.} \cite{ECCV2018LearnBulat} propose a two-stage process which firstly trains an HR-to-LR GAN to learn degradation using unpaired LR-HR images and then trains an LR-to-HR GAN for SR using paired LR-HR images conducted base on the first GAN.
Specifically, for the HR-to-LR GAN, HR images are fed into the generator to produce LR outputs, which are required to match not only the LR images obtained by downscaling the HR images (by average pooling) but also the distribution of real LR images.
After finishing training, the generator is used as a degradation model to generate LR-HR image pairs.
Then for the LR-to-HR GAN, the generator (i.e., the SR model) takes the generated LR images as input and predicts HR outputs, which are required to match not only the corresponding HR images but also the distribution of the HR images.

By applying this two-stage process, the proposed unsupervised model effectively increases the quality of super-resolving real-world LR images and obtains large improvement over previous state-of-the-art works.

\textbf{Cycle-in-cycle Super-resolution.}
Another approach for unsupervised super-resolution is to treat the LR space and the HR space as two domains, and use a cycle-in-cycle structure to learn the mappings between each other.
In this case, the training objectives include pushing the mapped results to match the target domain distribution and making the images recoverable through round-trip mappings.

Motivated by CycleGAN \cite{ICCV2017UnpairedZhu}, Yuan \textit{et al.} \cite{CVPRW2018UnsupervisedYuan} propose a cycle-in-cycle SR network (CinCGAN) composed of 4 generators and 2 discriminators, making up two CycleGANs for \textit{noisy LR} $\rightleftharpoons$ \textit{clean LR} and \textit{clean LR} $\rightleftharpoons$ \textit{clean HR} mappings, respectively.
Specifically, in the first CycleGAN, the noisy LR image is fed into a generator, and the output is required to be consistent with the distribution of real clean LR images.
Then it's fed into another generator and required to recover the original input.
Several loss functions (e.g., adversarial loss, cycle consistency loss, identity loss) are employed for guaranteeing the cycle consistency, distribution consistency, and mapping validity.
The other CycleGAN is similarly designed, except that the mapping domains are different.

Because of avoiding the predefined degradation, the unsupervised CinCGAN not only achieves comparable performance to supervised methods, but also is applicable to various cases even under very harsh conditions.
However, due to the ill-posed essence of SR problem and the complicated architecture of CinCGAN, some advanced strategies are needed for reducing the training difficulty and instability.

\subsection{Deep Image Prior}

Considering that the CNN structure is sufficient to capture a great deal of low-level image statistics prior for inverse problems, Ulyanov \textit{et al.} \cite{CVPR2018DeepUlyanov} employ a randomly-initialized CNN as handcrafted prior to perform SR.
Specifically, they define a generator network which takes a random vector $z$ as input and tries to generate the target HR image $I_y$.
The goal is to train the network to find an $\hat{I_y}$ that the downsampled $\hat{I_y}$ is identical to the LR image $I_x$.
Since the network is randomly initialized and never trained, the only prior is the CNN structure itself.
Although the performance of this method is still worse than the supervised methods ($2$ dB), it outperforms traditional bicubic upsampling considerably ($1$ dB).
Besides, it shows the rationality of the CNN architectures itself, and prompts us to improve SR by combining the deep learning methodology with handcrafted priors such as CNN structures or self-similarity.

\section{Domain-Specific Applications}
\label{sec_domain_specific_apps}

\subsection{Depth Map Super-resolution}

Depth maps record the depth (i.e., distance) between the viewpoint and objects in the scene, and plays important roles in many tasks like pose estimation \cite{CVPR2011RealShotton,CVPR2018V2VMoon} and semantic segmentation \cite{ECCV2014LearningGupta,ECCV2018DepthWang}.  
However, due to economic and production constraints, the depth maps produced by depth sensors are often low-resolution and suffer degradation effects such as noise, quantization and missing values.
Thus super-resolution is introduced for increasing the spatial resolution of depth maps.

Nowadays one of the most popular practices for depth map SR is to use another economical RGB camera to obtain HR images of the same scenes for guiding super-resolving the LR depth maps.
Specifically, Song \textit{et al.} \cite{ACCV2016DeepSong} exploit the depth field statistics and local correlations between depth maps and RGB images to constrain the global statistics and local structures.
Hui \textit{et al.} \cite{ECCV2016DepthHui} utilize two CNNs to simultaneously upsample LR depth maps and downsample HR RGB images, then use RGB features as the guidance for upsampling depth maps with the same resolution.
And Haefner \textit{et al.} \cite{CVPR2018FightHaefner} further exploit the color information and guide SR by resorting to the shape-from-shading technique.
In contrast, Riegler \textit{et al.} \cite{ECCV2016ATGVRiegler} combine CNNs with an energy minimization model in the form of a powerful variational model to recover HR depth maps without other reference images.

\subsection{Face Image Super-resolution}

Face image super-resolution, a.k.a. face hallucination (FH), can often help other face-related tasks \cite{TIP2008ExamplePark,CVPR2018FSRNetChen,ECCV2018SuperZhang}.  
Compared to generic images, face images have more face-related structured information, so incorporating facial prior knowledge (e.g., landmarks, parsing maps, identities) into FH is a very popular and promising approach.

One of the most straightforward way is to constrain the generated images to have the identical face-related attributes to ground truth.
Specifically, the CBN \cite{ECCV2016DeepZhu} utilizes the facial prior by alternately optimizing FH and dense correspondence field estimation.
The Super-FAN \cite{CVPR2018SuperBulat} and MTUN \cite{ECCV2018FaceYu} both introduce FAN to guarantee the consistency of facial landmarks by end-to-end multi-task learning.
And the FSRNet \cite{CVPR2018FSRNetChen} uses not only facial landmark heatmaps but also face parsing maps as prior constraints.
The SICNN \cite{ECCV2018SuperZhang}, which aims at recovering the real identity, adopts a super-identity loss function and a domain-integrated training approach to stable the joint training.

Besides explicitly using facial prior, the implicit methods are also widely studied.
The TDN \cite{AAAI2017FaceYu} incorporates spatial transformer networks \cite{NIPS2015SpatialJaderberg} for automatic spatial transformations and thus solves the face unalignment problem.
Based on TDN, the TDAE \cite{CVPR2017HallucinatingYu} adopts a decoder-encoder-decoder framework, where the first decoder learns to upsample and denoise, the encoder projects it back to aligned and noise-free LR faces, and the last decoder generates hallucinated HR images.
In contrast, the LCGE \cite{IJCAI2017LearningSong} employs component-specific CNNs to perform SR on five facial components, uses k-NN search on an HR facial component dataset to find corresponding patches, synthesizes finer-grained components and finally fuses them to FH results.
Similarly, Yang \textit{et al.} \cite{IJCV2018HallucinatingYang} decompose deblocked face images into facial components and background, use the component landmarks to retrieve adequate HR exemplars in external datasets, perform generic SR on the background, and finally fuse them to complete HR faces.

In addition, researchers also improve FH from other perspectives.
Motivated by the human attention shifting mechanism \cite{Nature2005OptimalNajemnik}, the Attention-FH \cite{CVPR2017AttentionCao} resorts to a recurrent policy network for sequentially discovering attended face patches and performing local enhancement, and thus fully exploits the global interdependency of face images.
The UR-DGN \cite{ECCV2016UltraYu} adopts a network similar to SRGAN \cite{CVPR2017PhotoLedig} with adversarial learning.
And Xu \textit{et al.} \cite{ICCV2017LearningXu} propose a multi-class GAN-based FH model composed of a generic generator and class-specific discriminators.
Both Lee \textit{et al.} \cite{CVPR2018AttributeLee} and Yu \textit{et al.} \cite{CVPR2018SuperYu} utilize additional facial attribute information to perform FH with the specified attributes, based on the conditional GAN \cite{Arxiv2014ConditionalMirza}.

\subsection{Hyperspectral Image Super-resolution}

Compared to panchromatic images (PANs, i.e., RGB images with 3 bands), hyperspectral images (HSIs) containing hundreds of bands provide abundant spectral features and help various vision tasks \cite{PI2013AdvancesFauvel,CVPR2016ExploitingFu,CVPRW2017AerialUzkent}. 
However, due to hardware limitations, collecting high-quality HSIs is much more difficult than PANs and the resolution is also lower.
Thus super-resolution is introduced into this field, and researchers tend to combine HR PANs and LR HSIs to predict HR HSIs.
Among them, Masi \textit{et al.} \cite{RS2016PansharpeningMasi} employ SRCNN \cite{ECCV2014LearningDong} and incorporate several maps of nonlinear radiometric indices for boosting performance.
Qu \textit{et al.} \cite{CVPR2018UnsupervisedQu} jointly train two encoder-decoder networks to perform SR on PANs and HSIs, respectively, and transfer the SR knowledge from PAN to HSI by sharing the decoder and applying constraints such as angle similarity loss and reconstruction loss.
Recently, Fu \textit{et al.} \cite{CVPR2019HyperspectralFu} evaluate the effect of camera spectral response (CSR) functions for HSI SR and propose a CSR optimization layer which can automatically select or design the optimal CSR, and outperform the state-of-the-arts.

\subsection{Real-world Image Super-resolution}

Generally, the LR images for training SR models are generated by downsampling RGB images manually (e.g., by bicubic downsampling).
However, real-world cameras actually capture 12-bit or 14-bit RAW images, and performs a series of operations (e.g., demosaicing, denoising and compression) through camera ISPs (image signal processors) and finally produce 8-bit RGB images.
Through this process, the RGB images have lost lots of original signals and are very different from the original images taken by the camera.
Therefore, it is suboptimal to directly use the manually downsampled RGB image for SR.

To solve this problem, researchers study how to use real-world images for SR.
Among them, Chen \textit{et al.} \cite{CVPR2019CameraChen} analyze the relationships between image resolution (R) and field-of-view (V) in imaging systems (namely R-V degradation), propose data acquisition strategies to conduct a real-world dataset City100, and experimentally demonstrate the superiority of the proposed image synthesis model.
Zhang \textit{et al.} \cite{CVPR2019ZoomZhang} build another real-world image dataset SR-RAW (i.e., paired HR RAW images and LR RGB images) through optical zoom of cameras, and propose contextual bilateral loss to solve the misalignment problem.  
In contrast, Xu \textit{et al.} \cite{CVPR2019TowardsXu} propose a pipeline to generate realistic training data by simulating the imaging process and develop a dual CNN to exploit the originally captured radiance information in RAW images.
They also propose to learn a spatially-variant color transformation for effective color corrections and generalization to other sensors.

\subsection{Video Super-resolution}

For video super-resolution, multiple frames provide much more scene information, and there are not only intra-frame spatial dependency but also inter-frame temporal dependency (e.g., motions, brightness and color changes).
Thus the existing works mainly focus on making better use of spatio-temporal dependency, including explicit motion compensation (e.g., optical flow-based, learning-based) and recurrent methods, etc.

Among the optical flow-based methods, Liao \textit{et al.} \cite{ICCV2015VideoLiao} employ optical flow methods to generate HR candidates and ensemble them by CNNs.
VSRnet \cite{TCIVideoKappeler} and CVSRnet \cite{ICIP2016SuperKappeler} deal with motion compensation by Druleas algorithm \cite{ITSC2011TotalDrulea}, and uses CNNs to take successive frames as input and predict HR frames.
While Liu \textit{et al.} \cite{ICCV2017RobustLiu,TIP2018LearningLiu} perform rectified optical flow alignment, and propose a temporal adaptive net to generate HR frames in various temporal scales and aggregate them adaptively.

Besides, others also try to directly learn the motion compensation.
The VESPCN \cite{CVPR2017RealCaballero} utilizes a trainable spatial transformer \cite{NIPS2015SpatialJaderberg} to learn motion compensation based on adjacent frames, and enters multiple frames into a spatio-temporal ESPCN \cite{CVPR2016RealShi} for end-to-end prediction.
And Tao \textit{et al.} \cite{ICCV2017DetailTao} root from accurate LR imaging model and propose a sub-pixel-like module to simultaneously achieve motion compensation and super-resolution, and thus fuse the aligned frames more effectively.

Another trend is to use recurrent methods to capture the spatial-temporal dependency without explicit motion compensation.
Specifically, the BRCN \cite{NIPS2015BidirectionalHuang,TPAMI2018VideoHuang} employs a bidirectional framework, and uses CNN, RNN, and conditional CNN to model the spatial, temporal and spatial-temporal dependency, respectively.
Similarly, STCN \cite{AAAI2017BuildingGuo} uses a deep CNN and a bidirectional LSTM \cite{ICANN2005BidirectionalGraves} to extract spatial and temporal information.
And FRVSR \cite{CVPR2018FrameSajjadi} uses previously inferred HR estimates to reconstruct the subsequent HR frames by two deep CNNs in a recurrent manner.
Recently the FSTRN \cite{CVPR2019FastLi} employs two much smaller 3D convolution filters to replace the original large filter, and thus enhances the performance through deeper CNNs while maintaining low computational cost.
While the RBPN \cite{CVPR2019RecurrentHaris} extracts spatial and temporal contexts by a recurrent encoder-decoder, and combines them with an iterative refinement framework based on the back-projection mechanism (Sec. \ref{sec_framework_up_down}).

In addition, the FAST \cite{CVPRW2017FASTZhang} exploits compact descriptions of the structure and pixel correlations extracted by compression algorithms, transfers the SR results from one frame to adjacent frames, and much accelerates the state-of-the-art SR algorithms with little performance loss.  
And Jo \textit{et al.} \cite{CVPR2018DeepJo} generate dynamic upsampling filters and the HR residual image based on the local spatio-temporal neighborhoods of each pixel, and also avoid explicit motion compensation.

\subsection{Other Applications}

Deep learning based super-resolution is also adopted to other domain-specific applications and shows great performance.
Specifically, the Perceptual GAN \cite{CVPR2017PerceptualLi} addresses the small object detection problem by super-resolving representations of small objects to have similar characteristics as large objects and be more discriminative for detection.
Similarly, the FSR-GAN \cite{CVPR2018FeatureTan} super-resolves small-size images in the feature space instead of the pixel space, and thus transforms the raw poor features to highly discriminative ones, which greatly benefits image retrieval.
Besides, Jeon \textit{et al.} \cite{CVPR2018EnhancingJeon} utilize a parallax prior in stereo images to reconstruct HR images with sub-pixel accuracy in registration.
Wang \textit{et al.} \cite{CVPR2019LearningWang} propose a parallax-attention model to tackle the stereo image super-resolution problem.
Li \textit{et al.} \cite{CVPR20193DLi} incorporate the 3D geometric information and super-resolve 3D object texture maps.
And Zhang \textit{et al.} \cite{CVPR2019ResidualZhang} separate view images in one light field into groups, learn inherent mapping for every group and finally combine the residuals in every group to reconstruct higher-resolution light fields.
All in all, super-resolution technology can play an important role in all kinds of applications, especially when we can deal with large objects well but cannot handle small objects.

\section{Conclusion and Future Directions}

In this paper, we have given an extensive survey on recent advances in image super-resolution with deep learning.
We mainly discussed the improvement of supervised and unsupervised SR, and also introduced some domain-specific applications.
Despite great success, there are still many unsolved problems.
Thus in this section, we will point out these problems explicitly and introduce some promising trends for future evolution.
We hope that this survey not only provides a better understanding of image SR for researchers but also facilitates future research activities and application developments in this field.


\subsection{Network Design}

Good network design not only determines a hypothesis space with great performance upper bound, but also helps efficiently learn representations without excessive spatial and computational redundancy.
Below we will introduce some promising directions for network improvements.

\textit{Combining Local and Global Information.}
Large receptive field provides more contextual information and helps generate more realistic results.
Thus it is promising to combine local and global information for providing contextual information of different scales for image SR.

\textit{Combining Low- and High-level Information.}
Shallow layers in CNNs tend to extract low-level features like colors and edges, while deeper layers learn higher-level representations like object identities.
Thus combining low-level details with high-level semantics can be of great help for HR reconstruction.

\textit{Context-specific Attention.}
In different contexts, people tend to care about different aspects of the images.
For example, for the grass area people may be more concerned with local colors and textures, while in the animal body area people may care more about the species and corresponding hair details.
Therefore, incorporating attention mechanism to enhance the attention to key features facilitates the generation of realistic details.  

\textit{More Efficient Architectures.}
Existing SR modes tend to pursue ultimate performance, while ignoring the model size and inference speed.
For example, the EDSR \cite{CVPRW2017EnhancedLim} takes 20s per image for $4\times$ SR on DIV2K \cite{CVPRW2017NTIREAgustsson} with a Titan GTX GPU \cite{CVPRW2017NTIRETimofte}, and DBPN \cite{CVPR2018DeepHaris} takes 35s for $8\times$ SR \cite{CVPRW2018NTIREAncuti}.
Such long prediction time is unacceptable in practical applications, thus more efficient architectures are imperative.
How to reduce model sizes and speed up prediction while maintaining performance remains a problem.


\textit{Upsampling Methods.}
Existing upsampling methods (Sec. \ref{sec_upsampling_methods}) have more or less disadvantages:
interpolation methods result in expensive computation and cannot be end-to-end learned, the transposed convolution produces checkerboard artifacts, the sub-pixel layer brings uneven distribution of receptive fields, and the meta upscale module may cause instability or inefficiency and have further room for improvement.
How to perform effective and efficient upsampling still needs to be studied, especially with high scaling factors.

Recently, the neural architecture search (NAS) technique for deep learning has become more and more popular, greatly improving the performance or efficiency with little artificial intervention \cite{ICML2018EfficientPham,ICLR2019DartsLiu,NIPS2019NATGuo}.
For the SR field, combining the exploration of the above directions with NAS is of great potential.

\subsection{Learning Strategies}

Besides good hypothesis spaces, robust learning strategies are also needed for achieving satisfactory results.
Next we'll introduce some promising directions of learning strategies.

\textit{Loss Functions.}
Existing loss functions can be regarded as establishing constraints among LR/HR/SR images, and guide optimization based on whether these constraints are met.
In practice, these loss functions are often weighted combined and the best loss function for SR is still unclear.
Therefore, one of the most promising directions is to explore the potential correlations between these images and seek more accurate loss functions.

\textit{Normalization.}
Although BN is widely used in vision tasks, which greatly speeds up training and improves performance, it is proven to be sub-optimal for super-resolution \cite{CVPRW2017EnhancedLim,CVPRW2018FullyWang,CVPRW2018PersistentChen}.
Thus other effective normalization techniques for SR are needed to be studied.

\subsection{Evaluation Metrics}

Evaluation metrics are one of the most fundamental components for machine learning.
If the performance cannot be measured accurately, researchers will have great difficulty verifying improvements.
Metrics for super-resolution face such challenges and need more exploration.

\textit{More Accurate Metrics.}
Nowadays the PSNR and SSIM have been the most widely used metrics for SR.
However, the PSNR tends to result in excessive smoothness and the results can vary wildly between almost indistinguishable images.
The SSIM \cite{TIP2004ImageWang} performs evaluation in terms of brightness, contrast and structure, but still cannot measure perceptual quality accurately \cite{CVPR2017PhotoLedig,ICCV2017EnhancenetSajjadi}.
Besides, the MOS is the closest to human visual response, but needs to take a lot of efforts and is non-reproducible.
Although researchers have proposed various metrics (Sec. \ref{sec_iqa}), but currently there is no unified and admitted evaluation metrics for SR quality.
Thus more accurate metrics for evaluating reconstruction quality are urgently needed.

\textit{Blind IQA Methods.}
Today most metrics used for SR are all-reference methods, i.e., assuming that we have paired LR-HR images with perfect quality.
But since it's difficult to obtain such datasets, the commonly used datasets for evaluation are often conducted by manual degradation.
In this case, the task we perform evaluation on is actually the inverse process of the predefined degradation.
Therefore, developing blind IQA methods also has great demands.



\subsection{Unsupervised Super-resolution}

As mentioned in Sec. \ref{sec_unsupervised_sr}, it is often difficult to collect images with different resolutions of the same scene, so bicubic interpolation is widely used for constructing SR datasets.
However, the SR models trained on these datasets may only learn the inverse process of the predefined degradation.
Therefore, how to perform unsupervised super-resolution (i.e., trained on datasets without paired LR-HR images) is a promising direction for future development.

\subsection{Towards Real-world Scenarios}

Image super-resolution is greatly limited in real-world scenarios, such as suffering unknown degradation, missing paired LR-HR images.
Below we'll introduce some directions towards real-world scenarios.

\textit{Dealing with Various Degradation.}
Real-world images tend to suffer degradation like blurring, additive noise and compression artifacts.
Thus the models trained on datasets conducted manually often perform poorly in real-world scenes.
Some works have been proposed for solving this \cite{CVPR2018LearningZhang,ECCV2018LearnBulat,CVPRW2018NewBei,CVPRW2018UnsupervisedYuan}, but these methods have some inherent drawbacks, such as great training difficulty and over-perfect assumptions.
This issue is urgently needed to be resolved.

\textit{Domain-specific Applications.}
Super-resolution can not only be used in domain-specific data and scenes directly, but also help other vision tasks greatly (Sec. \ref{sec_domain_specific_apps}).
Therefore, it is also a promising direction to apply SR to more specific domains, such as video surveillance, object tracking, medical imaging and scene rendering.


\appendices

\ifCLASSOPTIONcaptionsoff
    \newpage
\fi

\section*{Acknowledgment}
Prof. Jian Chen is supported by the Guangdong special branch plans young talent with scientific and technological innovation (Grant No. 2016TQ03X445), the Guangzhou science and technology planning project (Grant No. 201904010197) and Natural Science Foundation of Guangdong Province, China (2016A030313437).

\bibliographystyle{IEEEtran}
\bibliography{reference}

\begin{thebibliography}{100}
\providecommand{\url}[1]{#1}
\csname url@samestyle\endcsname
\providecommand{\newblock}{\relax}
\providecommand{\bibinfo}[2]{#2}
\providecommand{\BIBentrySTDinterwordspacing}{\spaceskip=0pt\relax}
\providecommand{\BIBentryALTinterwordstretchfactor}{4}
\providecommand{\BIBentryALTinterwordspacing}{\spaceskip=\fontdimen2\font plus
\BIBentryALTinterwordstretchfactor\fontdimen3\font minus
  \fontdimen4\font\relax}
\providecommand{\BIBforeignlanguage}[2]{{%
\expandafter\ifx\csname l@#1\endcsname\relax
\typeout{** WARNING: IEEEtran.bst: No hyphenation pattern has been}%
\typeout{** loaded for the language `#1'. Using the pattern for}%
\typeout{** the default language instead.}%
\else
\language=\csname l@#1\endcsname
\fi
#2}}
\providecommand{\BIBdecl}{\relax}
\BIBdecl

\bibitem{TCJ2008SuperGreenspan}
H.~Greenspan, ``Super-resolution in medical imaging,'' \emph{The Computer
  Journal}, vol.~52, 2008.

\bibitem{ICTSD2015SuperIsaac}
J.~S. Isaac and R.~Kulkarni, ``Super resolution techniques for medical image
  processing,'' in \emph{ICTSD}, 2015.

\bibitem{CVPR2017SimultaneousHuang}
Y.~Huang, L.~Shao, and A.~F. Frangi, ``Simultaneous super-resolution and
  cross-modality synthesis of 3d medical images using weakly-supervised joint
  convolutional sparse coding,'' in \emph{CVPR}, 2017.

\bibitem{ESP2010SuperZhang}
L.~Zhang, H.~Zhang, H.~Shen, and P.~Li, ``A super-resolution reconstruction
  algorithm for surveillance images,'' \emph{Elsevier Signal Processing},
  vol.~90, 2010.

\bibitem{AMDO2016ConvolutionalRasti}
P.~Rasti, T.~Uiboupin, S.~Escalera, and G.~Anbarjafari, ``Convolutional neural
  network super resolution for face recognition in surveillance monitoring,''
  in \emph{AMDO}, 2016.

\bibitem{WACV2016ImageDai}
D.~Dai, Y.~Wang, Y.~Chen, and L.~Van~Gool, ``Is image super-resolution helpful
  for other vision tasks?'' in \emph{WACV}, 2016.

\bibitem{Arxiv2018TaskHaris}
M.~Haris, G.~Shakhnarovich, and N.~Ukita, ``Task-driven super resolution:
  Object detection in low-resolution images,'' \emph{Arxiv:1803.11316}, 2018.

\bibitem{ICCV2017EnhancenetSajjadi}
M.~S. Sajjadi, B.~Sch{\"o}lkopf, and M.~Hirsch, ``Enhancenet: Single image
  super-resolution through automated texture synthesis,'' in \emph{ICCV}, 2017.

\bibitem{ECCV2018SODZhang}
Y.~Zhang, Y.~Bai, M.~Ding, and B.~Ghanem, ``Sod-mtgan: Small object detection
  via multi-task generative adversarial network,'' in \emph{ECCV}, 2018.

\bibitem{TASSP1981CubicKeys}
R.~Keys, ``Cubic convolution interpolation for digital image processing,''
  \emph{IEEE Transactions on Acoustics, Speech, and Signal Processing},
  vol.~29, 1981.

\bibitem{JAM1979LanczosDuchon}
C.~E. Duchon, ``Lanczos filtering in one and two dimensions,'' \emph{Journal of
  Applied Meteorology}, vol.~18, 1979.

\bibitem{CVGIP1991ImprovingIrani}
M.~Irani and S.~Peleg, ``Improving resolution by image registration,''
  \emph{CVGIP: Graphical Models and Image Processing}, vol.~53, 1991.

\bibitem{TOG2011ImageFreedman}
G.~Freedman and R.~Fattal, ``Image and video upscaling from local
  self-examples,'' \emph{TOG}, vol.~30, 2011.

\bibitem{CVPR2008ImageSun}
J.~Sun, Z.~Xu, and H.-Y. Shum, ``Image super-resolution using gradient profile
  prior,'' in \emph{CVPR}, 2008.

\bibitem{TPAMI2010SingleKim}
K.~I. Kim and Y.~Kwon, ``Single-image super-resolution using sparse regression
  and natural image prior,'' \emph{TPAMI}, vol.~32, 2010.

\bibitem{TIP2010RobustXiong}
Z.~Xiong, X.~Sun, and F.~Wu, ``Robust web image/video super-resolution,''
  \emph{IEEE Transactions on Image Processing}, vol.~19, 2010.

\bibitem{CGA2002ExampleFreeman}
W.~T. Freeman, T.~R. Jones, and E.~C. Pasztor, ``Example-based
  super-resolution,'' \emph{IEEE Computer Graphics and Applications}, vol.~22,
  2002.

\bibitem{CVPR2004SuperChang}
H.~Chang, D.-Y. Yeung, and Y.~Xiong, ``Super-resolution through neighbor
  embedding,'' in \emph{CVPR}, 2004.

\bibitem{ICCV2009SuperGlasner}
D.~Glasner, S.~Bagon, and M.~Irani, ``Super-resolution from a single image,''
  in \emph{ICCV}, 2009.

\bibitem{CVPR2008ImageJianchao}
Y.~Jianchao, J.~Wright, T.~Huang, and Y.~Ma, ``Image super-resolution as sparse
  representation of raw image patches,'' in \emph{CVPR}, 2008.

\bibitem{TIP2010ImageYang}
J.~Yang, J.~Wright, T.~S. Huang, and Y.~Ma, ``Image super-resolution via sparse
  representation,'' \emph{IEEE Transactions on Image Processing}, vol.~19,
  2010.

\bibitem{ECCV2014LearningDong}
C.~Dong, C.~C. Loy, K.~He, and X.~Tang, ``Learning a deep convolutional network
  for image super-resolution,'' in \emph{ECCV}, 2014.

\bibitem{TPAMI2016ImageDong}
------, ``Image super-resolution using deep convolutional networks,''
  \emph{TPAMI}, vol.~38, 2016.

\bibitem{NIPS2014GenerativeGoodfellow}
I.~Goodfellow, J.~Pouget-Abadie, M.~Mirza, B.~Xu, D.~Warde-Farley, S.~Ozair,
  A.~Courville, and Y.~Bengio, ``Generative adversarial nets,'' in \emph{NIPS},
  2014.

\bibitem{CVPR2017PhotoLedig}
C.~Ledig, L.~Theis, F.~Husz{\'a}r, J.~Caballero, A.~Cunningham, A.~Acosta,
  A.~P. Aitken, A.~Tejani, J.~Totz, Z.~Wang \emph{et~al.}, ``Photo-realistic
  single image super-resolution using a generative adversarial network,'' in
  \emph{CVPR}, 2017.

\bibitem{CVPR2016AccurateKim}
J.~Kim, J.~Kwon~Lee, and K.~Mu~Lee, ``Accurate image super-resolution using
  very deep convolutional networks,'' in \emph{CVPR}, 2016.

\bibitem{CVPR2017DeepLai}
W.-S. Lai, J.-B. Huang, N.~Ahuja, and M.-H. Yang, ``Deep laplacian pyramid
  networks for fast and accurate superresolution,'' in \emph{CVPR}, 2017.

\bibitem{ECCV2018FastAhn}
N.~Ahn, B.~Kang, and K.-A. Sohn, ``Fast, accurate, and lightweight
  super-resolution with cascading residual network,'' in \emph{ECCV}, 2018.

\bibitem{ECCV2016PerceptualJohnson}
J.~Johnson, A.~Alahi, and L.~Fei-Fei, ``Perceptual losses for real-time style
  transfer and super-resolution,'' in \emph{ECCV}, 2016.

\bibitem{CVPR2018SuperBulat}
A.~Bulat and G.~Tzimiropoulos, ``Super-fan: Integrated facial landmark
  localization and super-resolution of real-world low resolution faces in
  arbitrary poses with gans,'' in \emph{CVPR}, 2018.

\bibitem{CVPRW2017EnhancedLim}
B.~Lim, S.~Son, H.~Kim, S.~Nah, and K.~M. Lee, ``Enhanced deep residual
  networks for single image super-resolution,'' in \emph{CVPRW}, 2017.

\bibitem{CVPRW2018FullyWang}
Y.~Wang, F.~Perazzi, B.~McWilliams, A.~Sorkine-Hornung, O.~Sorkine-Hornung, and
  C.~Schroers, ``A fully progressive approach to single-image
  super-resolution,'' in \emph{CVPRW}, 2018.

\bibitem{ISPM2003SuperPark}
S.~C. Park, M.~K. Park, and M.~G. Kang, ``Super-resolution image
  reconstruction: A technical overview,'' \emph{IEEE Signal Processing
  Magazine}, vol.~20, 2003.

\bibitem{MVA2014SuperNasrollahi}
K.~Nasrollahi and T.~B. Moeslund, ``Super-resolution: A comprehensive survey,''
  \emph{Machine Vision and Applications}, vol.~25, 2014.

\bibitem{SIVP2011SurveyTian}
J.~Tian and K.-K. Ma, ``A survey on super-resolution imaging,'' \emph{Signal,
  Image and Video Processing}, vol.~5, 2011.

\bibitem{IVC2006ImageVan}
J.~Van~Ouwerkerk, ``Image super-resolution survey,'' \emph{Image and Vision
  Computing}, vol.~24, 2006.

\bibitem{ECCV2014SingleYang}
C.-Y. Yang, C.~Ma, and M.-H. Yang, ``Single-image super-resolution: A
  benchmark,'' in \emph{ECCV}, 2014.

\bibitem{CEE2016PerformanceThapa}
D.~Thapa, K.~Raahemifar, W.~R. Bobier, and V.~Lakshminarayanan, ``A performance
  comparison among different super-resolution techniques,'' \emph{Computers \&
  Electrical Engineering}, vol.~54, 2016.

\bibitem{CVPR2018LearningZhang}
K.~Zhang, W.~Zuo, and L.~Zhang, ``Learning a single convolutional
  super-resolution network for multiple degradations,'' in \emph{CVPR}, 2018.

\bibitem{ICCV2001DatabaseMartin}
D.~Martin, C.~Fowlkes, D.~Tal, and J.~Malik, ``A database of human segmented
  natural images and its application to evaluating segmentation algorithms and
  measuring ecological statistics,'' in \emph{ICCV}, 2001.

\bibitem{TPAMI2011ContourArbelaez}
P.~Arbelaez, M.~Maire, C.~Fowlkes, and J.~Malik, ``Contour detection and
  hierarchical image segmentation,'' \emph{TPAMI}, vol.~33, 2011.

\bibitem{CVPRW2017NTIREAgustsson}
E.~Agustsson and R.~Timofte, ``Ntire 2017 challenge on single image
  super-resolution: Dataset and study,'' in \emph{CVPRW}, 2017.

\bibitem{ECCV2016AcceleratingDong}
C.~Dong, C.~C. Loy, and X.~Tang, ``Accelerating the super-resolution
  convolutional neural network,'' in \emph{ECCV}, 2016.

\bibitem{CVPR2016SevenTimofte}
R.~Timofte, R.~Rothe, and L.~Van~Gool, ``Seven ways to improve example-based
  single image super resolution,'' in \emph{CVPR}, 2016.

\bibitem{MANPU2016Manga109Fujimoto}
A.~Fujimoto, T.~Ogawa, K.~Yamamoto, Y.~Matsui, T.~Yamasaki, and K.~Aizawa,
  ``Manga109 dataset and creation of metadata,'' in \emph{MANPU}, 2016.

\bibitem{CVPR2018RecoveringWang}
X.~Wang, K.~Yu, C.~Dong, and C.~C. Loy, ``Recovering realistic texture in image
  super-resolution by deep spatial feature transform,'' 2018.

\bibitem{ECCVW2018PIRMBlau}
Y.~Blau, R.~Mechrez, R.~Timofte, T.~Michaeli, and L.~Zelnik-Manor, ``2018 pirm
  challenge on perceptual image super-resolution,'' in \emph{ECCV Workshop},
  2018.

\bibitem{BMVC2012LowBevilacqua}
M.~Bevilacqua, A.~Roumy, C.~Guillemot, and M.~L. Alberi-Morel, ``Low-complexity
  single-image super-resolution based on nonnegative neighbor embedding,'' in
  \emph{BMVC}, 2012.

\bibitem{ICCS2010SingleZeyde}
R.~Zeyde, M.~Elad, and M.~Protter, ``On single image scale-up using
  sparse-representations,'' in \emph{International Conference on Curves and
  Surfaces}, 2010.

\bibitem{CVPR2015SingleHuang}
J.-B. Huang, A.~Singh, and N.~Ahuja, ``Single image super-resolution from
  transformed self-exemplars,'' in \emph{CVPR}, 2015.

\bibitem{CVPR2009ImagenetDeng}
J.~Deng, W.~Dong, R.~Socher, L.-J. Li, K.~Li, and L.~Fei-Fei, ``Imagenet: A
  large-scale hierarchical image database,'' in \emph{CVPR}, 2009.

\bibitem{ECCV2014MicrosoftLin}
T.-Y. Lin, M.~Maire, S.~Belongie, J.~Hays, P.~Perona, D.~Ramanan,
  P.~Doll{\'a}r, and C.~L. Zitnick, ``Microsoft coco: Common objects in
  context,'' in \emph{ECCV}, 2014.

\bibitem{IJCV2015PascalEveringham}
M.~Everingham, S.~A. Eslami, L.~Van~Gool, C.~K. Williams, J.~Winn, and
  A.~Zisserman, ``The pascal visual object classes challenge: A
  retrospective,'' \emph{IJCV}, vol. 111, 2015.

\bibitem{ICCV2015DeepLiu}
Z.~Liu, P.~Luo, X.~Wang, and X.~Tang, ``Deep learning face attributes in the
  wild,'' in \emph{ICCV}, 2015.

\bibitem{ICCV2017MemNetTai}
Y.~Tai, J.~Yang, X.~Liu, and C.~Xu, ``Memnet: A persistent memory network for
  image restoration,'' in \emph{ICCV}, 2017.

\bibitem{CVPR2017ImageTai}
Y.~Tai, J.~Yang, and X.~Liu, ``Image super-resolution via deep recursive
  residual network,'' in \emph{CVPR}, 2017.

\bibitem{CVPR2018DeepHaris}
M.~Haris, G.~Shakhnarovich, and N.~Ukita, ``Deep backp-rojection networks for
  super-resolution,'' in \emph{CVPR}, 2018.

\bibitem{TIP2004ImageWang}
Z.~Wang, A.~C. Bovik, H.~R. Sheikh, and E.~P. Simoncelli, ``Image quality
  assessment: From error visibility to structural similarity,'' \emph{IEEE
  Transactions on Image Processing}, vol.~13, 2004.

\bibitem{ICASSP2002WhyWang}
Z.~Wang, A.~C. Bovik, and L.~Lu, ``Why is image quality assessment so
  difficult?'' in \emph{ICASSP}, 2002.

\bibitem{TIP2006StatisticalSheikh}
H.~R. Sheikh, M.~F. Sabir, and A.~C. Bovik, ``A statistical evaluation of
  recent full reference image quality assessment algorithms,'' \emph{IEEE
  Transactions on Image Processing}, vol.~15, 2006.

\bibitem{ISPM2009MeanWang}
Z.~Wang and A.~C. Bovik, ``Mean squared error: Love it or leave it? a new look
  at signal fidelity measures,'' \emph{IEEE Signal Processing Magazine},
  vol.~26, 2009.

\bibitem{ICCV2015DeepWang}
Z.~Wang, D.~Liu, J.~Yang, W.~Han, and T.~Huang, ``Deep networks for image
  super-resolution with sparse prior,'' in \emph{ICCV}, 2015.

\bibitem{ICCV2017LearningXu}
X.~Xu, D.~Sun, J.~Pan, Y.~Zhang, H.~Pfister, and M.-H. Yang, ``Learning to
  super-resolve blurry face and text images,'' in \emph{ICCV}, 2017.

\bibitem{ICCV2017PixelDahl}
R.~Dahl, M.~Norouzi, and J.~Shlens, ``Pixel recursive super resolution,'' in
  \emph{ICCV}, 2017.

\bibitem{TPAMI2018FastLai}
W.-S. Lai, J.-B. Huang, N.~Ahuja, and M.-H. Yang, ``Fast and accurate image
  super-resolution with deep laplacian pyramid networks,'' \emph{TPAMI}, 2018.

\bibitem{CVIU2017LearningMa}
C.~Ma, C.-Y. Yang, X.~Yang, and M.-H. Yang, ``Learning a no-reference quality
  metric for single-image super-resolution,'' \emph{Computer Vision and Image
  Understanding}, 2017.

\bibitem{TIP2018NimaTalebi}
H.~Talebi and P.~Milanfar, ``Nima: Neural image assessment,'' \emph{IEEE
  Transactions on Image Processing}, vol.~27, 2018.

\bibitem{CVPR2017DeepKim}
J.~Kim and S.~Lee, ``Deep learning of human visual sensitivity in image quality
  assessment framework,'' in \emph{CVPR}, 2017.

\bibitem{CVPR2018TheZhang}
R.~Zhang, P.~Isola, A.~A. Efros, E.~Shechtman, and O.~Wang, ``The unreasonable
  effectiveness of deep features as a perceptual metric,'' in \emph{CVPR},
  2018.

\bibitem{ECCV2018ImageZhang}
Y.~Zhang, K.~Li, K.~Li, L.~Wang, B.~Zhong, and Y.~Fu, ``Image super-resolution
  using very deep residual channel attention networks,'' in \emph{ECCV}, 2018.

\bibitem{JVCIR2012EvaluationFookes}
C.~Fookes, F.~Lin, V.~Chandran, and S.~Sridharan, ``Evaluation of image
  resolution and super-resolution on face recognition performance,''
  \emph{Journal of Visual Communication and Image Representation}, vol.~23,
  2012.

\bibitem{ECCV2018SuperZhang}
K.~Zhang, Z.~ZHANG, C.-W. Cheng, W.~Hsu, Y.~Qiao, W.~Liu, and T.~Zhang,
  ``Super-identity convolutional neural network for face hallucination,'' in
  \emph{ECCV}, 2018.

\bibitem{CVPR2018FSRNetChen}
Y.~Chen, Y.~Tai, X.~Liu, C.~Shen, and J.~Yang, ``Fsrnet: End-to-end learning
  face super-resolution with facial priors,'' in \emph{CVPR}, 2018.

\bibitem{ACSSC2003MultiWang}
Z.~Wang, E.~Simoncelli, A.~Bovik \emph{et~al.}, ``Multi-scale structural
  similarity for image quality assessment,'' in \emph{Asilomar Conference on
  Signals, Systems, and Computers}, 2003.

\bibitem{TIP2011FSIMZhang}
L.~Zhang, L.~Zhang, X.~Mou, D.~Zhang \emph{et~al.}, ``Fsim: a feature
  similarity index for image quality assessment,'' \emph{IEEE transactions on
  Image Processing}, vol.~20, 2011.

\bibitem{SPL2013MakingMittal}
A.~Mittal, R.~Soundararajan, and A.~C. Bovik, ``Making a “completely blind”
  image quality analyzer,'' \emph{IEEE Signal Processing Letters}, 2013.

\bibitem{CVPR2018TheBlau}
Y.~Blau and T.~Michaeli, ``The perception-distortion tradeoff,'' in
  \emph{CVPR}, 2018.

\bibitem{NIPS2016ImageMao}
X.~Mao, C.~Shen, and Y.-B. Yang, ``Image restoration using very deep
  convolutional encoder-decoder networks with symmetric skip connections,'' in
  \emph{NIPS}, 2016.

\bibitem{ICCV2017ImageTong}
T.~Tong, G.~Li, X.~Liu, and Q.~Gao, ``Image super-resolution using dense skip
  connections,'' in \emph{ICCV}, 2017.

\bibitem{CVPRW2017NTIRETimofte}
R.~Timofte, E.~Agustsson, L.~Van~Gool, M.-H. Yang, L.~Zhang, B.~Lim, S.~Son,
  H.~Kim, S.~Nah, K.~M. Lee \emph{et~al.}, ``Ntire 2017 challenge on single
  image super-resolution: Methods and results,'' in \emph{CVPRW}, 2017.

\bibitem{ECCVW2018PIRMIgnatov}
A.~Ignatov, R.~Timofte, T.~Van~Vu, T.~Minh~Luu, T.~X~Pham, C.~Van~Nguyen,
  Y.~Kim, J.-S. Choi, M.~Kim, J.~Huang \emph{et~al.}, ``Pirm challenge on
  perceptual image enhancement on smartphones: report,'' in \emph{ECCV
  Workshop}, 2018.

\bibitem{CVPR2016DeeplyKim}
J.~Kim, J.~Kwon~Lee, and K.~Mu~Lee, ``Deeply-recursive convolutional network
  for image super-resolution,'' in \emph{CVPR}, 2016.

\bibitem{CVPR2018ZeroShocher}
A.~Shocher, N.~Cohen, and M.~Irani, ``“zero-shot” super-resolution using
  deep internal learning,'' in \emph{CVPR}, 2018.

\bibitem{CVPR2016RealShi}
W.~Shi, J.~Caballero, F.~Husz{\'a}r, J.~Totz, A.~P. Aitken, R.~Bishop,
  D.~Rueckert, and Z.~Wang, ``Real-time single image and video super-resolution
  using an efficient sub-pixel convolutional neural network,'' in \emph{CVPR},
  2016.

\bibitem{CVPR2018ImageHan}
W.~Han, S.~Chang, D.~Liu, M.~Yu, M.~Witbrock, and T.~S. Huang, ``Image
  super-resolution via dual-state recurrent networks,'' in \emph{CVPR}, 2018.

\bibitem{CVPR2019FeedbackLi}
Z.~Li, J.~Yang, Z.~Liu, X.~Yang, G.~Jeon, and W.~Wu, ``Feedback network for
  image super-resolution,'' in \emph{CVPR}, 2019.

\bibitem{CVPR2019RecurrentHaris}
M.~Haris, G.~Shakhnarovich, and N.~Ukita, ``Recurrent back-projection network
  for video super-resolution,'' in \emph{CVPR}, 2019.

\bibitem{ACCV2014AdjustedTimofte}
R.~Timofte, V.~De~Smet, and L.~Van~Gool, ``A+: Adjusted anchored neighborhood
  regression for fast super-resolution,'' in \emph{ACCV}, 2014.

\bibitem{CVPR2015FastSchulter}
S.~Schulter, C.~Leistner, and H.~Bischof, ``Fast and accurate image upscaling
  with super-resolution forests,'' in \emph{CVPR}, 2015.

\bibitem{CVPRW2010DeconvolutionalZeiler}
M.~D. Zeiler, D.~Krishnan, G.~W. Taylor, and R.~Fergus, ``Deconvolutional
  networks,'' in \emph{CVPRW}, 2010.

\bibitem{ECCV2014VisualizingZeiler}
M.~D. Zeiler and R.~Fergus, ``Visualizing and understanding convolutional
  networks,'' in \emph{ECCV}, 2014.

\bibitem{Distill2016DeconvolutionOdena}
A.~Odena, V.~Dumoulin, and C.~Olah, ``Deconvolution and checkerboard
  artifacts,'' \emph{Distill}, 2016.

\bibitem{CVPR2018ResidualZhang}
Y.~Zhang, Y.~Tian, Y.~Kong, B.~Zhong, and Y.~Fu, ``Residual dense network for
  image super-resolution,'' in \emph{CVPR}, 2018.

\bibitem{TPAMI2019PixelGao}
H.~Gao, H.~Yuan, Z.~Wang, and S.~Ji, ``Pixel transposed convolutional
  networks,'' \emph{TPAMI}, 2019.

\bibitem{CVPR2019MetaHu}
X.~Hu, H.~Mu, X.~Zhang, Z.~Wang, T.~Tan, and J.~Sun, ``Meta-sr: A
  magnification-arbitrary network for super-resolution,'' in \emph{CVPR}, 2019.

\bibitem{CVPR2016DeepHe}
K.~He, X.~Zhang, S.~Ren, and J.~Sun, ``Deep residual learning for image
  recognition,'' in \emph{CVPR}, 2016.

\bibitem{ICCV2013AnchoredTimofte}
R.~Timofte, V.~De~Smet, and L.~Van~Gool, ``Anchored neighborhood regression for
  fast example-based super-resolution,'' in \emph{ICCV}, 2013.

\bibitem{CVPR2018FastHui}
Z.~Hui, X.~Wang, and X.~Gao, ``Fast and accurate single image super-resolution
  via information distillation network,'' in \emph{CVPR}, 2018.

\bibitem{ECCV2018MultiLi}
J.~Li, F.~Fang, K.~Mei, and G.~Zhang, ``Multi-scale residual network for image
  super-resolution,'' in \emph{ECCV}, 2018.

\bibitem{CVPRW2017ImageRen}
H.~Ren, M.~El-Khamy, and J.~Lee, ``Image super resolution based on fusing
  multiple convolution neural networks,'' in \emph{CVPRW}, 2017.

\bibitem{CVPR2015GoingSzegedy}
C.~Szegedy, W.~Liu, Y.~Jia, P.~Sermanet, S.~Reed, D.~Anguelov, D.~Erhan,
  V.~Vanhoucke, and A.~Rabinovich, ``Going deeper with convolutions,'' in
  \emph{CVPR}, 2015.

\bibitem{CVPR2017DenselyHuang}
G.~Huang, Z.~Liu, L.~Van Der~Maaten, and K.~Q. Weinberger, ``Densely connected
  convolutional networks,'' in \emph{CVPR}, 2017.

\bibitem{ECCVW2018ESRGANWang}
X.~Wang, K.~Yu, S.~Wu, J.~Gu, Y.~Liu, C.~Dong, C.~C. Loy, Y.~Qiao, and X.~Tang,
  ``Esrgan: Enhanced super-resolution generative adversarial networks,'' in
  \emph{ECCV Workshop}, 2018.

\bibitem{CVPR2018SqueezeHu}
J.~Hu, L.~Shen, and G.~Sun, ``Squeeze-and-excitation networks,'' in
  \emph{CVPR}, 2018.

\bibitem{CVPR2019SecondDai}
T.~Dai, J.~Cai, Y.~Zhang, S.-T. Xia, and L.~Zhang, ``Second-order attention
  network for single image super-resolution,'' in \emph{CVPR}, 2019.

\bibitem{ICLR2019ResidualZhang}
Y.~Zhang, K.~Li, K.~Li, B.~Zhong, and Y.~Fu, ``Residual non-local attention
  networks for image restoration,'' \emph{ICLR}, 2019.

\bibitem{CVPR2017LearningZhang}
K.~Zhang, W.~Zuo, S.~Gu, and L.~Zhang, ``Learning deep cnn denoiser prior for
  image restoration,'' in \emph{CVPR}, 2017.

\bibitem{CVPR2017AggregatedXie}
S.~Xie, R.~Girshick, P.~Doll{\'a}r, Z.~Tu, and K.~He, ``Aggregated residual
  transformations for deep neural networks,'' in \emph{CVPR}, 2017.

\bibitem{CVPR2017XceptionChollet}
F.~Chollet, ``Xception: Deep learning with depthwise separable convolutions,''
  in \emph{CVPR}, 2017.

\bibitem{Arxiv2017MobilenetsHoward}
A.~G. Howard, M.~Zhu, B.~Chen, D.~Kalenichenko, W.~Wang, T.~Weyand,
  M.~Andreetto, and H.~Adam, ``Mobilenets: Efficient convolutional neural
  networks for mobile vision applications,'' \emph{Arxiv:1704.04861}, 2017.

\bibitem{NIPS2016ConditionalAaron}
A.~van~den Oord, N.~Kalchbrenner, L.~Espeholt, O.~Vinyals, A.~Graves
  \emph{et~al.}, ``Conditional image generation with pixelcnn decoders,'' in
  \emph{NIPS}, 2016.

\bibitem{Nature2005OptimalNajemnik}
J.~Najemnik and W.~S. Geisler, ``Optimal eye movement strategies in visual
  search,'' \emph{Nature}, vol. 434, 2005.

\bibitem{CVPR2017AttentionCao}
Q.~Cao, L.~Lin, Y.~Shi, X.~Liang, and G.~Li, ``Attention-aware face
  hallucination via deep reinforcement learning,'' in \emph{CVPR}, 2017.

\bibitem{ECCV2014SpatialHe}
K.~He, X.~Zhang, S.~Ren, and J.~Sun, ``Spatial pyramid pooling in deep
  convolutional networks for visual recognition,'' in \emph{ECCV}, 2014.

\bibitem{CVPR2017PyramidZhao}
H.~Zhao, J.~Shi, X.~Qi, X.~Wang, and J.~Jia, ``Pyramid scene parsing network,''
  in \emph{CVPR}, 2017.

\bibitem{CVPRW2018EfficientPark}
D.~Park, K.~Kim, and S.~Y. Chun, ``Efficient module based single image super
  resolution for multiple problems,'' in \emph{CVPRW}, 2018.

\bibitem{SBook1992TenDaubechies}
I.~Daubechies, \emph{Ten lectures on wavelets}.\hskip 1em plus 0.5em minus
  0.4em\relax SIAM, 1992.

\bibitem{EBook1999WaveletMallat}
S.~Mallat, \emph{A wavelet tour of signal processing}.\hskip 1em plus 0.5em
  minus 0.4em\relax Elsevier, 1999.

\bibitem{CVPRW2017BeyondBae}
W.~Bae, J.~J. Yoo, and J.~C. Ye, ``Beyond deep residual learning for image
  restoration: Persistent homology-guided manifold simplification,'' in
  \emph{CVPRW}, 2017.

\bibitem{CVPRW2017DeepGuo}
T.~Guo, H.~S. Mousavi, T.~H. Vu, and V.~Monga, ``Deep wavelet prediction for
  image super-resolution,'' in \emph{CVPRW}, 2017.

\bibitem{ICCV2017WaveletHuang}
H.~Huang, R.~He, Z.~Sun, T.~Tan \emph{et~al.}, ``Wavelet-srnet: A wavelet-based
  cnn for multi-scale face super resolution,'' in \emph{ICCV}, 2017.

\bibitem{CVPRW2018MultiLiu}
P.~Liu, H.~Zhang, K.~Zhang, L.~Lin, and W.~Zuo, ``Multi-level wavelet-cnn for
  image restoration,'' in \emph{CVPRW}, 2018.

\bibitem{ECCVW2018FastVu}
T.~Vu, C.~Van~Nguyen, T.~X. Pham, T.~M. Luu, and C.~D. Yoo, ``Fast and
  efficient image quality enhancement via desubpixel convolutional neural
  networks,'' in \emph{ECCV Workshop}, 2018.

\bibitem{CVPR2018xUnitKligvasser}
I.~Kligvasser, T.~Rott~Shaham, and T.~Michaeli, ``xunit: Learning a spatial
  activation function for efficient image restoration,'' in \emph{CVPR}, 2018.

\bibitem{IJCV2005LucasBruhn}
A.~Bruhn, J.~Weickert, and C.~Schn{\"o}rr, ``Lucas/kanade meets horn/schunck:
  Combining local and global optic flow methods,'' \emph{IJCV}, vol.~61, 2005.

\bibitem{TCI2017LossZhao}
H.~Zhao, O.~Gallo, I.~Frosio, and J.~Kautz, ``Loss functions for image
  restoration with neural networks,'' \emph{IEEE Transactions on Computational
  Imaging}, vol.~3, 2017.

\bibitem{NIPS2016GeneratingDosovitskiy}
A.~Dosovitskiy and T.~Brox, ``Generating images with perceptual similarity
  metrics based on deep networks,'' in \emph{NIPS}, 2016.

\bibitem{ICLR2015VerySimonyan}
K.~Simonyan and A.~Zisserman, ``Very deep convolutional networks for
  large-scale image recognition,'' in \emph{ICLR}, 2015.

\bibitem{NIPS2015TextureGatys}
L.~Gatys, A.~S. Ecker, and M.~Bethge, ``Texture synthesis using convolutional
  neural networks,'' in \emph{NIPS}, 2015.

\bibitem{CVPR2016ImageGatys}
L.~A. Gatys, A.~S. Ecker, and M.~Bethge, ``Image style transfer using
  convolutional neural networks,'' in \emph{CVPR}, 2016.

\bibitem{CVPRW2018UnsupervisedYuan}
Y.~Yuan, S.~Liu, J.~Zhang, Y.~Zhang, C.~Dong, and L.~Lin, ``Unsupervised image
  super-resolution using cycle-in-cycle generative adversarial networks,'' in
  \emph{CVPRW}, 2018.

\bibitem{ICCV2017LeastMao}
X.~Mao, Q.~Li, H.~Xie, R.~Y. Lau, Z.~Wang, and S.~P. Smolley, ``Least squares
  generative adversarial networks,'' in \emph{ICCV}, 2017.

\bibitem{ECCV2018SRFeatPark}
S.-J. Park, H.~Son, S.~Cho, K.-S. Hong, and S.~Lee, ``Srfeat: Single image
  super resolution with feature discrimination,'' in \emph{ECCV}, 2018.

\bibitem{Arxiv2018RelativisticJolicoeur}
A.~Jolicoeur-Martineau, ``The relativistic discriminator: a key element missing
  from standard gan,'' \emph{Arxiv:1807.00734}, 2018.

\bibitem{ICML2017WassersteinArjovsky}
M.~Arjovsky, S.~Chintala, and L.~Bottou, ``Wasserstein generative adversarial
  networks,'' in \emph{ICML}, 2017.

\bibitem{NIPS2017ImprovedGulrajani}
I.~Gulrajani, F.~Ahmed, M.~Arjovsky, V.~Dumoulin, and A.~C. Courville,
  ``Improved training of wasserstein gans,'' in \emph{NIPS}, 2017.

\bibitem{ICLR2018SpectralMiyato}
T.~Miyato, T.~Kataoka, M.~Koyama, and Y.~Yoshida, ``Spectral normalization for
  generative adversarial networks,'' in \emph{ICLR}, 2018.

\bibitem{ICCV2017UnpairedZhu}
J.-Y. Zhu, T.~Park, P.~Isola, and A.~A. Efros, ``Unpaired image-to-image
  translation using cycle-consistent adversarial networks,'' in \emph{ICCV},
  2017.

\bibitem{PDNP1992NonlinearRudin}
L.~I. Rudin, S.~Osher, and E.~Fatemi, ``Nonlinear total variation based noise
  removal algorithms,'' \emph{Physica D: Nonlinear Phenomena}, vol.~60, 1992.

\bibitem{TIP2005ImageAly}
H.~A. Aly and E.~Dubois, ``Image up-sampling using total-variation
  regularization with a new observation model,'' \emph{IEEE Transactions on
  Image Processing}, vol.~14, 2005.

\bibitem{Arxiv2018DualGuo}
Y.~Guo, Q.~Chen, J.~Chen, J.~Huang, Y.~Xu, J.~Cao, P.~Zhao, and M.~Tan, ``Dual
  reconstruction nets for image super-resolution with gradient sensitive
  loss,'' \emph{arXiv:1809.07099}, 2018.

\bibitem{ECCVW2018AnalyzingVasu}
S.~Vasu, N.~T. Madam \emph{et~al.}, ``Analyzing perception-distortion tradeoff
  using enhanced perceptual super-resolution network,'' in \emph{ECCV
  Workshop}, 2018.

\bibitem{ECCVW2018GenerativeCheon}
M.~Cheon, J.-H. Kim, J.-H. Choi, and J.-S. Lee, ``Generative adversarial
  network-based image super-resolution using perceptual content losses,'' in
  \emph{ECCV Workshop}, 2018.

\bibitem{ECCVW2018DeepChoi}
J.-H. Choi, J.-H. Kim, M.~Cheon, and J.-S. Lee, ``Deep learning-based image
  super-resolution considering quantitative and perceptual quality,'' in
  \emph{ECCV Workshop}, 2018.

\bibitem{ICML2015BatchSergey}
I.~Sergey and S.~Christian, ``Batch normalization: Accelerating deep network
  training by reducing internal covariate shift,'' in \emph{ICML}, 2015.

\bibitem{ICLR2017AmortisedSonderby}
C.~K. S{\o}nderby, J.~Caballero, L.~Theis, W.~Shi, and F.~Husz{\'a}r,
  ``Amortised map inference for image super-resolution,'' in \emph{ICLR}, 2017.

\bibitem{CVPRW2018PersistentChen}
R.~Chen, Y.~Qu, K.~Zeng, J.~Guo, C.~Li, and Y.~Xie, ``Persistent memory
  residual network for single image super resolution,'' in \emph{CVPRW}, 2018.

\bibitem{ICML2009CurriculumBengio}
Y.~Bengio, J.~Louradour, R.~Collobert, and J.~Weston, ``Curriculum learning,''
  in \emph{ICML}, 2009.

\bibitem{CVPRW2018NewBei}
Y.~Bei, A.~Damian, S.~Hu, S.~Menon, N.~Ravi, and C.~Rudin, ``New techniques for
  preserving global structure and denoising with low information loss in
  single-image super-resolution,'' in \emph{CVPRW}, 2018.

\bibitem{CVPRW2018ImageAhn}
N.~Ahn, B.~Kang, and K.-A. Sohn, ``Image super-resolution via progressive
  cascading residual network,'' in \emph{CVPRW}, 2018.

\bibitem{ICLR2018ProgressiveKarras}
T.~Karras, T.~Aila, S.~Laine, and J.~Lehtinen, ``Progressive growing of gans
  for improved quality, stability, and variation,'' in \emph{ICLR}, 2018.

\bibitem{ML1997MultitaskCaruana}
R.~Caruana, ``Multitask learning,'' \emph{Machine Learning}, vol.~28, 1997.

\bibitem{ICCV2017MaskHe}
K.~He, G.~Gkioxari, P.~Doll{\'a}r, and R.~Girshick, ``Mask r-cnn,'' in
  \emph{ICCV}, 2017.

\bibitem{ECCV2014FicialZhang}
Z.~Zhang, P.~Luo, C.~C. Loy, and X.~Tang, ``Facial landmark detection by deep
  multi-task learning,'' in \emph{ECCV}, 2014.

\bibitem{CVPR2019DeepWang}
X.~Wang, K.~Yu, C.~Dong, X.~Tang, and C.~C. Loy, ``Deep network interpolation
  for continuous imagery effect transition,'' in \emph{CVPR}, 2019.

\bibitem{CVPR2017RealCaballero}
J.~Caballero, C.~Ledig, A.~P. Aitken, A.~Acosta, J.~Totz, Z.~Wang, and W.~Shi,
  ``Real-time video super-resolution with spatio-temporal networks and motion
  compensation,'' in \emph{CVPR}, 2017.

\bibitem{Code2019OpCounterZhu}
L.~Zhu, ``pytorch-opcounter,''
  \url{https://github.com/Lyken17/pytorch-OpCounter}, 2019.

\bibitem{ICCV2013NonparametricMichaeli}
T.~Michaeli and M.~Irani, ``Nonparametric blind super-resolution,'' in
  \emph{ICCV}, 2013.

\bibitem{ECCV2018LearnBulat}
A.~Bulat, J.~Yang, and G.~Tzimiropoulos, ``To learn image super-resolution, use
  a gan to learn how to do image degradation first,'' in \emph{ECCV}, 2018.

\bibitem{CVPR2018DeepUlyanov}
D.~Ulyanov, A.~Vedaldi, and V.~Lempitsky, ``Deep image prior,'' in \emph{CVPR},
  2018.

\bibitem{CVPR2011RealShotton}
J.~Shotton, A.~Fitzgibbon, M.~Cook, T.~Sharp, M.~Finocchio, R.~Moore,
  A.~Kipman, and A.~Blake, ``Real-time human pose recognition in parts from
  single depth images,'' in \emph{CVPR}, 2011.

\bibitem{CVPR2018V2VMoon}
G.~Moon, J.~Yong~Chang, and K.~Mu~Lee, ``V2v-posenet: Voxel-to-voxel prediction
  network for accurate 3d hand and human pose estimation from a single depth
  map,'' in \emph{CVPR}, 2018.

\bibitem{ECCV2014LearningGupta}
S.~Gupta, R.~Girshick, P.~Arbel{\'a}ez, and J.~Malik, ``Learning rich features
  from rgb-d images for object detection and segmentation,'' in \emph{ECCV},
  2014.

\bibitem{ECCV2018DepthWang}
W.~Wang and U.~Neumann, ``Depth-aware cnn for rgb-d segmentation,'' in
  \emph{ECCV}, 2018.

\bibitem{ACCV2016DeepSong}
X.~Song, Y.~Dai, and X.~Qin, ``Deep depth super-resolution: Learning depth
  super-resolution using deep convolutional neural network,'' in \emph{ACCV},
  2016.

\bibitem{ECCV2016DepthHui}
T.-W. Hui, C.~C. Loy, and X.~Tang, ``Depth map super-resolution by deep
  multi-scale guidance,'' in \emph{ECCV}, 2016.

\bibitem{CVPR2018FightHaefner}
B.~Haefner, Y.~Qu{\'e}au, T.~M{\"o}llenhoff, and D.~Cremers, ``Fight
  ill-posedness with ill-posedness: Single-shot variational depth
  super-resolution from shading,'' in \emph{CVPR}, 2018.

\bibitem{ECCV2016ATGVRiegler}
G.~Riegler, M.~R{\"u}ther, and H.~Bischof, ``Atgv-net: Accurate depth
  super-resolution,'' in \emph{ECCV}, 2016.

\bibitem{TIP2008ExamplePark}
J.-S. Park and S.-W. Lee, ``An example-based face hallucination method for
  single-frame, low-resolution facial images,'' \emph{IEEE Transactions on
  Image Processing}, vol.~17, 2008.

\bibitem{ECCV2016DeepZhu}
S.~Zhu, S.~Liu, C.~C. Loy, and X.~Tang, ``Deep cascaded bi-network for face
  hallucination,'' in \emph{ECCV}, 2016.

\bibitem{ECCV2018FaceYu}
X.~Yu, B.~Fernando, B.~Ghanem, F.~Porikli, and R.~Hartley, ``Face
  super-resolution guided by facial component heatmaps,'' in \emph{ECCV}, 2018.

\bibitem{AAAI2017FaceYu}
X.~Yu and F.~Porikli, ``Face hallucination with tiny unaligned images by
  transformative discriminative neural networks,'' in \emph{AAAI}, 2017.

\bibitem{NIPS2015SpatialJaderberg}
M.~Jaderberg, K.~Simonyan, A.~Zisserman \emph{et~al.}, ``Spatial transformer
  networks,'' in \emph{NIPS}, 2015.

\bibitem{CVPR2017HallucinatingYu}
X.~Yu and F.~Porikli, ``Hallucinating very low-resolution unaligned and noisy
  face images by transformative discriminative autoencoders,'' in \emph{CVPR},
  2017.

\bibitem{IJCAI2017LearningSong}
Y.~Song, J.~Zhang, S.~He, L.~Bao, and Q.~Yang, ``Learning to hallucinate face
  images via component generation and enhancement,'' in \emph{IJCAI}, 2017.

\bibitem{IJCV2018HallucinatingYang}
C.-Y. Yang, S.~Liu, and M.-H. Yang, ``Hallucinating compressed face images,''
  \emph{IJCV}, vol. 126, 2018.

\bibitem{ECCV2016UltraYu}
X.~Yu and F.~Porikli, ``Ultra-resolving face images by discriminative
  generative networks,'' in \emph{ECCV}, 2016.

\bibitem{CVPR2018AttributeLee}
C.-H. Lee, K.~Zhang, H.-C. Lee, C.-W. Cheng, and W.~Hsu, ``Attribute augmented
  convolutional neural network for face hallucination,'' in \emph{CVPRW}, 2018.

\bibitem{CVPR2018SuperYu}
X.~Yu, B.~Fernando, R.~Hartley, and F.~Porikli, ``Super-resolving very
  low-resolution face images with supplementary attributes,'' in \emph{CVPR},
  2018.

\bibitem{Arxiv2014ConditionalMirza}
M.~Mirza and S.~Osindero, ``Conditional generative adversarial nets,''
  \emph{Arxiv:1411.1784}, 2014.

\bibitem{PI2013AdvancesFauvel}
M.~Fauvel, Y.~Tarabalka, J.~A. Benediktsson, J.~Chanussot, and J.~C. Tilton,
  ``Advances in spectral-spatial classification of hyperspectral images,''
  \emph{Proceedings of the IEEE}, vol. 101, 2013.

\bibitem{CVPR2016ExploitingFu}
Y.~Fu, Y.~Zheng, I.~Sato, and Y.~Sato, ``Exploiting spectral-spatial
  correlation for coded hyperspectral image restoration,'' in \emph{CVPR},
  2016.

\bibitem{CVPRW2017AerialUzkent}
B.~Uzkent, A.~Rangnekar, and M.~J. Hoffman, ``Aerial vehicle tracking by
  adaptive fusion of hyperspectral likelihood maps,'' in \emph{CVPRW}, 2017.

\bibitem{RS2016PansharpeningMasi}
G.~Masi, D.~Cozzolino, L.~Verdoliva, and G.~Scarpa, ``Pansharpening by
  convolutional neural networks,'' \emph{Remote Sensing}, vol.~8, 2016.

\bibitem{CVPR2018UnsupervisedQu}
Y.~Qu, H.~Qi, and C.~Kwan, ``Unsupervised sparse dirichlet-net for
  hyperspectral image super-resolution,'' in \emph{CVPR}, 2018.

\bibitem{CVPR2019HyperspectralFu}
Y.~Fu, T.~Zhang, Y.~Zheng, D.~Zhang, and H.~Huang, ``Hyperspectral image
  super-resolution with optimized rgb guidance,'' in \emph{CVPR}, 2019.

\bibitem{CVPR2019CameraChen}
C.~Chen, Z.~Xiong, X.~Tian, Z.-J. Zha, and F.~Wu, ``Camera lens
  super-resolution,'' in \emph{CVPR}, 2019.

\bibitem{CVPR2019ZoomZhang}
X.~Zhang, Q.~Chen, R.~Ng, and V.~Koltun, ``Zoom to learn, learn to zoom,'' in
  \emph{CVPR}, 2019.

\bibitem{CVPR2019TowardsXu}
X.~Xu, Y.~Ma, and W.~Sun, ``Towards real scene super-resolution with raw
  images,'' in \emph{CVPR}, 2019.

\bibitem{ICCV2015VideoLiao}
R.~Liao, X.~Tao, R.~Li, Z.~Ma, and J.~Jia, ``Video super-resolution via deep
  draft-ensemble learning,'' in \emph{ICCV}, 2015.

\bibitem{TCIVideoKappeler}
A.~Kappeler, S.~Yoo, Q.~Dai, and A.~K. Katsaggelos, ``Video super-resolution
  with convolutional neural networks,'' \emph{IEEE Transactions on
  Computational Imaging}, vol.~2, 2016.

\bibitem{ICIP2016SuperKappeler}
------, ``Super-resolution of compressed videos using convolutional neural
  networks,'' in \emph{ICIP}, 2016.

\bibitem{ITSC2011TotalDrulea}
M.~Drulea and S.~Nedevschi, ``Total variation regularization of local-global
  optical flow,'' in \emph{ITSC}, 2011.

\bibitem{ICCV2017RobustLiu}
D.~Liu, Z.~Wang, Y.~Fan, X.~Liu, Z.~Wang, S.~Chang, and T.~Huang, ``Robust
  video super-resolution with learned temporal dynamics,'' in \emph{ICCV},
  2017.

\bibitem{TIP2018LearningLiu}
D.~Liu, Z.~Wang, Y.~Fan, X.~Liu, Z.~Wang, S.~Chang, X.~Wang, and T.~S. Huang,
  ``Learning temporal dynamics for video super-resolution: A deep learning
  approach,'' \emph{IEEE Transactions on Image Processing}, vol.~27, 2018.

\bibitem{ICCV2017DetailTao}
X.~Tao, H.~Gao, R.~Liao, J.~Wang, and J.~Jia, ``Detail-revealing deep video
  super-resolution,'' in \emph{ICCV}, 2017.

\bibitem{NIPS2015BidirectionalHuang}
Y.~Huang, W.~Wang, and L.~Wang, ``Bidirectional recurrent convolutional
  networks for multi-frame super-resolution,'' in \emph{NIPS}, 2015.

\bibitem{TPAMI2018VideoHuang}
------, ``Video super-resolution via bidirectional recurrent convolutional
  networks,'' \emph{TPAMI}, vol.~40, 2018.

\bibitem{AAAI2017BuildingGuo}
J.~Guo and H.~Chao, ``Building an end-to-end spatial-temporal convolutional
  network for video super-resolution,'' in \emph{AAAI}, 2017.

\bibitem{ICANN2005BidirectionalGraves}
A.~Graves, S.~Fern{\'a}ndez, and J.~Schmidhuber, ``Bidirectional lstm networks
  for improved phoneme classification and recognition,'' in \emph{ICANN}, 2005.

\bibitem{CVPR2018FrameSajjadi}
M.~S. Sajjadi, R.~Vemulapalli, and M.~Brown, ``Frame-recurrent video
  super-resolution,'' in \emph{CVPR}, 2018.

\bibitem{CVPR2019FastLi}
S.~Li, F.~He, B.~Du, L.~Zhang, Y.~Xu, and D.~Tao, ``Fast spatio-temporal
  residual network for video super-resolution,'' in \emph{CVPR}, 2019.

\bibitem{CVPRW2017FASTZhang}
Z.~Zhang and V.~Sze, ``Fast: A framework to accelerate super-resolution
  processing on compressed videos,'' in \emph{CVPRW}, 2017.

\bibitem{CVPR2018DeepJo}
Y.~Jo, S.~W. Oh, J.~Kang, and S.~J. Kim, ``Deep video super-resolution network
  using dynamic upsampling filters without explicit motion compensation,'' in
  \emph{CVPR}, 2018.

\bibitem{CVPR2017PerceptualLi}
J.~Li, X.~Liang, Y.~Wei, T.~Xu, J.~Feng, and S.~Yan, ``Perceptual generative
  adversarial networks for small object detection,'' in \emph{CVPR}, 2017.

\bibitem{CVPR2018FeatureTan}
W.~Tan, B.~Yan, and B.~Bare, ``Feature super-resolution: Make machine see more
  clearly,'' in \emph{CVPR}, 2018.

\bibitem{CVPR2018EnhancingJeon}
D.~S. Jeon, S.-H. Baek, I.~Choi, and M.~H. Kim, ``Enhancing the spatial
  resolution of stereo images using a parallax prior,'' in \emph{CVPR}, 2018.

\bibitem{CVPR2019LearningWang}
L.~Wang, Y.~Wang, Z.~Liang, Z.~Lin, J.~Yang, W.~An, and Y.~Guo, ``Learning
  parallax attention for stereo image super-resolution,'' in \emph{CVPR}, 2019.

\bibitem{CVPR20193DLi}
Y.~Li, V.~Tsiminaki, R.~Timofte, M.~Pollefeys, and L.~V. Gool, ``3d appearance
  super-resolution with deep learning,'' in \emph{CVPR}, 2019.

\bibitem{CVPR2019ResidualZhang}
S.~Zhang, Y.~Lin, and H.~Sheng, ``Residual networks for light field image
  super-resolution,'' in \emph{CVPR}, 2019.

\bibitem{CVPRW2018NTIREAncuti}
C.~Ancuti, C.~O. Ancuti, R.~Timofte, L.~Van~Gool, L.~Zhang, M.-H. Yang, V.~M.
  Patel, H.~Zhang, V.~A. Sindagi, R.~Zhao \emph{et~al.}, ``Ntire 2018 challenge
  on image dehazing: Methods and results,'' in \emph{CVPRW}, 2018.

\bibitem{ICML2018EfficientPham}
H.~Pham, M.~Y. Guan, B.~Zoph, Q.~V. Le, and J.~Dean, ``Efficient neural
  architecture search via parameter sharing,'' in \emph{ICML}, 2018.

\bibitem{ICLR2019DartsLiu}
H.~Liu, K.~Simonyan, and Y.~Yang, ``Darts: Differentiable architecture
  search,'' \emph{ICLR}, 2019.

\bibitem{NIPS2019NATGuo}
Y.~Guo, Y.~Zheng, M.~Tan, Q.~Chen, J.~Chen, P.~Zhao, and J.~Huang, ``Nat:
  Neural architecture transformer for accurate and compact architectures,'' in
  \emph{NIPS}, 2019, pp. 735--747.

\end{thebibliography}

\begin{IEEEbiography}
    [{\includegraphics[width=1.1in,clip,keepaspectratio]{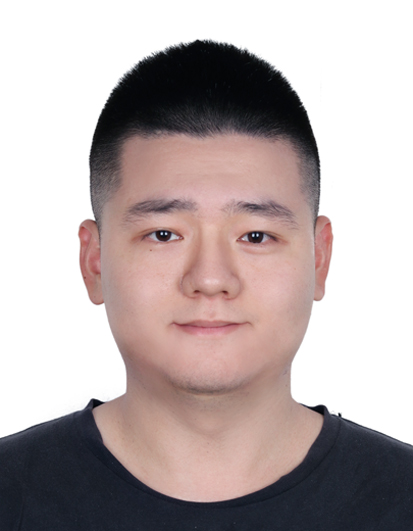}}]
    {Zhihao Wang} received the BE degree in South China University of Technology (SCUT), China, in 2017, and is working toward the ME degree at the School of Software Engineering, SCUT.
    Now he is as a visiting student at the School of Information Systems, Singapore Management University, Singapore. His research interests are computer vision based on deep learning, including visual recognition and image super-resolution.
\vspace{-0.2in}
\end{IEEEbiography}

\begin{IEEEbiography}
    [{\includegraphics[width=1.1in,clip,keepaspectratio]{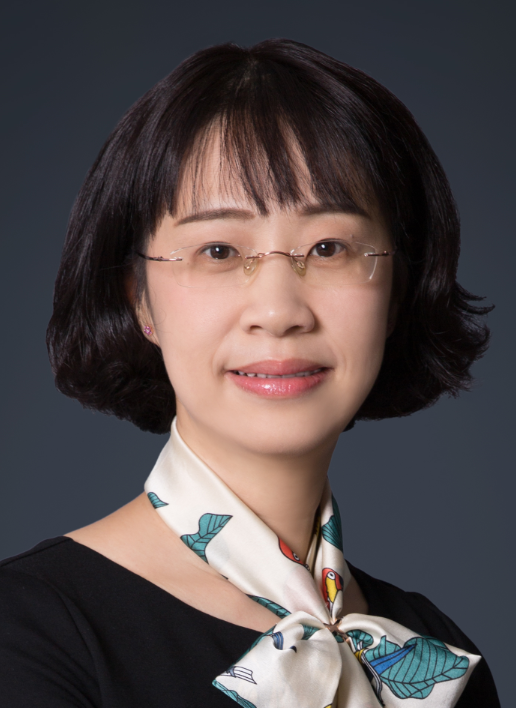}}]
    {Jian Chen} is currently a Professor of the School of Software Engineering at South China University of Technology where she started as an Assistant Professor in 2005.
    She received her B.S. and Ph.D. degrees, both in Computer Science, from Sun Yat-Sen University, China, in 2000 and 2005 respectively.
    Her research interests can be summarized as developing effective and efficient data analysis techniques for complex data and the related applications.
\vspace{-0.2in}
\end{IEEEbiography}

\begin{IEEEbiography}
    [{\includegraphics[width=1.1in,clip,keepaspectratio]{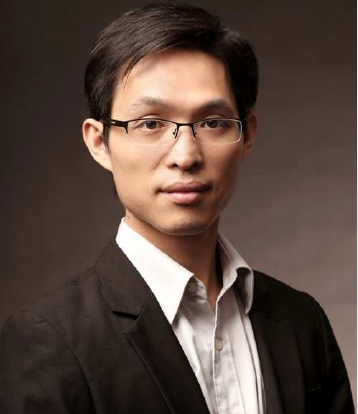}}]
    {Steven C. H.~Hoi} is currently the Managing Director of Salesforce Research Asia, and an Associate Professor (on leave) of the School of Information Systems, Singapore Management University, Singapore. Prior to joining SMU, he was an Associate Professor with Nanyang Technological University, Singapore. He received his Bachelor degree from Tsinghua University, P.R. China, in 2002, and his Ph.D degree in computer science and engineering from The Chinese University of Hong Kong, in 2006.
    His research interests are machine learning and data mining and their applications to multimedia information retrieval (image and video retrieval), social media and web mining, and computational finance, etc., and he has published over 150 refereed papers in top conferences and journals in these related areas.
    He has served as the Editor-in-Chief for Neurocomputing Journal, general co-chair for ACM SIGMM Workshops on Social Media (WSM'09, WSM'10, WSM'11), program co-chair for the fourth Asian Conference on Machine Learning (ACML'12), book editor for ``Social Media Modeling and Computing'', guest editor for ACM Transactions on Intelligent Systems and Technology (ACM TIST), technical PC member for many international conferences, and external reviewer for many top journals and worldwide funding agencies, including NSF in US and RGC in Hong Kong. He is an IEEE Fellow and ACM Distinguished Member.
\vspace{-0.2in}
\end{IEEEbiography}

\clearpage

\setcounter{table}{0}
\setcounter{page}{1}

\begin{table*}[h]
    \renewcommand{\arraystretch}{1.3}
    \caption{Notations.}
    \label{tab_notations}
    \centering
    \begin{tabular}{|c|l|}
        \hline
        Notation        & Description \\
        \hline
        $I_x$           & LR image \\
        $I_y$           & ground truth HR image, abbreviated as $I$ \\
        $\hat{I_y}$     & reconstructed HR image, abbreviated as $\hat{I}$ \\
        $I_s$           & randomly sampled HR image from the real HR images \\
        \hline
        $I(i)$          & intensity of the $i$-th pixel of image $I$ \\
        $D$             & discriminator network of GAN \\
        $\phi$          & image classification network \\
        $\phi^{(l)}$    & extracted representations on $l$-th layer by $\phi$ \\
        $\operatorname{vec}$ &  vectorization operation \\
        $G^{(l)}$       & Gram matrix of representations on $l$-th layer \\
        $l$             & layer of CNN \\
        $h$, $w$, $c$   & width, height and number of channels of feature maps \\
        $h_l$, $w_l$, $c_l$  & width, height and number of channels of feature maps in $l$-th layer \\
        \hline
        $\mathcal{D}$   & degradation process \\
        $\delta$        & parameters of $\mathcal{D}$ \\
        $\mathcal{F}$   & super-resolution process \\
        $\theta$        & parameters of $\mathcal{F}$ \\
        \hline
        $\otimes$       & convolution operation \\
        $\kappa$        & convolution kernel \\
        $\downarrow$    & downsampling operation \\
        $s$             & scaling factor \\
        $n$             & Gaussian noise \\
        $\varsigma$     & standard deviation of $n$ \\
        $z$             & a random vector \\
        \hline
        $\mathcal{L}$   & loss function \\
        $\mathcal{L}_{\text{content}}$     & content loss \\
        $\mathcal{L}_{\text{cycle}}$       & content consistency loss \\
        $\mathcal{L}_{\text{pixel\_l1}}$   & pixel L1 loss \\
        $\mathcal{L}_{\text{pixel\_l2}}$   & pixel L2 loss \\
        $\mathcal{L}_{\text{pixel\_Cha}}$  & pixel Charbonnier loss \\
        $\mathcal{L}_{\text{gan\_ce\_g}}$, $\mathcal{L}_{\text{gan\_ce\_d}}$  & adversarial loss of the generator and discriminator based on cross entropy \\
        $\mathcal{L}_{\text{gan\_hi\_g}}$, $\mathcal{L}_{\text{gan\_hi\_d}}$  & adversarial loss of the generator and discriminator based on hinge error \\
        $\mathcal{L}_{\text{gan\_ls\_g}}$, $\mathcal{L}_{\text{gan\_ls\_g}}$  & adversarial loss of the generator and discriminator based on least square error \\
        $\mathcal{L}_{\text{TV}}$          & total variation loss \\
        $\Phi$          & regularization term \\
        $\lambda$       & tradeoff parameter of $\Phi$ \\
        $\epsilon$      & small instant for stability \\
        \hline
        $\mu_I$         & luminance of image $I$, i.e., mean of intensity \\
        $\sigma_I$      & contrast of image $I$, i.e., standard deviation of intensity \\
        $\sigma_{I,\hat{I}}$ & covariance between images $I$ and $\hat{I}$ \\
        $\mathcal{C}_l$, $\mathcal{C}_c$, $\mathcal{C}_s$ & comparison function of luminance, contrast, structure \\
        $\alpha$, $\beta$, $\gamma$                       & weights of $\mathcal{C}_l$, $\mathcal{C}_c$, $\mathcal{C}_s$ \\
        $C_1$, $C_2$, $C_3$  & constants \\
        $k_1$, $k_2$         & constants \\
        \hline
        $L$  & maximum possible pixel value \\
        $N$  & number of pixels \\
        $M$  & number of bins \\
        \hline
    \end{tabular}
\end{table*}

\begin{table*}[h]
    \renewcommand{\arraystretch}{1.3}
    \caption{Abbreviations.}
    \label{tab_abbreviations}
    \centering
    \begin{tabular}{|clcl|}
        \hline
        Abbreviation & Full name   & Abbreviation & Full name \\
        \hline
        FH & face hallucination    & PAN & panchromatic image \\
        HR & high-resolution       & SR & super-resolution \\
        HSI & hyperspectral image  & TV & total variation \\
        HVS & human visual system  & WT & wavelet transformation \\
        LR & low-resolution        & & \\
        \hline
        FSIM \cite{TIP2011FSIMZhang} & feature similarity                     & MS-SSIM \cite{ACSSC2003MultiWang} & multi-scale SSIM \\
        IQA & image quality assessment                                        & NIQE \cite{SPL2013MakingMittal} & natural image quality evaluator \\
        MOS & mean opinion score                                              & PSNR & peak signal-to-noise ratio \\
        MSSIM \cite{TIP2004ImageWang} & mean SSIM                             & SSIM \cite{TIP2004ImageWang} & structural similarity \\
        \hline
        BN \cite{ICML2015BatchSergey} & batch normalization           & GAN \cite{NIPS2014GenerativeGoodfellow} & generative adversarial net \\
        CNN & convolutional neural network                            & LSTM & long short term memory network \\
        CycleGAN \cite{ICCV2017UnpairedZhu} & cycle-in-cycle GAN      & ResNet \cite{CVPR2016DeepHe} & residual network \\
        DenseNet \cite{CVPR2017DenselyHuang} & densely connected CNN  & SENet \cite{CVPR2018SqueezeHu} & squeeze-and-excitation network \\
        FAN & face alignment network                                  & SPMC \cite{ICCV2017DetailTao} & sub-pixel motion compensation \\
        \hline
        ADRSR \cite{CVPRW2018NewBei} & automated decomposition and reconstruction     & LCGE \cite{IJCAI2017LearningSong} & learn FH via component generation and enhancement \\
        Attention-FH \cite{CVPR2017AttentionCao} & attention-aware FH                 & MemNet \cite{ICCV2017MemNetTai} & memory network \\
        BRCN \cite{NIPS2015BidirectionalHuang,TPAMI2018VideoHuang} & bidirectional recurrent CNN & MS-LapSRN \cite{TPAMI2018FastLai} & multi-scale LapSRN \\
        CARN \cite{ECCV2018FastAhn} & cascading residual network                      & MSRN \cite{ECCV2018MultiLi} & multiscale residual network \\
        CARN-M \cite{ECCV2018FastAhn} & CARM based on MobileNet                       & MTUN \cite{ECCV2018FaceYu} & multi-task upsampling network \\
        CBN \cite{ECCV2016DeepZhu} & cascaded bi-network                              & MWCNN \cite{CVPRW2018MultiLiu} & multi-level wavelet CNN \\
        CinCGAN \cite{CVPRW2018UnsupervisedYuan} & cycle-in-cycle GAN                 & ProSR \cite{CVPRW2018FullyWang} & progressive SR \\
        CNF \cite{CVPRW2017ImageRen} & context-wise network fusion                    & RBPN \cite{CVPR2019RecurrentHaris} & recurrent back-projection network \\
        CVSRnet \cite{ICIP2016SuperKappeler} & compressed VSRnet                      & RCAN \cite{ECCV2018ImageZhang} & residual channel attention networks \\
        DBPN \cite{CVPR2018DeepHaris} & deep back-projection network                  & RDN \cite{CVPR2018ResidualZhang} & residual dense network \\
        DNSR \cite{CVPRW2018NewBei} & denoising for SR                                & RNAN \cite{ICLR2019ResidualZhang} & residual non-local attention networks \\
        DRCN \cite{CVPR2016DeeplyKim} & deeply-recursive CNN                          & SAN \cite{CVPR2019SecondDai} & Second-order Attention Network \\
        DRRN \cite{CVPR2017ImageTai} & deep recursive residual network                & SFT-GAN \cite{CVPR2018RecoveringWang} & GAN with spatial feature transformation \\
        DSRN \cite{CVPR2018ImageHan} & dual-state recurrent network                   & SICNN \cite{ECCV2018SuperZhang} & super-identity CNN \\
        DWSR \cite{CVPRW2017DeepGuo} & deep wavelet prediction for SR                 & SOCA \cite{CVPR2019SecondDai} & second-order channel attention \\
        EDSR \cite{CVPRW2017EnhancedLim} & enhanced deep SR network                   & SRCNN \cite{ECCV2014LearningDong,TPAMI2016ImageDong} & SR CNN \\
        EDSR-PP \cite{CVPRW2018EfficientPark} & EDSR with pyramid pooling             & SRFBN \cite{CVPR2019FeedbackLi} & SR feedback network \\
        ESPCN \cite{CVPR2017RealCaballero} & efficient sub-pixel CNN                  & SRGAN \cite{CVPR2017PhotoLedig} & SR GAN \\
        ESRGAN \cite{ECCVW2018ESRGANWang} & enhanced SRGAN                            & SRDenseNet \cite{ICCV2017ImageTong} & SR DenseNet \\
        FAST \cite{CVPRW2017FASTZhang} & free adaptive SR via transfer                & STCN \cite{AAAI2017BuildingGuo} & spatial-temporal CNN \\
        FRVSR \cite{CVPR2018FrameSajjadi} & frame-recurrent video SR                  & TDAE \cite{CVPR2017HallucinatingYu} & transformative discriminative auto-encoder \\
        FSRCNN \cite{ECCV2016AcceleratingDong} & fast SRCNN                           & TDN \cite{AAAI2017FaceYu} & transformative discriminative network \\
        FSR-GAN \cite{CVPR2018FeatureTan} & feature SRGAN                             & Super-FAN \cite{CVPR2018SuperBulat} & SR with FAN \\
        FSRNet \cite{CVPR2018FSRNetChen} & face SR network                            & UR-DGN \cite{ECCV2016UltraYu} & ultra-resolving by discriminative generative networks \\
        FSTRN \cite{CVPR2019FastLi} & fast spatio-temporal ResNet                     & VESPCN \cite{CVPR2017RealCaballero} & video ESPCN \\
        IDN \cite{CVPR2018FastHui} & information distillation network                 & VSRnet \cite{TCIVideoKappeler} & video SR network \\
        LapSRN \cite{CVPR2017DeepLai,TPAMI2018FastLai} & Laplacian pyramid SR network & ZSSR \cite{CVPR2018ZeroShocher} & zero-shot SR \\
        MDSR \cite{CVPRW2017EnhancedLim} & multi-scale deep SR system                 & & \\
        \hline
    \end{tabular}
\end{table*}



\end{document}